%% file: main.tex
\documentclass[10pt,twocolumn,letterpaper]{article}
\usepackage{leon_arxiv}
\input{preamble}
\definecolor{linkblue}{rgb}{0.21,0.49,0.74}
\usepackage[pagebackref,breaklinks,colorlinks,allcolors=linkblue]{hyperref}
\hypersetup{
  pdftitle={LEON: Operator-Structured Latent Evolution for World Modeling},
  pdfauthor={Xiaoxiao Lu, Jiahao Shi, Yunlong Dong, Ye Yuan},
  pdfsubject={World modeling},
  pdfkeywords={world models, latent dynamics, robot learning}
}
\title{LEON: Operator-Structured Latent Evolution for World Modeling}
\input{authors}

\begin{document}
\maketitle
\input{sections/00_abstract}
\input{sections/01_introduction}

\input{sections/02_related_work}

\input{sections/03_dynamics}
\input{sections/04_method}
\input{sections/05_experiments}
\input{sections/06_conclusion}

\clearpage
\input{llm_usage}
{\small
\bibliographystyle{ieeenat_fullname}
\bibliography{references}
}

\clearpage
\onecolumn
\appendix
\input{appendix_setup}
\input{appendices/appendix_a_theory}
\input{appendices/appendix_b_specification}
\input{appendices/appendix_c_integration}

\input{appendices/appendix_d_protocols}
\input{appendices/appendix_e_results_ablations}
\input{appendices/appendix_f_dynamical_system_diagnostics}
\end{document}

%% file: preamble.tex
\usepackage[T1]{fontenc}
\usepackage[utf8]{inputenc}
\usepackage{microtype}
\usepackage{amsthm,mathtools,bm}
\usepackage{tabularx,array,multirow,makecell,colortbl}
\usepackage{ragged2e}
\usepackage{adjustbox}
\newtheorem{proposition}{Proposition}[section]
\newcolumntype{Y}{>{\raggedright\arraybackslash}X}

\newcommand{\LEONfigure}[3]{%
  \IfFileExists{#3}{%
    \includegraphics[width=#1]{#3}%
  }{%
    \begingroup
    \setlength{\fboxsep}{0pt}%
    \fbox{\parbox[c][#2][c]{\dimexpr#1-2\fboxrule\relax}{%
      \centering\sffamily\small Figure placeholder\par\medskip
      \ttfamily\footnotesize\detokenize{#3}\par\medskip
      \normalfont\footnotesize Upload this file to replace the placeholder.
    }}%
    \endgroup
  }%
}

\usepackage{float,longtable}
\newfloat{algorithm}{htbp}{loa}[section]
\floatname{algorithm}{Algorithm}
\usepackage{algpseudocode}
\providecommand{\R}{\mathbb R}
\providecommand{\E}{\mathbb E}
\providecommand{\diag}{\operatorname{diag}}
\providecommand{\spn}{\operatorname{span}}
\providecommand{\rank}{\operatorname{rank}}
\providecommand{\LN}{\operatorname{LN}}

\providecommand{\Tref}{T_{\mathrm{ref}}}

%% file: authors.tex
\author{%
Xiaoxiao Lu\textsuperscript{1},
Jiahao Shi\textsuperscript{1},
Yunlong Dong\textsuperscript{2},
Ye Yuan\textsuperscript{1,}\thanks{Corresponding author}\\[4pt]
{\small\textsuperscript{1}School of Artificial Intelligence and Automation, Huazhong University of Science and Technology}\\
{\small\textsuperscript{2}Principia AI}\\[2pt]
{\small\textsuperscript{1}\{xiaolu, jiahaoshi, yye\}@hust.edu.cn\qquad
\textsuperscript{2}yunlongdong@outlook.com}%
}
\date{}

%% file: sections/00_abstract.tex
\begin{abstract}
Modeling action-conditioned physical evolution is fundamental to latent world models. World Action Models (WAMs) couple such models with policy learning through predictive supervision or future-conditioned action generation. Yet their latent transitions typically rely on transformer-based predictors, leaving dynamical structure to be learned implicitly from data. 
We derive a latent-transition formulation from controlled mechanical evolution. A common finite-dimensional projection of observable dynamics yields a shared operator family affine in physical control and an explicit projection residual. 
Guided by this principle, we introduce the Latent Evolution Operator Network (LEON), which adapts pretrained visual features into learned observables and evolves them through an affine family of shared operator matrices and complementary additive updates. Coefficients inferred from evolving observables and action-related context modulate the operators. 
We instantiate LEON in two WAM architectures while preserving their distinct original prediction targets and policy interfaces. 
Across single-arm and bimanual manipulation, LEON improves average policy success over corresponding WAM baselines, including under diverse perturbations. Ablations support its observable adaptation and operator-structured evolution, while controlled-system evaluations demonstrate improved extrapolation beyond training conditions. 
\end{abstract}

%% file: sections/01_introduction.tex
\section{Introduction}

Anticipating how the physical environment evolves during interaction is central to world modeling for embodied agents~\citep{hafner2020dreamer}.
Latent world models realize this modeling through future prediction in encoded visual feature spaces, without requiring pixel-level reconstruction~\citep{zhou2025dinowm,assran2025v,sun2026vlajepa}. Such predictions are integrated into policy learning in robot manipulation, serving either as training-time supervision or as predicted-future conditions for action generation, as exemplified by recent World Action Models (WAMs)~\citep{sun2026vlajepa,chen2026lawam,yuan2026fastwam}. 

\begin{figure}[t]
    \centering
    \includegraphics[width=\columnwidth]{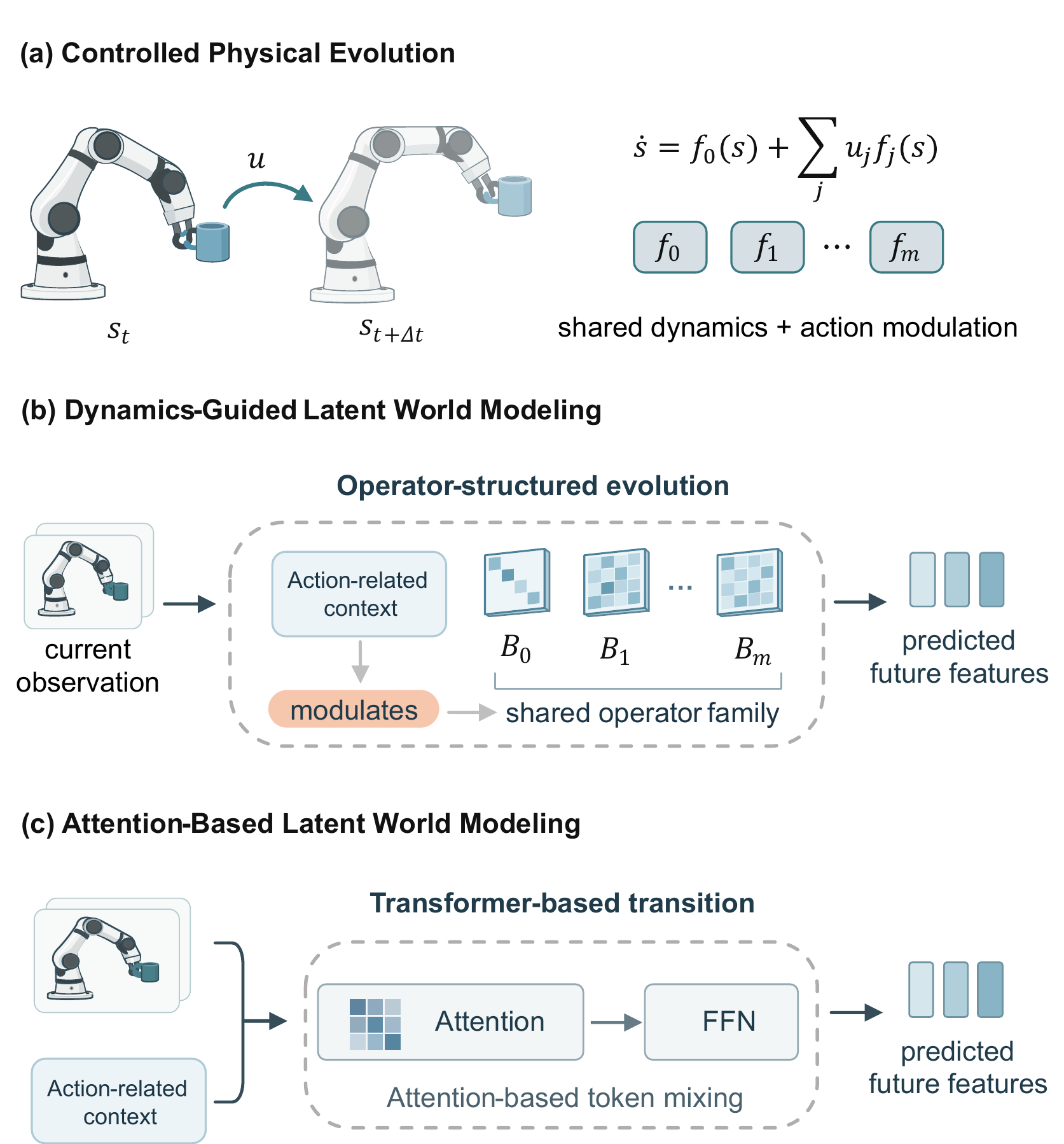}
    \caption{
    \textbf{From physical evolution to operator-structured latent prediction.}
    Control-affine dynamics motivate LEON's observable-space evolution through shared, context-modulated operators and complementary additive updates with repeated Euler steps.
    In contrast, the Transformer-based methods predict future features using stacked attention and feed-forward blocks.
    }
    \label{fig:intro_transition}
\end{figure}

At the core of this future prediction is a latent transition from current visual representations and action-related context to future representations. The physical dynamics underlying these transitions follow common equations across different states and controls within a given mechanical system~\citep{spong2020robot}. Visual observation and encoding recast the same dynamics in the features used by the model~\citep{watter2015embed,lusch2018deep}, while its underlying dynamics remain unchanged. Representative world models employ Transformer-based predictors that learn latent transitions through attention-based feature interactions~\citep{sun2026vlajepa,chen2026lawam}, leaving this shared dynamical organization to be learned implicitly from data rather than explicitly encoded in the predictor. This raises a central question: How can controlled physical dynamics guide the internal structure of latent transitions in visual world models?


For robot--environment systems, mechanical dynamics admit a control-affine form, where physical control inputs weight a shared set of state-dependent vector fields~\citep{spong2020robot}. To express this structure in latent prediction, we describe the physical state through differentiable observables and characterize their instantaneous evolution using Koopman generators~\citep{proctor2018koopmancontrol,peitz2020interpolated,otto2024bilinear}. By the chain rule, the Koopman generator decomposes affinely in physical control into shared generator components. Projecting the observable derivatives onto a common finite-dimensional observable subspace yields a control-affine family of evolution matrices and a residual for unresolved change~\citep{klus2020data}. As the matrix representation and residual depend on the chosen observable subspace, we use this decomposition to guide a latent transition design with jointly learned observables, condition-modulated shared operator evolution, and complementary updates.

Building on this principle, we introduce the Latent Evolution Operator Network (LEON), which instantiates the finite-observable evolution derived above within a learnable latent world model. LEON adapts pretrained visual features into learned observables. Specifically, we construct an affine operator family from shared evolution components, with coefficients inferred from the evolving observables and action-related representations to adapt the operator to different states and conditions. We further introduce a complementary additive update to model changes beyond the restricted operator family, forming a shared rule for latent evolution. We instantiate LEON in VLA-JEPA and LaWAM, representing prediction-supervised policy learning and predicted-future-conditioned action generation, respectively, fully replacing their Transformer-based transitions while preserving the original prediction targets and policy interfaces~\citep{sun2026vlajepa,chen2026lawam}.

Our contributions are summarized as follows:
\begin{itemize}
\item We derive a dynamics-grounded formulation for latent transition design, realizing controlled mechanical evolution with shared operator structure in observable space.
    
\item We propose LEON, which evolves learned visual observables through a condition-modulated shared affine operator family and complementary nonlinear additive updates.

\item We validate LEON across two prediction--policy interfaces, demonstrating improved single-arm and bimanual manipulation performance and robustness to diverse perturbations.

\end{itemize}

%% file: sections/02_related_work.tex
\section{Related Work}

\begin{figure*}[t]
    \centering
    \includegraphics[width=\textwidth]{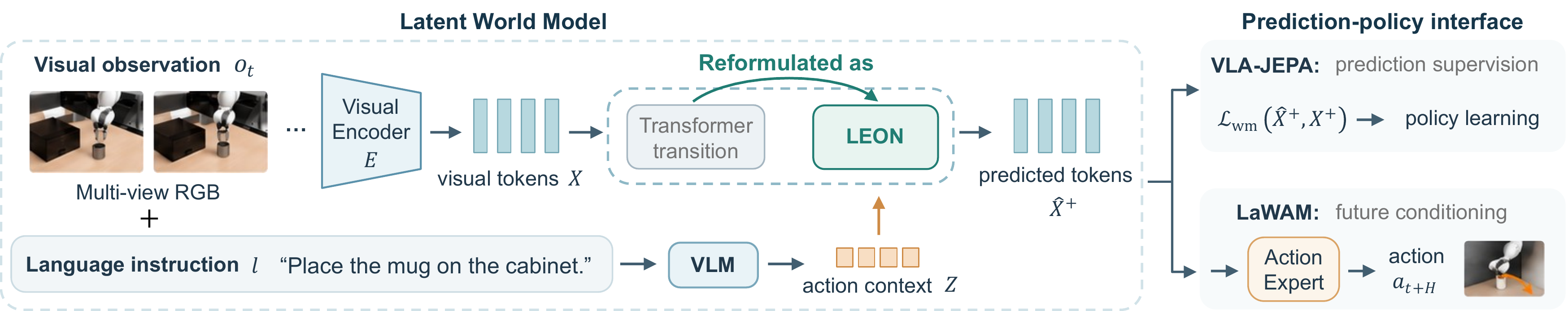}
    \caption{\textbf{Operator-structured latent transitions across prediction--policy interfaces.}
LEON reformulates Transformer-based latent prediction through operator-structured evolution conditioned on visual features $X$ and action-related context $Z$. Future prediction supervises policy learning in VLA-JEPA and conditions action generation in LaWAM.}
    \label{fig:leon_wam_interfaces}
\end{figure*}

\noindent\textbf{World models for robot manipulation.}
World models have become an increasingly prominent component of robot manipulation, where future prediction augments policy learning and execution with explicit modeling of action consequences. Existing methods differ in the representation of the predicted future: WorldVLA and IRASim generate future images or videos~\citep{cen2025worldvla,zhu2025irasim}, whereas DINO-WM and VLA-JEPA predict future features extracted by pretrained visual encoders~\citep{zhou2025dinowm,sun2026vlajepa}.
LaDi-WM also models future states in a latent space aligned with pretrained visual foundation models~\citep{huang2025ladi}. Their use of prediction within the policy also varies. VLA-JEPA and Fast-WAM employ predictive objectives to supervise representation learning for action generation, without requiring future prediction at inference~\citep{sun2026vlajepa,yuan2026fastwam}, whereas Seer, CoT-VLA, and LaWAM condition action generation on predicted visual states or latent subgoals~\citep{tian2025predictive,zhao2025cotvla,chen2026lawam}. These works establish different predictive representations and prediction--policy interfaces, while latent evolution is typically modeled by Transformer-based latent predictors.
Recent work further examines how action-conditioning mechanisms and predictor depth affect planning performance in world models~\citep{terver2025drives}.
However, these predictor design choices leave the shared dynamical structure of controlled state evolution implicit.
LEON instead draws on this dynamical structure to jointly design learned visual observables and their action-conditioned evolution.

\par\addvspace{3pt}\noindent\textbf{Koopman-based dynamics.}
Koopman-based methods seek representations in which nonlinear dynamics admit simpler linear predictions. 
KOROL learns image-derived object features so a fitted linear model can reproduce demonstrated robot trajectories~\citep{chen2024korol}.
RoboKoop learns spectral visual coordinates that support linear feedback control within reinforcement learning~\citep{kumawat2024robokoop}.
PK-NN learns a common representation for parameter-varying systems, enabling parameter-dependent linear one-step prediction~\citep{guo2025learning}.
Koopman Dreamer constrains latent dynamical modes to regulate error accumulation during long-horizon imagination~\citep{li2026koopman}. Peitz et al. develop reduced-order dynamical models to support control optimization, while Otto et al. identify such dynamics from partially observed trajectories through latent-state estimation~\citep{peitz2020interpolated,otto2024bilinear}. LEON builds on controlled mechanical dynamics to guide visual latent-transition design, linking observable adaptation to action-dependent evolution and complementary modeling of unresolved dynamics.
This connection motivates a learned evolution rule whose operator coefficients adapt to the evolving visual state and action-related context, alongside complementary nonlinear updates.

%% file: sections/03_dynamics.tex
\section{Transition Design Principles from Dynamics}
\label{sec:leon_physical_dynamics}

World models predict how physical systems evolve under action. We formulate this evolution as a control-affine system and derive its corresponding structure in observable space. The decomposition yields observable-dependent shared operators, control-modulated evolution, and unresolved change, guiding the transition architecture in Section~4.

\par\addvspace{3pt}\noindent\textbf{Controlled Evolution problem formulation.}
Consider a robot--environment system within a smooth interaction regime. Its state evolution is governed by
$M(q)\ddot q+c(q,\dot q)=B(q)u$,
where $q$ denotes generalized coordinates, $u\in\mathbb R^m$ physical controls, $M$ a nonsingular inertia matrix, $c$ non-actuation terms, and $B$ the control-to-force mapping~\citep{spong2020robot}. With the physical state $s=(q,\dot q)\in\mathbb R^{d_s}$, the second-order dynamics take the first-order control-affine form
\begin{equation}
\dot s=f(s,u)=f_0(s)+\sum_{j=1}^{m}u_jf_j(s).
\label{eq:leon_control_affine}
\end{equation}
For a given system, the drift $f_0$ and input vector fields $\{f_j\}_{j=1}^m$ are fixed, potentially nonlinear functions of the state, while each control signal $u_j$ modulates the corresponding field. Latent transitions aim to model this evolution.

\par\addvspace{3pt}\noindent\textbf{Control affinity under observable mapping.}
To represent controlled state evolution through observables, let $\Phi=(\varphi_1,\ldots,\varphi_d)^\top:\mathbb R^{d_s}\to\mathbb R^d$ be a differentiable observable map and define $h=\Phi(s)$. For fixed input $u$, the Koopman generator $L_u$ gives the instantaneous rate of change of an observable along the dynamics~\citep{proctor2018koopmancontrol}. By the chain rule,
$\dot h=(L_u\Phi)(s)=J_\Phi(s)f(s,u)$,
where $J_\Phi$ denotes the Jacobian of $\Phi$. Substituting Eq.~\eqref{eq:leon_control_affine} yields
$L_u=L_0+\sum_{j=1}^{m}u_jL_j$, where $(L_j\Phi)(s)=J_\Phi(s)f_j(s)$ for $j=0,\ldots,m$. Thus, physical controls modulate shared generator components $\{L_j\}_{j=1}^{m}$, preserving affine control dependence in instantaneous observable evolution~\citep{peitz2020interpolated}. To obtain these shared generators in a latent transition, we seek matrices that represent their action on $h$. Since such matrices may not fully represent observable evolution~\citep{otto2024bilinear}, we introduce a common projection and retain the residual.

\par\addvspace{3pt}\noindent\textbf{Shared operator–residual decomposition.}
Let $\mathcal V_\Phi=\operatorname{span}\{\phi_1,\ldots,\phi_d\}$ be the finite observable subspace, and let $P_\Phi$ be a fixed linear projection onto this subspace, applied componentwise and shared across all generators. Each projected generator admits a matrix representation $P_\Phi L_j\Phi=G_j\Phi$,
with $G_j\in\mathbb R^{d\times d}$~\citep{klus2020data}. Defining
$r_j=(I-P_\Phi)L_j\Phi$, where $I$ is the identity, gives the exact decomposition
\begin{equation}
\dot h=G(u)h+r(s,u),
\label{eq:leon_bilinear1}
\end{equation}
where $G(u)=G_0+\sum_{j=1}^{m}u_jG_j$ is the shared operator
family and $r(s,u)=r_0(s)+\sum_{j=1}^{m}u_jr_j(s)$ is the
projection residual.
The common linear projection preserves the original affine control dependence. An orthogonal projection construction is detailed in the supplementary material.

The control-dependent part $G(u)-G_0$ lies in
$\mathcal S_G=\operatorname{span}\{G_1,\ldots,G_m\}$, with $r_*=\dim(\mathcal S_G)\le m$ denoting the number of independent operator directions. Choosing a basis
$\{\widetilde G_k\}_{k=1}^{r_*}$ gives
$G(u)=G_0+\sum_{k=1}^{r_*}(Tu)_k\widetilde G_k$,
where $T\in\mathbb R^{r_*\times m}$ maps physical controls to coefficients in this basis. The choice of observables affects which changes the shared operators express and which remain unresolved, motivating joint learning of the observable representation and its control-dependent evolution in latent transition.

\par\addvspace{3pt}\noindent\textbf{Discrete observable evolution.}
\label{discrete}
If $\dot h$ is Lipschitz continuous on $[t,t+\Delta t]$ for $\Delta t>0$, integrating Eq.~\eqref{eq:leon_bilinear1} gives
\begin{equation}
h^{+}
=
h+\Delta t\left[G(u)h+r(s,u)\right]
+\mathcal O(\Delta t^2),
\label{eq:leon_discrete_principle}
\end{equation}
where $h^{+}=h(t+\Delta t)$ and the increment is evaluated at $t$. Neglecting the remainder yields a forward-Euler update combining operator evolution and residual change. The decomposition and discretization motivate the joint design of observable representations and their control-modulated evolution, with complementary updates for unresolved change. Section~4 translates this structure into a transition architecture with context-conditioned operators and shared updates.

\begin{figure*}[t]
\centering
\LEONfigure{\textwidth}{62mm}{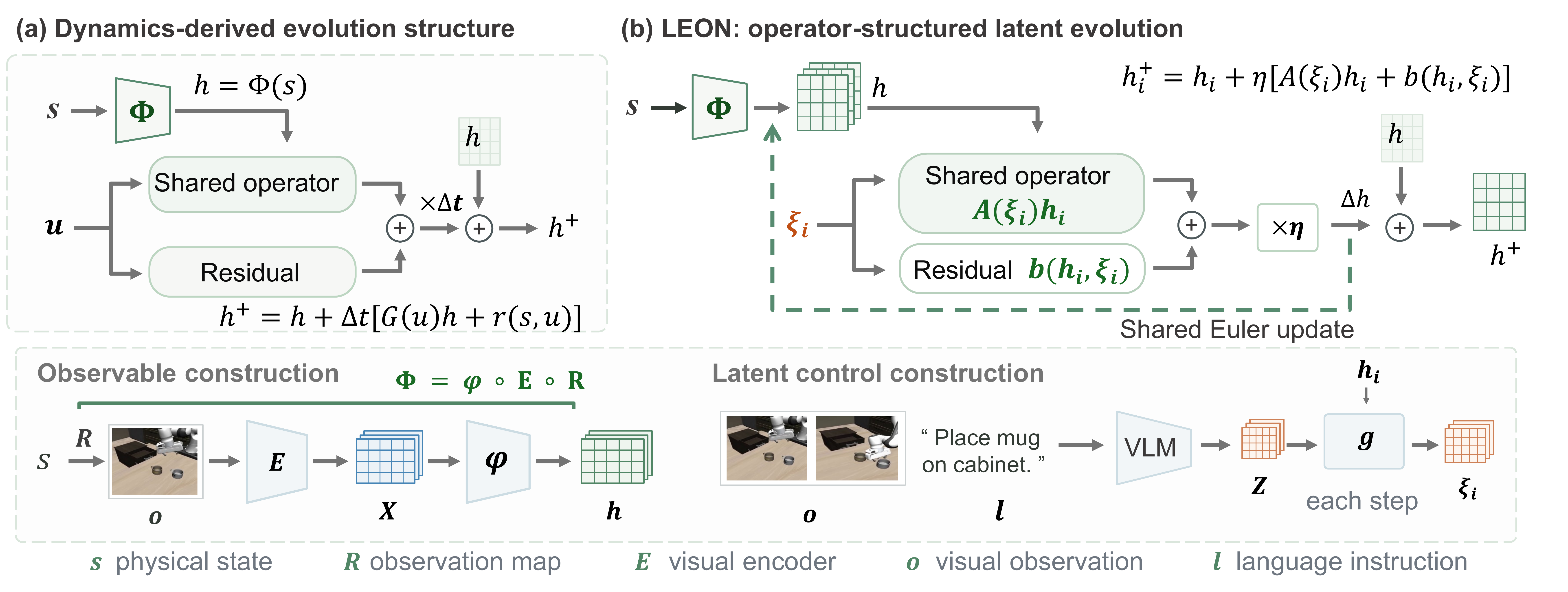}
\caption{\textbf{From controlled dynamics to operator-structured latent evolution.}
(a) Controlled dynamics motivate observable evolution as shared operator propagation with complementary residual change.
(b) LEON realizes this principle in learned visual observables using shared latent-control-modulated operators and additive residual updates. The observable and latent-control constructions are shown below.}
\label{fig:leon-overview}
\end{figure*}

%% file: sections/04_method.tex
\section{Latent Evolution Operator Network}
\label{sec:leon}


We introduce the \emph{Latent Evolution Operator Network} (LEON), a dynamics-guided architecture for latent transition modeling. We first formulate the prediction problem and outline the proposed framework, followed by its operator-structured evolution design and instantiation across two prediction--policy interfaces.


\subsection{Overview}

\par\addvspace{3pt}\noindent\textbf{Problem formulation.}
We consider visual latent prediction of the controlled robot--environment evolution formulated in Section~\ref{sec:leon_physical_dynamics}. Given an observation $o=R(s)$ of physical state $s$ and a language instruction $l$, the pretrained visual encoder $E$ provides features $X=E(o)$, while the WAM supplies action-related context $Z$ through its VLM conditioned on $(o,l)$. Here, $R$ denotes the physical visual observation map. The prediction task is to estimate the future visual representation $X^+$ from $(X,Z)$. To instantiate the observable representation of Section~3, LEON constructs $h=\varphi(X)=\Phi(s)$, where $\phi$ is a learned adaptation mapping and $\Phi=\varphi\circ E\circ R$.

\par\addvspace{3pt}\noindent\textbf{Framework overview.}
The decomposition in Eq.~\eqref{eq:leon_bilinear1} separates observable evolution into a projected operator term $G(u)h$ and a residual $r(s,u)$. LEON translates this structure into learnable components: $A(\xi)h$ models operator-structured evolution through shared matrices modulated by latent coefficients $\xi$, while $b(h,\xi)$ provides a complementary update motivated by the projection residual. The coefficients $\xi$ are inferred from evolving observables and WAM action-related context $Z$, without recovering the physical control $u$. Following the forward-Euler form of Eq.~\eqref{eq:leon_discrete_principle}, LEON updates observables as
\begin{equation}
h' = h+\eta\left[A(\xi)h+b(h,\xi)\right],
\label{eq:leon_formulation}
\end{equation}
where $A(\xi)$ is affine in $\xi$, and $\eta$ denotes the normalized-time step size. A learned readout maps the updated observable back to the retained prediction space.

\subsection{Operator-Structured Latent Transitions}
\label{sec:leon_transition}

\paragraph{Learned observable space.}
Given the retained visual feature tokens
$X=[x_1,\ldots,x_N]^\top=E(o)\in\mathbb{R}^{N\times d_x}$,
LEON constructs observables through the token-wise mapping
$h_i=\varphi(x_i)\in\mathbb{R}^{d_h}$,
where $\varphi$ combines a linear projection of layer-normalized features with an MLP residual branch.
Applying $\varphi$ once to $X$ with shared parameters yields the initial observable representation
$h=[h_1,\ldots,h_N]^\top=\Phi(s)$ for subsequent updates,
where $\Phi=\varphi\circ E\circ R$.
Learning $\varphi$ jointly with the evolution operators allows this representation to adapt to the restricted operator family.

\par\addvspace{3pt}\noindent\textbf{Observable-conditioned latent control.}
Let $Z$ denote the WAM's action-related representation inferred from the current observation and language instruction. For each observable $h_i$, we construct latent control coefficients
$\xi_i=\tanh\!\big(f_\xi(g_{\mathrm{fuse}}(h_i,Z))\big)\in(-1,1)^M$,
where $g_{\mathrm{fuse}}$ combines $h_i$ with $Z$, and $f_\xi$ maps the fused representation to $\mathbb{R}^M$. Conditioning on the current observable allows the same action-related context to modulate evolution differently across tokens and evolving states. The bounded, signed components of $\xi_i$ weight the shared operator bases, and $\xi_i$ also conditions the complementary update $b(h_i,\xi_i)$. Although the control construction is nonlinear in $(h_i,Z)$, the operator $A(\xi_i)$ remains affine in $\xi_i$. Section~\ref{sec:leon_integration} specifies the fusion map for each WAM instantiation.

\par\addvspace{3pt}\noindent\textbf{Shared operator family and complementary update.}
To realize the shared affine structure in the learned observable space, LEON parameterizes the operator family as
\begin{equation}
A(\xi_i)
=
D+\sum_{k=1}^{M}\xi_{ik}B_k,
\qquad B_k\in\mathbb{R}^{d_h\times d_h}.
\label{eq:leon_operator}
\end{equation}
Here, $D=\operatorname{diag}(d)$ is a learned diagonal matrix, and $\{B_k\}_{k=1}^{M}$ are learned operator matrices shared across observables and evolution updates. The diagonal term provides a latent-control-independent linear contribution to individual observable components, while the shared matrices allow their interactions to be modulated by the latent control. Let $\mathcal{S}_A=\operatorname{span}\{B_1,\ldots,B_M\}$. Then $A(\xi_i)-D\in\mathcal{S}_A$, with $\dim(\mathcal{S}_A)\le M$. Thus, the latent control modulates the operator within a shared matrix subspace of dimension at most $M$, while each operator component can represent general interactions among the learned observables.

The unresolved contribution $r(s,u)$ in Section~\ref{sec:leon_physical_dynamics} motivates a learned complementary update $b(h_i,\xi_i)$ for changes beyond the restricted operator family. We implement $b$ as a normalized two-layer MLP conditioned on the observable and latent control, with $b(h_i,\xi_i)=g_b(h_i+g_p(\xi_i))$, where $g_p$ embeds the latent control coefficients into the observable space. The operator and complementary branches are both initialized to produce zero observable increments.

\par\addvspace{3pt}\noindent\textbf{Composition of operator-structured updates.}
For an input observation at time $t_0$, let $\Delta t_{\mathrm{pred}}$ denote the time interval to its future prediction target, determined by the video sampling interval and target-frame offset. We normalize the prediction interval as $\tau=(t-t_0)/\Delta t_{\mathrm{pred}}\in[0,1]$ and divide it into $S=8$ uniform Euler steps with $\eta=1/S$. Starting from $h_i^{(0)}=h_i$, LEON composes the operator-structured updates in Eq.~\eqref{eq:leon_formulation} as
\begin{equation}
h_i^{(n+1)}
=
h_i^{(n)}
+\eta\left[
A(\xi_i^{(n)})h_i^{(n)}
+b(h_i^{(n)},\xi_i^{(n)})
\right],
\label{eq:leon_composition}
\end{equation}
for $n=0,\ldots,S-1$. The operator matrices $D$ and $\{B_k\}$, together with the latent-control and complementary-update networks, are shared across all steps, while the coefficients $\xi_i^{(n)}=\xi(h_i^{(n)},Z)$ are recomputed from the evolving observables under the fixed action-related context $Z$. After the final step, a single linear readout maps the accumulated observable change to the predictor feature space, yielding $\widetilde{x}_i^+=x_i+W_\psi(h_i^{(S)}-h_i^{(0)})$. The retained WAM output projection maps the updated features to the target representation space, yielding predicted future visual representations.

\subsection{Prediction--Policy Interface Instantiations}
\label{sec:leon_integration}

We instantiate LEON in two representative WAM architectures with distinct prediction--policy interfaces, while preserving their original prediction targets and policy interfaces. Both instantiations use the same operator-structured transition formulation, but differ in the construction of the conditioning map $g_{\mathrm{fuse}}$ and in how future predictions contribute to policy learning and action generation.

\par\addvspace{3pt}\noindent\textbf{Prediction-supervised policy learning.}
In VLA-JEPA~\citep{sun2026vlajepa}, future prediction supervises policy representation learning without providing an explicit future condition to the action head at inference. Given the current observation and task instruction, the VLM provides $K$ latent-action tokens $Z\in\mathbb{R}^{K\times d_Z}$ to condition future visual feature prediction. We construct $g_{\mathrm{fuse}}$ by combining observable-specific and global action summaries. Projecting and normalizing $Z$ to the observable dimension gives $\bar Z$. The local summary is $c_i^{\mathrm{loc}}=\operatorname{MHA}(\operatorname{LN}(h_i),\bar Z,\bar Z)$, where $\operatorname{MHA}$ denotes multi-head attention with the normalized observable as the query and $\bar Z$ as the keys and values. In parallel, a two-layer MLP computes the global summary $c^{\mathrm{glob}}=g_{\mathrm{glob}}(\operatorname{Mean}(\bar Z))$. The summaries are combined as $g_{\mathrm{fuse}}(h_i,Z)=g_{\mathrm{merge}}([c^{\mathrm{glob}};c_i^{\mathrm{loc}}])$, where $[\cdot;\cdot]$ denotes concatenation along the feature dimension, and $g_{\mathrm{merge}}$ applies normalization followed by an affine projection. Finally, $f_\xi$ and $\tanh$ map the fused representation to the latent control coefficients $\xi_i$.

The retained WAM output mapping transforms LEON's updated features into the future visual representation $\widehat X^{+}$ in the frozen V-JEPA target space. We retain the task objective $\mathcal L_{\mathrm{task}}=\mathcal L_{\mathrm{act}}+\beta\mathcal L_{\mathrm{WM}}$, where $\mathcal L_{\mathrm{WM}}$ supervises future prediction using $X^{+}$ as the target, and $\mathcal L_{\mathrm{act}}$ supervises action generation. We additionally regularize the LEON coefficient
preactivations, giving
$\mathcal{L}_{\mathrm{train}} = \mathcal{L}_{\mathrm{task}} +
\lambda_{\xi}\mathcal{L}_{\xi}$ (Appendix~\ref{app:leon-objectives}). Gradients of $\mathcal L_{\mathrm{WM}}$ propagate through LEON and the latent-action pathway to the shared VLM, shaping the embodied-action representations that condition the action head. At inference, the action head generates action sequences conditioned on the embodied-action representations, without conditioning on predicted future features.

\par\addvspace{3pt}\noindent\textbf{Future-conditioned action generation.}
In LaWAM~\citep{chen2026lawam}, the predicted future representation directly conditions action generation. The current-state features and future prediction targets remain in the frozen DINOv3 feature space. The WAM supplies a single aggregated latent-action representation $Z$ from the current observation and language instruction, so no further token-level action aggregation is needed. After the retained action projection $P_Z$ and normalization, a two-layer MLP forms the global summary
$c^{\mathrm{glob}}=g_{\mathrm{glob}}(\operatorname{LN}(P_Z(Z)))\in\mathbb R^{d_h}$.
This summary is fused with each observable as
$g_{\mathrm{fuse}}(h_i,Z)=g_{\mathrm{merge}}([h_i;c^{\mathrm{glob}}])$,
where $g_{\mathrm{merge}}:\mathbb R^{2d_h}\to\mathbb R^{d_h}$ is a normalized two-layer MLP. Following Section~\ref{sec:leon_transition}, $f_\xi$ and $\tanh$ map the fused representation to the latent control $\xi_i$, which weights the shared operator directions and, together with $h_i$, conditions the complementary update $b(h_i,\xi_i)$.

LEON predicts the future representation $\widehat X^{+}$ from $(X,Z)$. Training retains the task objective
$\mathcal L_{\mathrm{task}}=\mathcal L_{\mathrm{act}}+\lambda_{\mathrm{WM}}\mathcal L_{\mathrm{WM}}+\lambda_{\mathrm{distill}}\mathcal L_{\mathrm{distill}}$,
where $\mathcal L_{\mathrm{WM}}$ supervises future prediction using $X^{+}$ as the target, and $\mathcal L_{\mathrm{act}}$ supervises action generation. The distillation loss $\mathcal L_{\mathrm{distill}}$ aligns $Z$ with a teacher latent action inferred from the observed current--future pair. The retained flow-matching action expert is conditioned on $X$, $\widehat X^{+}$, and the VLM context $H_{\mathrm{VLM}}$ from which $Z$ is obtained. As in LaWAM, $\mathcal L_{\mathrm{act}}$ does not backpropagate through the predicted-future input, while $\mathcal L_{\mathrm{WM}}$ updates LEON and the latent-action pathway. At inference, $Z$ and $\widehat X^{+}$ are computed from the current observation and language instruction, and the action expert generates actions conditioned on $(X,\widehat X^{+},H_{\mathrm{VLM}})$. Appendix~C.3 details the feature pipelines, losses, and gradient paths for both WAMs.

%% file: sections/05_experiments.tex
\section{Experiments}
\label{sec:experiments}

\begin{table*}[t]
\centering
\caption{
\textbf{LIBERO success rates (\%).}
Prior-method results are taken from the original papers unless marked $\dagger$, indicating our re-evaluation of the released VLA-JEPA and LaWAM checkpoints under their respective evaluation protocols.
The $\dagger$ and LEON results use three rollout seeds and 50 trials per task per seed; seed-wise results are reported in Table E.1.
Avg.\ denotes the mean across 40 tasks.
Best and second-best results are shown in bold and underlined, respectively.
}
\label{tab:libero}

\small
\setlength{\tabcolsep}{5pt}
\renewcommand{\arraystretch}{1.08}

\begin{tabular*}{0.84\textwidth}{@{\extracolsep{\fill}}lccccc@{}}
\toprule
\textbf{Method} &
\textbf{Spatial} &
\textbf{Object} &
\textbf{Goal} &
\textbf{LIBERO-10} &
\textbf{Avg.} \\
\midrule

OpenVLA-OFT \citep{kim2025openvlaoft}
& 97.6 & 98.4 & 97.9 & 94.5 & 97.1 \\

$\pi_{0.5}$ \citep{black2025pi05}
& 98.8 & 98.2 & 98.0 & 92.4 & 96.9 \\

UniVLA \citep{bu2025univla}
& 96.5 & 96.8 & 95.6 & 92.0 & 95.2 \\

WorldVLA \citep{cen2025worldvla}
& 87.6 & 96.2 & 83.4 & 60.0 & 81.8 \\

Fast-WAM \citep{yuan2026fastwam}
& 98.2 & \textbf{100.0} & 97.0 & 95.2 & 97.6 \\

Cosmos-Policy \citep{kim2026cosmospolicy}
& 98.1 & \textbf{100.0} & 98.2 & 97.6 & 98.5 \\

LingBot-VA \citep{li2026lingbotva}
& 98.5 & 99.6 & 97.2 & \textbf{98.5} & 98.5 \\

\midrule

VLA-JEPA$^\dagger$ \citep{sun2026vlajepa}
& 98.0 & 99.7 & 98.4 & 95.4 & 97.9 \\

LaWAM$^\dagger$ \citep{chen2026lawam}
& 98.1 & 98.9 & 98.1 & 97.1 & 98.1 \\

\addlinespace[2pt]

\textbf{VLA-JEPA + LEON}
& \textbf{99.2} & 99.7 & \textbf{98.9} & \underline{98.1} & \textbf{99.0} \\

\textbf{LaWAM + LEON}
& \underline{99.1} & \underline{99.8} & \underline{98.5} & 97.5 & \underline{98.8} \\

\bottomrule
\end{tabular*}
\end{table*}

We organize the experiments around three questions: (i) whether LEON improves manipulation success across distinct prediction--policy interfaces, evaluated in VLA-JEPA and LaWAM on LIBERO and in LaWAM on RoboTwin 2.0; (ii) whether the proposed transition design and its components improve policy performance and perturbation robustness, assessed through Transformer and Residual MLP comparisons and component ablations on LIBERO and LIBERO-Plus; and (iii) whether the structured evolution supports extrapolative state prediction and how its update components contribute, examined on a nonlinear pendulum and a driven oscillator independently of policy learning. For the VLA-JEPA transition comparisons, the retained WAM components and prediction targets are preserved, with LIBERO-trained variants evaluated on LIBERO-Plus without additional adaptation. Detailed training and evaluation protocols are provided in the supplementary material.


\subsection{Policy Performance Across WAMs}
\label{sec:policy_performance}

LIBERO~\citep{liu2023libero} spans spatial, object, goal, and long-horizon manipulation, providing a shared multi-task setting for evaluating policy performance. Both LEON instantiations are jointly fine-tuned on the four LIBERO suites, following the corresponding WAM data-processing protocols. Both integrations preserve their original prediction targets and policy interfaces, enabling a focused evaluation of the latent-transition design. As shown in Table~\ref{tab:libero}, LEON increases the average SR from 97.9\% to 99.0\% in VLA-JEPA and from 98.1\% to 98.8\% in LaWAM, relative to their re-evaluated released checkpoints. VLA-JEPA + LEON improves three suites while maintaining comparable performance on Object, whereas LaWAM + LEON improves all four suites. Both variants achieve the highest average success rates among the compared methods. These results support the effectiveness of the proposed latent-transition formulation across the two distinct prediction--policy interfaces.

\begin{table}[!t]
\centering
\caption{
\textbf{RoboTwin 2.0 success rates (\%).}
Clean and Randomized report mean success rates across 50 tasks, with 100 trials per task under each condition.
Avg.\ denotes the arithmetic mean of the two condition scores.
Prior-method results follow LaWAM.
Best and second-best results are shown in bold and underlined, respectively.
}
\label{tab:robotwin}
\small
\setlength{\tabcolsep}{5pt}
\renewcommand{\arraystretch}{1.12}
\begin{tabular*}{0.96\columnwidth}{@{\extracolsep{\fill}}lccc@{}}
\toprule
\textbf{Method} & \textbf{Clean} & \textbf{Randomized} & \textbf{Avg.} \\
\midrule
Fast-WAM        & 91.98 & \underline{90.52} & \underline{91.25} \\
GigaWorld-Policy & 86.36 & 85.04 & 85.70 \\
LingBot-VA      & 91.50 & \textbf{90.92} & 91.21 \\
$\pi_{0.5}$     & 82.74 & 76.76 & 79.75 \\
Motus           & 88.66 & 87.02 & 87.84 \\
LaWAM           & \underline{92.64} & 89.80 & 91.22 \\
\midrule
\textbf{LaWAM + LEON}
                & \textbf{93.32} & 90.42 & \textbf{91.87} \\
\bottomrule
\end{tabular*}
\vspace{-7pt}
\end{table}


RoboTwin 2.0~\citep{chen2025robotwin2} extends the policy-facing evaluation to 50 diverse bimanual manipulation tasks requiring coordinated dual-arm control. Compared with the single-arm LIBERO tasks, this benchmark introduces more complex coordination requirements across diverse object interactions and diverse visual and spatial conditions. Following mixed clean-and-randomized training, we evaluate both conditions with 100 trials per task. LEON preserves the future-conditioned action-generation interface of LaWAM during policy inference, enabling evaluation of the proposed latent-transition design in coordinated bimanual manipulation. As shown in Table~\ref{tab:robotwin}, LaWAM + LEON increases SR from 92.64\% to 93.32\% on Clean and from 89.80\% to 90.42\% on Randomized, yielding the highest overall average of 91.87\% among the compared methods. Fast-WAM and LingBot-VA remain higher on Randomized. These results support the effectiveness of the proposed latent-transition design for future-conditioned action generation across diverse bimanual tasks and domain-randomized environments.


\begin{table*}[t]
\centering
\caption{
\textbf{VLA-JEPA transition comparisons and ablations (success rate, \%).}
LIBERO reports the mean success rate across its four task suites.
LIBERO-Plus reports four representative perturbation categories and the mean across all seven categories.
LIBERO-Plus evaluations use LIBERO-trained checkpoints without additional adaptation. All results in this table use rollout seed 0 and the final LIBERO checkpoint of each variant. Best and second-best results are shown in bold and underlined, respectively.
}
\label{tab:transition-ablation}

\small
\setlength{\tabcolsep}{3.1pt}
\renewcommand{\arraystretch}{1.08}

\begin{tabular*}{\textwidth}{@{\extracolsep{\fill}}llcccccc@{}}
\toprule
\multirow{2}{*}{\textbf{Transition variant}}
& \multirow{2}{*}{\textbf{Transition form}}
& \multicolumn{1}{c}{\textbf{LIBERO}}
& \multicolumn{5}{c}{\textbf{LIBERO-Plus}} \\
\cmidrule(lr){4-8}
& & Avg.
& Camera & Light & Background & Layout & Avg. \\
\midrule

Transformer
& Transformer-based blocks
& 97.40 & 63.30 & 94.40 & 93.50 & 84.59 & 79.54 \\

Residual MLP
& Residual blocks
& 97.35 & 65.54 & 93.70 & 92.75 & 83.93 & 79.44 \\

\midrule

w/o Observable Adaptation
& $A(\xi)x+b(x,\xi)$
& 98.05 & \textbf{68.98} & 93.78 & 94.70 & 84.33 & 80.94 \\

w/o Operator Modulation
& $\bar A h+b(h,\xi)$
& 97.50 & 65.35 & 92.29 & 93.40 & 85.44 & 79.93 \\

w/o Additive
& $A(\xi)h$
& \underline{98.25} & 67.10 & \underline{95.36}
& 94.33 & \underline{86.43} & \underline{81.41} \\

w/o Operator
& $b(h,\xi)$
& 97.10 & 67.73 & 95.01
& \underline{94.98} & 85.64 & 80.32 \\

\midrule

\textbf{LEON}
& $A(\xi)h+b(h,\xi)$
& \textbf{99.05} & \underline{68.67}
& \textbf{98.34} & \textbf{97.03}
& \textbf{87.54} & \textbf{83.45} \\

\bottomrule
\end{tabular*}
\end{table*}

\subsection{Transition Comparisons and Ablations}
\label{sec:ablation}

\begin{figure}[!t]
\centering
\LEONfigure{\linewidth}{42mm}{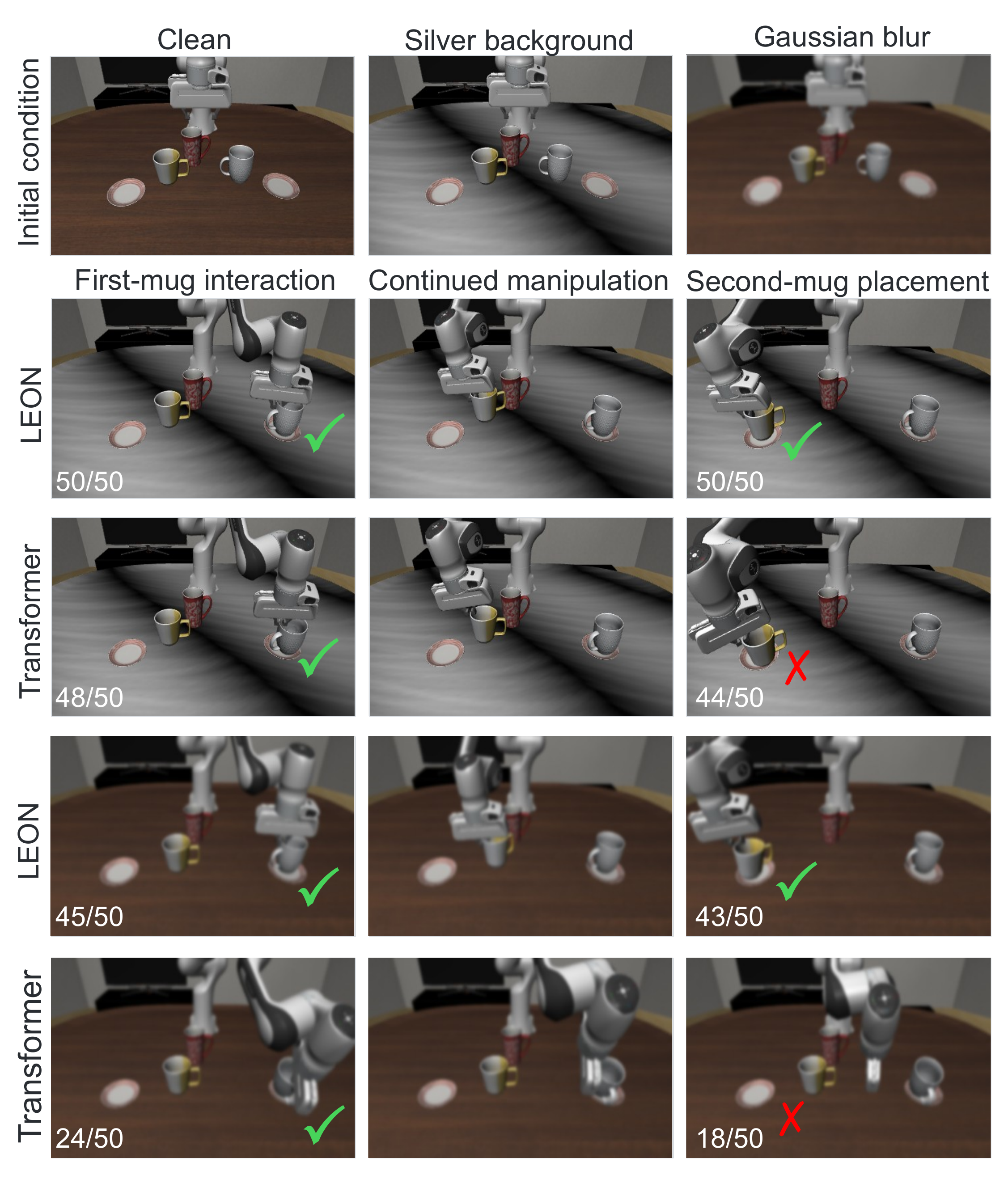}
\caption{
\textbf{Qualitative comparison of VLA-JEPA with Transformer and LEON latent transitions.}
Closed-loop executions on a long-horizon two-mug placement task from LIBERO-10 under silver-background and Gaussian-blur perturbations, with the clean observation shown for reference.
Both variants are evaluated without additional adaptation.
Annotations $x/50$ report successful trials out of 50 for first-mug placement or full-task completion.
}
\label{fig:qualitative-rollouts}
\vspace{-7pt}
\end{figure}


We evaluate the LEON transition within VLA-JEPA through comparisons with alternative transition architectures and ablations of its observable adaptation, operator modulation, and update components. All variants are trained independently using the same LIBERO data, training schedules, and pretrained initialization of the retained VLA-JEPA components. We further evaluate the corresponding LIBERO-trained checkpoints on LIBERO-Plus~\citep{fei2026liberoplus} without additional adaptation to assess policy robustness under distribution shifts across seven perturbation categories. Detailed variant definitions and results are reported in Appendix E.2, with training and evaluation protocols in Appendices D.3--D.4.

\par\addvspace{3pt}\noindent\textbf{Transition architecture comparison.}
To assess the effectiveness of the operator-structured transition against alternative architectures, we compare LEON with \emph{Transformer} and \emph{Residual MLP} baselines within VLA-JEPA. The Transformer baseline retains the 12-block predictor architecture of VLA-JEPA, but reinitializes its transition blocks to avoid an advantage from pretrained predictor weights. The Residual MLP retains the observable map $\varphi$, condition-fusion function $g_{\mathrm{fuse}}$, and increment readout $W_\psi$ of LEON, while replacing the shared operator--complementary evolution with eight independently parameterized residual blocks, each comprising four affine layers. All variants preserve the VLA-JEPA prediction targets and policy interface, and are trained independently using the same data, training schedule, evaluation protocol, and pretrained initialization of the retained VLA-JEPA components. Detailed baseline architectures and experimental protocols are provided in Appendix E.2 and Appendices D.3--D.4, respectively.

On LIBERO, LEON outperforms the Transformer and Residual MLP baselines by 1.65 and 1.70 percentage points, respectively. On LIBERO-Plus, which evaluates policy robustness under controlled distribution shifts, LEON achieves higher average success rates by 3.91 and 4.01 points without additional adaptation. The improvements extend across all seven perturbation categories, with the largest gains under sensor noise. These results support the robustness of the complete LEON transition relative to the evaluated alternatives, while subsequent component ablations examine the contributions of its individual design choices.

\par\addvspace{3pt}\noindent\textbf{Observable adaptation.}
We examine the contribution of the learned observable representation to operator-structured evolution. The \emph{w/o Observable Adaptation} variant removes the learned observable map $\varphi$ and increment readout $W_\psi$, applying the operator and complementary updates directly to the projected features from the retained VLA-JEPA input mapping. This variant outperforms the Transformer and Residual MLP baselines on both LIBERO and LIBERO-Plus, supporting the effectiveness of operator-structured evolution in the original feature space. Incorporating observable adaptation further increases the average SR from 98.05\% to 99.05\% on LIBERO and from 80.94\% to 83.45\% on LIBERO-Plus. The larger improvement on LIBERO-Plus suggests that observable adaptation contributes to performance under the evaluated perturbations, consistent with the dependence of finite-dimensional operator approximations on the chosen observable representation discussed in Section~\ref{sec:leon_physical_dynamics}. Detailed variant definitions are provided in Appendix E.2.


\par\addvspace{3pt}\noindent\textbf{Operator modulation and update components.}
We examine the contributions of condition-dependent operator modulation and the two update components. The \emph{w/o Operator Modulation} variant replaces the dynamically inferred coefficients in $A(\xi_i)$ with a learned constant $\bar\xi=\tanh(\bar q)$, yielding $\bar A=D+\sum_{k=1}^{M}\bar\xi_kB_k$. The resulting operator is independent of the observation and action context, while the complementary update $b(h_i,\xi_i)$ retains its original conditioning. On LIBERO-Plus, this variant reduces the average SR from 83.45\% to 79.93\%, supporting the contribution of conditional operator modulation to policy robustness. To examine the update components, \emph{w/o Additive} removes $b(h_i,\xi_i)$, whereas \emph{w/o Operator} removes the complete operator contribution $A(\xi_i)h_i$, including the diagonal term, while retaining the original conditioned additive branch. Their LIBERO-Plus averages decrease to 81.41\% and 80.32\%, respectively. All three variants underperform LEON on average LIBERO success and across all seven LIBERO-Plus perturbation categories. These results support the complementary contributions of operator propagation and additive updates, together with the benefit of condition-dependent operator modulation. Detailed variant definitions are provided in Appendix E.2.

\subsection{Structured Evolution Diagnostics}
\label{sec:operator_diagnostics}

\begin{figure}[t]
\centering
\LEONfigure{\linewidth}{42mm}{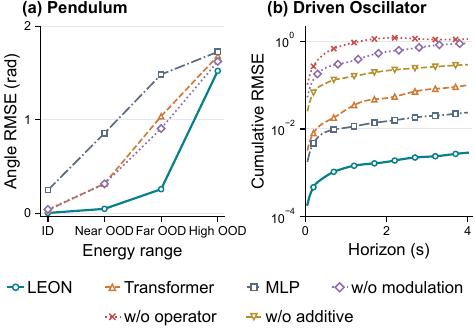}
\caption{\textbf{Controlled diagnostics of structured prediction.}
(a) Energy extrapolation: mean 8-s wrapped-angle RMSE on the damped pendulum across initial-energy conditions.
(b) Parameter extrapolation: mean cumulative state RMSE across prediction horizons on a driven oscillator with stiffness beyond the training range.}
\label{fig:controlled-tests}
\vspace{-3pt}
\end{figure}

We use physical systems with known governing dynamics to examine how LEON models state evolution through shared operators and complementary updates, independently of policy learning. Transformer and Residual MLP baselines use the same training trajectories and number of optimization steps, with comparable numbers of trainable parameters. Results are averaged over five training seeds, with detailed protocols in Appendix F.1. In the damped-pendulum experiment, we evaluate nonlinear prediction with initial energies beyond the training range. LEON achieves the lowest mean cumulative 8-s wrapped-angle RMSE across all evaluation conditions (Fig.~\ref{fig:controlled-tests}a), reducing error relative to the Transformer by 84.7\% and 75.0\% under Near-OOD and Far-OOD, respectively, while maintaining the lowest wrapped-angle RMSE among all compared methods under High-OOD.

In the driven-oscillator experiment, stiffness is fixed within each trajectory, while a piecewise-constant external force drives the system. We evaluate stiffness extrapolation by training on $\kappa\in[1,4]$ and testing on $\kappa\in[4.5,6]$. Checkpoint selection uses both in-distribution validation trajectories
and a validation set with stiffness $\kappa \in [4.05, 4.4]$. Training supervises eight successive prediction steps, whereas evaluation extends to 80 steps (4~s). LEON achieves lower OOD rollout RMSE than Transformer and Residual MLP baselines throughout the prediction horizon (Fig.~\ref{fig:controlled-tests}b). Disabling either the operator term or the complementary additive term
at test time substantially increases the full-horizon error, with a
larger increase when the operator term is disabled. Additional results
are reported in Appendices~F.3 and~F.4. 

%% file: sections/06_conclusion.tex
\section{Conclusion}

Our results highlight dynamics-guided latent-transition structure as a useful inductive bias for world modeling. LEON improves average manipulation success and perturbation robustness across two distinct prediction--policy interfaces, while preserving their prediction targets. Ablations and controlled experiments further support the effectiveness of shared operator evolution and complementary updates.

%% file: llm_usage.tex
\section*{LLM Usage Statement}
ChatGPT was used solely for language editing and polishing of the manuscript, including improvements to grammar, wording, clarity, and readability. It was not used to develop the theoretical framework, formulate mathematical claims, design the methodology or experiments, implement methods, analyze data, interpret results, or generate scientific conclusions. All AI-assisted edits were reviewed and verified by the authors. The authors take full responsibility for the final content of the manuscript.

%% file: appendix_setup.tex
\numberwithin{equation}{section}
\numberwithin{table}{section}
\setlength{\parindent}{0pt}
\setlength{\parskip}{4pt plus 1pt minus 1pt}
\setlength{\emergencystretch}{1.5em}
\setlength{\textfloatsep}{12pt plus 2pt minus 2pt}
\setlength{\floatsep}{10pt plus 2pt minus 2pt}
\renewcommand{\arraystretch}{1.13}
\raggedbottom
\pagestyle{plain}

%% file: appendices/appendix_a_theory.tex
\section{Additional Derivations and Theoretical Foundations}
\label{app:leon-theory}

LEON structures latent transitions using shared operators and complementary updates, motivated by the observable dynamics of controlled mechanical systems. Starting from the actuated equations of motion, this appendix derives a finite-dimensional evolution law consisting of a shared matrix family affine in physical control and an explicit projection residual. We establish a sufficient representation of this family using dense matrices, derive its first-order discretization in normalized time, and relate the projected dynamics to LEON's shared conditional vector field. A conditional error estimate separates approximation of the vector field from time discretization. Controlled spring and pendulum examples illustrate exact finite closure and nonzero projection residuals, respectively. Appendix~\ref{app:leon-specification} specifies the single observable construction, shared field evaluations, and increment readout used to implement this formulation.

\subsection{From Mechanical Dynamics to Shared Generators}
\label{app:leon-generators}

\paragraph{Controlled mechanical dynamics.}
Consider an actuated mechanical system whose equations of motion, within a smooth interaction regime, are
\begin{equation}
 M(q)\ddot q+c(q,\dot q)=B_{\mathrm{act}}(q)u,
 \label{eq:leon-a-mechanics}
\end{equation}
where $q\in\R^{n_q}$ denotes generalized coordinates, $u\in\R^m$ denotes physical control inputs, $M(q)$ is a nonsingular inertia matrix, $c$ collects the non-actuation terms, and $B_{\mathrm{act}}$ maps control inputs to generalized forces \citep{spong2020robot}. With $v=\dot q$ and $s=(q,v)$, Eq.~\eqref{eq:leon-a-mechanics} becomes
\begin{equation}
 \begin{aligned}
 \dot s&=f_0(s)+\sum_{j=1}^{m}u_jf_j(s),\\
 f_0(s)&=\begin{bmatrix}v\\-M(q)^{-1}c(q,v)\end{bmatrix},
 &f_j(s)&=\begin{bmatrix}0\\M(q)^{-1}B_{\mathrm{act}}(q)e_j\end{bmatrix}.
 \end{aligned}
 \label{eq:leon-a-affine}
\end{equation}
Here $e_j$ is the $j$th coordinate vector. The drift vector field $f_0$ and input vector fields $f_j$ do not depend on the control value. Let $\Omega\subset\R^{2n_q}$ be an open state region on which these vector fields are continuously differentiable. The derivations below apply on intervals where the trajectory remains in this region; impact events and changes of contact mode are not included in this smooth description.

\paragraph{Koopman operators and infinitesimal generators.}
For a fixed control value $u$, let $\varphi_u^t$ denote the local flow. The associated Koopman operator is defined on scalar observables by
\begin{equation}
 (\mathcal K_u^t g)(s)=g(\varphi_u^t(s)),
 \label{eq:leon-a-koopman}
\end{equation}
whenever the flow is defined. For $g\in C^1(\Omega)$, its infinitesimal generator satisfies
\begin{equation}
 \begin{aligned}
 (\mathcal L_u g)(s)
 &=\left.\frac{\mathrm d}{\mathrm dt}g(\varphi_u^t(s))\right|_{t=0}\\
 &=Dg(s)f_0(s)+\sum_{j=1}^{m}u_jDg(s)f_j(s)
 =\mathcal L_0g(s)+\sum_{j=1}^{m}u_j\mathcal L_jg(s),
 \end{aligned}
 \label{eq:leon-a-generator}
\end{equation}
where $\mathcal L_jg=Dg\,f_j$ for $j=0,\ldots,m$. Each generator is linear in the observable function, and the generator family is affine in the control parameter \citep{proctor2018koopmancontrol,peitz2020interpolated,otto2024bilinear}.

Let $\Phi=(\phi_1,\ldots,\phi_d)^\top\in C^1(\Omega;\R^d)$ and $h=\Phi(s)$. For a continuous time-dependent control $u(t)$, the chain rule gives
\begin{equation}
 \dot h(t)=(\mathcal L_{u(t)}\Phi)(s(t)),
 \label{eq:leon-a-observable-derivative}
\end{equation}
with the generator applied componentwise. Equation~\eqref{eq:leon-a-generator} is first derived for a fixed control parameter and then evaluated at $u(t)$. Linearity of the generator on observable functions does not imply that a finite collection of observables satisfies closed linear dynamics.

\subsection{Finite Projection and Dense Operator Families}
\label{app:leon-projection}

\paragraph{Common finite-dimensional projection.}
Fix a probability measure $\mu$ on $\Omega$ and use it for every control value. Assume that $\phi_i$ and $\mathcal L_j\phi_i$ belong to $L^2(\mu)$ for all $i$ and $j$, and that $\{\phi_i\}_{i=1}^d$ are linearly independent in $L^2(\mu)$. Denote by $P_\Phi$ the orthogonal projection onto $V_\Phi=\spn\{\phi_1,\ldots,\phi_d\}$, applied componentwise to vector-valued observables. Define
\begin{equation}
 K_\Phi=\E_\mu[\Phi\Phi^\top],\qquad
 H_j=\E_\mu[(\mathcal L_j\Phi)\Phi^\top],\qquad j=0,\ldots,m.
 \label{eq:leon-a-projection}
\end{equation}
For every nonzero $a\in\R^d$,
\begin{equation}
 a^\top K_\Phi a=\E_\mu[(a^\top\Phi)^2]>0,
 \label{eq:leon-a-gram-positive}
\end{equation}
because $a^\top\Phi$ cannot vanish in $L^2(\mu)$ under the independence assumption. Hence $K_\Phi$ is positive definite.

The projected generator matrix is the unique minimizer
\begin{equation}
 G_j=\underset{C\in\R^{d\times d}}{\arg\min}\ J_j(C),\qquad
 J_j(C)=\E_\mu\bigl[\|\mathcal L_j\Phi-C\Phi\|_2^2\bigr].
 \label{eq:leon-a-least-squares}
\end{equation}
Expanding this quadratic objective gives
\begin{equation}
 \begin{aligned}
 J_j(C)&=\E_\mu\bigl[\|\mathcal L_j\Phi\|_2^2\bigr]
 -2\operatorname{tr}(CH_j^\top)+\operatorname{tr}(CK_\Phi C^\top),\\
 \nabla_CJ_j(C)&=2(CK_\Phi-H_j).
 \end{aligned}
 \label{eq:leon-a-normal-equations}
\end{equation}
The normal equations therefore yield $G_jK_\Phi=H_j$ and $G_j=H_jK_\Phi^{-1}$. Equivalently, $P_\Phi\mathcal L_j\Phi=G_j\Phi$ in $L^2(\mu)$. For the residual defined pointwise by $r_j(s)=(\mathcal L_j\Phi)(s)-G_j\Phi(s)$,
\begin{equation}
 \E_\mu[r_j\Phi^\top]=H_j-G_jK_\Phi=0.
 \label{eq:leon-a-orthogonality}
\end{equation}
This is the least-squares projection of the generator applied to the chosen observables \citep{klus2020data}. Invariance of $\mu$ under the system dynamics is not required.

For each fixed control parameter, linearity of the common projection gives
\begin{equation}
 P_\Phi\mathcal L_u\Phi
 =P_\Phi\mathcal L_0\Phi+\sum_{j=1}^{m}u_jP_\Phi\mathcal L_j\Phi
 =\left(G_0+\sum_{j=1}^{m}u_jG_j\right)\Phi.
 \label{eq:leon-a-common-projection}
\end{equation}
All generator components are projected using the same measure and observable subspace. Substitution into Eq.~\eqref{eq:leon-a-observable-derivative} gives the exact decomposition
\begin{equation}
 \begin{aligned}
 \dot h&=G(u)h+r(s,u),\\
 G(u)&=G_0+\sum_{j=1}^{m}u_jG_j,\qquad
 r(s,u)=r_0(s)+\sum_{j=1}^{m}u_jr_j(s).
 \end{aligned}
 \label{eq:leon-a-decomposition}
\end{equation}
The residuals vanish in $L^2(\mu)$ when $V_\Phi$ is invariant under every generator component. Exact closure at every state in a region requires the corresponding pointwise residuals to vanish there. Otherwise, the residual retains dependence on the physical state and need not be determined by $h$ alone.

\paragraph{Dimension of the matrix family.}
Let $r_\star=\dim\spn\{G_1,\ldots,G_m\}\le m$. Choosing a basis $\{\widetilde G_k\}_{k=1}^{r_\star}$ gives a matrix $T_{\mathrm{ctrl}}\in\R^{r_\star\times m}$ such that
\begin{equation}
 G(u)=G_0+\sum_{k=1}^{r_\star}(T_{\mathrm{ctrl}}u)_k\widetilde G_k.
 \label{eq:leon-a-basis}
\end{equation}
The quantity $r_\star$ is the dimension of a subspace of matrices, not a bound on the rank of its elements. LEON parameterizes an analogous family by
\begin{equation}
 A(\xi)=D+\sum_{k=1}^{M}\xi_kB_k,\qquad
 B_k\in\R^{d_h\times d_h},
 \label{eq:leon-a-dense}
\end{equation}
where $D$ is diagonal and the dense matrices $B_k$ have no imposed rank constraint. Thus $\dim\spn\{B_k\}_{k=1}^M\le M$, while $\rank(A(\xi)-D)$ can equal $d_h$ even when $M=1$.

The following result gives a sufficient representation condition for fixed projected matrices. Including $G_0-D$ accounts for drift components that cannot be represented by the diagonal base.

\begin{proposition}[Dense representation with bounded coefficients]
\label{prop:leon-a-dense}
Let $\mathcal U\subset\R^m$ be nonempty and compact, and let $G(u)=G_0+\sum_{j=1}^{m}u_jG_j$, where $G_j\in\R^{d\times d}$. Fix a diagonal matrix $D$ and define
\begin{equation}
 r_D=\dim\spn\{G_0-D,G_1,\ldots,G_m\}.
 \label{eq:leon-a-drift-space}
\end{equation}
If $M\ge r_D$, then for every $0<\rho<1$ there exist matrices $B_1,\ldots,B_M\in\R^{d\times d}$ and an affine coefficient map $\xi^\star:\mathcal U\to(-\rho,\rho)^M$ satisfying
\begin{equation}
 G(u)=D+\sum_{k=1}^{M}\xi_k^\star(u)B_k
 \qquad\text{for all }u\in\mathcal U.
 \label{eq:leon-a-exact-dense}
\end{equation}
\end{proposition}

\begin{proof}
If $r_D=0$, then $G_0=D$ and $G_j=0$ for $j\ge1$, so all $B_k$ and coefficients may be set to zero. Otherwise, choose a basis $\{C_k\}_{k=1}^{r_D}$ of the space in Eq.~\eqref{eq:leon-a-drift-space}. Each $G_j$ has fixed coordinates in this basis, so there are affine scalar functions $a_k(u)$ such that $G(u)-D=\sum_{k=1}^{r_D}a_k(u)C_k$. Compactness of $\mathcal U$ implies $L_k=\sup_{u\in\mathcal U}|a_k(u)|<\infty$. Choose $s_k>L_k/\rho$ and set $B_k=s_kC_k$ and $\xi_k^\star(u)=a_k(u)/s_k$. Then $|\xi_k^\star(u)|<\rho$, and substitution gives Eq.~\eqref{eq:leon-a-exact-dense}. Any unused matrices and coefficients are set to zero.\qedhere
\end{proof}

\paragraph{Scope of the representation result.}
By definition, $r_D\le m+1$. Proposition~\ref{prop:leon-a-dense} concerns the representation of fixed projected matrices; it does not establish finite observable closure or recovery of $\xi^\star(u)$ from visual context. The diagonal parameterization, number of basis matrices, and coefficient network are architectural choices rather than consequences of Koopman theory. The result applies to any fixed diagonal matrix allowed by the chosen implementation.

Unless otherwise indicated, vector norms are Euclidean and matrix norms are the associated induced norms. If $\|\xi\|_\infty\le\rho$, then
\begin{equation}
 \|A(\xi)\|\le\|D\|+\rho\sum_{k=1}^{M}\|B_k\|.
 \label{eq:leon-a-operator-norm}
\end{equation}
The coefficient bound alone therefore gives no uniform operator-norm bound independent of the basis matrices. In particular, the rescaling used in the proof is a representation argument, not a stability guarantee for the learned dynamics or their Euler discretization.

\subsection{Discrete Evolution, Normalized Time, and the LEON Formulation}
\label{app:leon-time}

\paragraph{Euler discretization in physical time.}
Suppose that $\dot h$ is Lipschitz continuous in time with constant $L_t$ on $[t,t+\Delta t]$. Integration of Eq.~\eqref{eq:leon-a-decomposition} gives
\begin{equation}
 h(t+\Delta t)
 =h(t)+\Delta t\,[G(u(t))h(t)+r(s(t),u(t))]+e_t.
 \label{eq:leon-a-euler}
\end{equation}
The remainder satisfies
\begin{equation}
 \begin{aligned}
 e_t&=\int_0^{\Delta t}\bigl[\dot h(t+\sigma)-\dot h(t)\bigr]\,\mathrm d\sigma,\\
 \|e_t\|&\le\int_0^{\Delta t}L_t\sigma\,\mathrm d\sigma
 =\tfrac12L_t(\Delta t)^2.
 \end{aligned}
 \label{eq:leon-a-euler-remainder}
\end{equation}
Neglecting $e_t$ yields a forward Euler update. The regularity condition is local to the stated interval and is not assumed across discontinuities in the control or mechanical dynamics. Affine dependence of the infinitesimal generator on control does not imply affine dependence of the exact finite-time transition, which generally involves higher-order compositions of the system dynamics.

\paragraph{Normalized time.}
For a reference duration $\Tref>0$, define $\tau=(t-t_0)/\Tref$, $\bar s(\tau)=s(t_0+\Tref\tau)$, $\bar u(\tau)=u(t_0+\Tref\tau)$, and $\bar h(\tau)=\Phi(\bar s(\tau))$. The chain rule gives
\begin{equation}
 \frac{\mathrm d\bar h}{\mathrm d\tau}
 =\Tref\bigl[G(\bar u)\bar h+r(\bar s,\bar u)\bigr]
 =\bar G(\bar u)\bar h+\bar r(\bar s,\bar u),
 \label{eq:leon-a-normalized}
\end{equation}
where $\bar G=\Tref G$ and $\bar r=\Tref r$. An interval of physical duration $\Tref$ corresponds to $\tau\in[0,1]$. For $S$ equal updates, the normalized step is $\eta=1/S$. Since the factor $\Tref$ is included in the normalized vector field, it is not multiplied into $\eta$ a second time. If $\dot h$ has physical-time Lipschitz constant $L_t$ on this interval, then $\bar h'$ has normalized-time Lipschitz constant at most $\Tref^2L_t$.

\paragraph{From projected dynamics to a learned vector field.}
For a fixed set of network parameters, let $X=E(R(s))$ denote the retained visual features, and let $h=\phi(X)$ be the learned observable representation. The theoretical construction applies to the composite map $\Phi=\phi\circ E\circ R$ provided that it satisfies the differentiability, integrability, and linear-independence assumptions stated above. Injectivity of the observation map is not required; unresolved state dependence is retained in the projection residual.

Let $c$ denote the conditioning representation supplied by the WAM and held fixed over a prediction interval. LEON approximates normalized observable evolution using the conditional vector field
\begin{equation}
 \begin{aligned}
 \xi&=\xi_\theta(h,c),\\
 F_\theta(h,c)&=A_\theta(\xi)h+b_\theta(h,\xi),\qquad
 A_\theta(\xi)=D+\sum_{k=1}^{M}\xi_kB_k.
 \end{aligned}
 \label{eq:leon-a-learned-field}
\end{equation}
The shared matrix family corresponds to the projected generator structure, while the learned complementary term permits changes not captured by that family. The latent coefficients are inferred from the available observation and action-related context; they are not assumed to identify physical control. Because they depend on $h$, the complete vector field is generally nonlinear in the observable state. The affine parameterization of $A_\theta$ in $\xi$ does not assert that the projection in Eq.~\eqref{eq:leon-a-common-projection} commutes with multiplication by state-dependent coefficients.

Using a single initial map $h_0=\phi(X)$ keeps all subsequent updates in one observable coordinate system. For fixed $c$, the same parameterized vector field is evaluated at each current state:
\begin{equation}
 h_{n+1}=h_n+\eta F_\theta(h_n,c),\qquad
 n=0,\ldots,S-1,\qquad \eta=1/S.
 \label{eq:leon-a-shared-update}
\end{equation}
The matrices and networks are shared across these evaluations, but $\xi_\theta(h_n,c)$ and the resulting effective matrix may vary with $n$. A field without an explicit time input is a modeling choice for the prediction interval, not a consequence of control-affine mechanics alone. LEON uses $S=8$ and reads the accumulated increment back to the predictor space as specified in Appendix~\ref{app:leon-specification}. Neither this update count nor the increment readout is uniquely determined by the physical derivation.

\paragraph{Sources of vector-field approximation error.}
Fix an observable representation and a set of model parameters, with $d=d_h$. Choose a reference coefficient map $\xi^\star:\mathcal U\to\R^M$ and define $A^\star(u)=D+\sum_k\xi_k^\star(u)B_k$ using the same $D$ and $B_k$ as the model. At a physical state $s$, control $u$, and context $c$, let $h=\Phi(s)$ and $\widehat\xi=\xi_\theta(h,c)$. The difference between the normalized observable derivative and the learned field is
\begin{equation}
 \begin{aligned}
 d_\theta(s,u,c)
 &:=\bar G(u)h+\bar r(s,u)-F_\theta(h,c)\\
 &=\bigl[\bar G(u)-A^\star(u)\bigr]h
 +\sum_{k=1}^{M}\bigl[\xi_k^\star(u)-\widehat\xi_k\bigr]B_kh
 +\bar r(s,u)-b_\theta(h,\widehat\xi).
 \end{aligned}
 \label{eq:leon-a-error-decomposition}
\end{equation}
Define $\varepsilon_A(u)=\|\bar G(u)-A^\star(u)\|$, $\varepsilon_\xi(h,u,c)=\|\xi^\star(u)-\widehat\xi\|_\infty$, and $\varepsilon_b(s,u,c)=\|\bar r(s,u)-b_\theta(h,\widehat\xi)\|$. The triangle inequality yields
\begin{equation}
 \|d_\theta(s,u,c)\|
 \le\left(\varepsilon_A(u)
 +\varepsilon_\xi(h,u,c)\sum_{k=1}^{M}\|B_k\|\right)\|h\|
 +\varepsilon_b(s,u,c).
 \label{eq:leon-a-field-error-bound}
\end{equation}
This separates matrix-family approximation, coefficient inference, and complementary-term approximation for a fixed observable representation. Equation~\eqref{eq:leon-a-error-decomposition} does not require exact representation by $A^\star$, and the two learned branches may compensate for one another. In particular, predictive training does not impose $b_\theta=\bar r$ or establish numerical bounds on these error terms.

\paragraph{Propagation through shared updates.}
The following conditional estimate makes the distinction between vector-field approximation and time discretization explicit. It is stated for a fixed context and does not assume closed physical dynamics in $h$.

\begin{proposition}[Finite-horizon error under regularity assumptions]
\label{prop:leon-a-error-propagation}
Let $\bar h:[0,1]\to\R^{d_h}$ be a continuously differentiable reference trajectory, with $\bar h'$ Lipschitz continuous with constant $C_\tau$. Fix $c$ and suppose $F_\theta(\cdot,c)$ is $L$-Lipschitz on a set containing this trajectory and the iterates in Eq.~\eqref{eq:leon-a-shared-update}, where $L\ge0$. Assume
\begin{equation}
 \sup_{\tau\in[0,1]}
 \|\bar h'(\tau)-F_\theta(\bar h(\tau),c)\|\le\varepsilon.
 \label{eq:leon-a-defect-assumption}
\end{equation}
If $h_0=\bar h(0)$ and $\eta=1/S$, then
\begin{equation}
 \|h_S-\bar h(1)\|
 \le\left(\varepsilon+\frac{C_\tau}{2S}\right)\Gamma(L),\qquad
 \Gamma(L)=\begin{cases}(e^L-1)/L,&L>0,\\1,&L=0.\end{cases}
 \label{eq:leon-a-global-error}
\end{equation}
\end{proposition}

\begin{proof}
Let $e_n=\|h_n-\bar h(n\eta)\|$. Integrating the reference derivative over one step, as in Eq.~\eqref{eq:leon-a-euler-remainder}, gives a remainder with norm at most $C_\tau\eta^2/2$. Subtracting the numerical update and using Eq.~\eqref{eq:leon-a-defect-assumption} and the Lipschitz bound gives
\begin{equation}
 e_{n+1}\le(1+L\eta)e_n+\eta\varepsilon+\tfrac12C_\tau\eta^2.
 \label{eq:leon-a-error-recursion}
\end{equation}
Since $e_0=0$, iteration yields $e_S\le\eta(\varepsilon+C_\tau\eta/2)\sum_{j=0}^{S-1}(1+L\eta)^j$. For $L>0$, the sum gives
\begin{equation}
 e_S\le\left(\varepsilon+\tfrac12C_\tau\eta\right)
 \frac{(1+L\eta)^S-1}{L}
 \le\left(\varepsilon+\tfrac12C_\tau\eta\right)\frac{e^L-1}{L},
 \label{eq:leon-a-gronwall}
\end{equation}
because $S\eta=1$. For $L=0$, summing Eq.~\eqref{eq:leon-a-error-recursion} gives $e_S\le\varepsilon+C_\tau\eta/2$. Both cases establish Eq.~\eqref{eq:leon-a-global-error}.\qedhere
\end{proof}

The estimate isolates an $O(S^{-1})$ discretization term while retaining the vector-field approximation term $\varepsilon$. It does not supply values for $\varepsilon$, $L$, or $C_\tau$ for a trained model and therefore does not certify the accuracy or stability of the particular choice $S=8$. Observable learning, parameter optimization, and the final prediction-space readout require separate analysis or empirical evaluation. Likewise, the reference duration defines the standard prediction horizon; boundary-clipped training pairs require an explicit sampling protocol rather than an assumption that every pair has that measured duration.

\subsection{Illustrative Finite-Observable Mechanical Systems}
\label{app:leon-examples}

\paragraph{Controlled damped spring: exact finite closure.}
Consider
\begin{equation}
 \dot q=v,\qquad m_s\dot v=-k_sq-c_sv+u,
 \qquad m_s>0,\quad k_s>0,\quad c_s\ge0.
 \label{eq:leon-a-spring-dynamics}
\end{equation}
With $\Phi(s)=(1,q,v)^\top$, the observable dynamics satisfy
\begin{equation}
 \dot h=(G_0+uG_1)h,\qquad
 G_0=\begin{bmatrix}
 0&0&0\\
 0&0&1\\
 0&-k_s/m_s&-c_s/m_s
 \end{bmatrix},\qquad
 G_1=\begin{bmatrix}
 0&0&0\\
 0&0&0\\
 1/m_s&0&0
 \end{bmatrix}.
 \label{eq:leon-a-spring}
\end{equation}
The constant observable allows multiplication by the control-dependent matrix to represent the additive input $u/m_s$. The selected observable space is invariant under both generator components, so the projection residual is identically zero. The same matrices describe trajectories with different initial states and control values for fixed mechanical parameters.

\paragraph{Controlled damped pendulum: nonzero projection residual.}
On a local angular coordinate chart, let
\begin{equation}
 \dot\theta=\omega,\qquad
 \dot\omega=-\omega_p^2\sin\theta-\beta_p\omega+u/I_p,
 \qquad I_p>0,\quad\omega_p>0,\quad\beta_p\ge0.
 \label{eq:leon-a-pendulum}
\end{equation}
Choose $\Phi=(1,\theta,\omega)^\top$. Let $\mu$ have bounded support within this coordinate chart and a density that is strictly positive on a nonempty open rectangle in $(\theta,\omega)$. These conditions satisfy the assumptions of Section~\ref{app:leon-projection}. The least-squares projection of $\sin\theta$ onto $V_\Phi$ is $a^\top\Phi$, with
\begin{equation}
 a^\top=\E_\mu[\sin\theta\,\Phi^\top]K_\Phi^{-1},\qquad
 a=(a_0,a_1,a_2)^\top.
 \label{eq:leon-a-pendulum-projection}
\end{equation}
The projected matrices and residuals are
\begin{equation}
 \begin{aligned}
 G_0&=\begin{bmatrix}
 0&0&0\\
 0&0&1\\
 -\omega_p^2a_0&-\omega_p^2a_1&-\beta_p-\omega_p^2a_2
 \end{bmatrix},
 &G_1&=\begin{bmatrix}0&0&0\\0&0&0\\1/I_p&0&0\end{bmatrix},\\[3pt]
 r_0(s)&=\begin{bmatrix}0\\0\\-\omega_p^2(\sin\theta-a^\top\Phi(s))\end{bmatrix},
 &r_1(s)&=0.
 \end{aligned}
 \label{eq:leon-a-pendulum-residual}
\end{equation}
Substitution recovers the original equations exactly through Eq.~\eqref{eq:leon-a-decomposition}. The control component is represented without a residual. In contrast, $\sin\theta$ cannot equal an affine function of $(\theta,\omega)$ throughout the stated rectangle, so $r_0$ is nonzero in $L^2(\mu)$. Its allocation between $G_0\Phi$ and $r_0$ depends on $\mu$. The coordinate $\theta$ is defined on the chosen chart and is not a globally smooth real-valued coordinate on the circle. These examples demonstrate that a shared matrix representation is compatible both with exact finite closure and with unresolved observable dynamics, motivating joint learning of the observables and their operator and complementary updates.

%% file: appendices/appendix_b_specification.tex
\section{LEON Architecture and Implementation}
\label{app:leon-specification}

Appendix~\ref{app:leon-theory} derives the projected observable dynamics that motivate LEON's shared matrix family and complementary term. This appendix specifies the network architecture that implements this formulation at the WAM predictor interface. The input features are mapped once to observable coordinates, evolved over a unit normalized interval through eight evaluations of a shared conditional vector field, and updated by a linear readout of the accumulated observable increment. We define the constituent modules and their dimensions, distinguish the conditioning mechanisms and parameterizations used in VLA-JEPA \citep{sun2026vlajepa} and LaWAM \citep{chen2026lawam}, and provide the complete forward algorithm. Initialization and numerical implementation are specified to make the correspondence between the theoretical formulation and the implemented transition explicit.

\subsection{Notation and Shared Hyperparameters}
\label{app:leon-notation}

As in the main text, $X=E(o)=[x_1,\ldots,x_N]^\top\in\R^{N\times d_x}$ denotes the visual encoder output. The target $X^+$ and prediction $\widehat X^+$ share this feature space, with $d_x=d_{\mathrm{vis}}=2048$ for VLA-JEPA and $768$ for LaWAM.

Write $X_{\mathrm{in}}=\mathcal P_{\mathrm{WAM}}(X)=[x_{1,\mathrm{in}},\ldots,x_{N,\mathrm{in}}]^\top\in\R^{N\times d_{\mathrm{pred}}}$, where $\mathcal P_{\mathrm{WAM}}$ comprises the retained input projection and any positional encoding, and $d_{\mathrm{pred}}=1024$. The main-text observable map is $\varphi=\phi_{\mathrm{core}}\circ\mathcal P_{\mathrm{WAM}}$, with $\phi_{\mathrm{core}}$ the observable module defined below. Token vectors are written as columns.

The observable state at internal step $n$ is $\mathbf{h}^{(n)}=[\mathbf{h}_1^{(n)},\ldots,\mathbf{h}_N^{(n)}]^\top\in\R^{N\times d_h}$. Parenthesized superscripts $(n)$ index internal Euler steps; subscripts $i$ and $k$ index visual tokens and basis matrices or coefficient components, respectively. Action tokens are indexed by $j$. Batch and prediction-pair indices are omitted. Let $Z$ denote the action-related representation supplied by the WAM and $c=\mathcal{E}_Z(Z)$ its encoded condition, held fixed throughout the prediction interval. The map $\mathcal{E}_Z$ is the retained WAM action encoder; the additional conditioning modules inside LEON are specified below.

With $c$ fixed over the prediction interval, LEON constructs the observable representation once and approximates its normalized dynamics by $S$ forward Euler steps of a shared conditional vector field. The accumulated observable increment is mapped back to the predictor space through a single linear readout:
\begin{equation}
 \begin{aligned}
  \mathbf{h}^{(0)}&=\varphi(X),\\
  \mathbf{h}^{(n+1)}&=\mathbf{h}^{(n)}+\eta F_\theta(\mathbf{h}^{(n)},c),
  \qquad n=0,\ldots,S-1,\\
  \widehat X^+&=\mathcal R_{\mathrm{WAM}}
  \bigl(X_{\mathrm{in}}+(\Delta\mathbf{h}^{(S)})W_\psi^\top\bigr).
 \end{aligned}
 \label{eq:leon-b-overall}
\end{equation}
Here $\eta=T/S$, with normalized horizon $T=1$ and $S=8$, and $\Delta\mathbf{h}^{(n)}:=\mathbf{h}^{(n)}-\mathbf{h}^{(0)}$ denotes the accumulated observable increment after $n$ steps. The observable module $\phi_{\mathrm{core}}$ and field $F_\theta$ act tokenwise, and the same field parameters are used at every step. The retained WAM output map $\mathcal R_{\mathrm{WAM}}$ is applied after the increment readout.

The number of basis matrices is denoted by $M$, the number of Euler steps by $S$, and the number of action tokens, when applicable, by $K$. In particular, $M$ does not denote matrix rank. Table~\ref{tab:leon-b-common} lists the shared hyperparameters. The two WAM instantiations are trained separately; parameter sharing refers to repeated evaluations within each instantiation.

\begin{table}[htbp]
 \centering
 \caption{Shared architectural and discretization hyperparameters of LEON. All durations and step sizes in the evolution equations are expressed in normalized time.}
 \label{tab:leon-b-common}
 \small
 \begin{tabularx}{\linewidth}{@{}>{\raggedright\arraybackslash}Xcc@{}}
  \toprule
  Hyperparameter & Symbol & Value \\
  \midrule
  Predictor input dimension & $d_{\mathrm{pred}}$ & $1024$ \\
  Observable dimension & $d_h$ & $768$ \\
  Number of basis matrices & $M$ & $96$ \\
  Number of Euler steps & $S$ & $8$ \\
  Normalized prediction horizon & $T$ & $1$ \\
  Euler step size & $\eta=T/S$ & $0.125$ \\
  LayerNorm numerical constant & $\epsilon_{\mathrm{LN}}$ & $10^{-5}$ \\
  Dropout probability in the conditional field & $p_{\mathrm{drop}}$ & $0$ \\
  \bottomrule
 \end{tabularx}
\end{table}

For the layer specifications, a two-layer MLP with dimensions $a\!\to\!w\!\to\!b$ denotes the map $z\mapsto W_2\sigma(W_1\LN(z)+\beta_1)+\beta_2$, where $\sigma$ is the stated hidden activation. Thus, each such MLP includes an input LayerNorm and two affine layers, with no activation after the second affine layer unless explicitly specified. LayerNorm acts independently along the feature dimension of each input vector, with trainable scale and shift parameters and the numerical constant in Table~\ref{tab:leon-b-common}. Different named modules have independent parameters. Repeated evaluations of a given module use the same parameters. All scalar activation functions act componentwise.

\subsection{Observable Construction and Increment Readout}
\label{app:leon-lift-readout}

\paragraph{Observable construction.}
For $\bar x_i=\LN_{\mathrm{in}}(x_{i,\mathrm{in}})$, the initial observable is
\begin{equation}
 \mathbf{h}_i^{(0)}=\mathbf{h}_i=\phi_{\mathrm{core}}(x_{i,\mathrm{in}})=W_\phi\bar x_i+\beta_\phi+g_\phi(\bar x_i),
 \label{eq:leon-b-observable}
\end{equation}
where $W_\phi\in\R^{d_h\times d_{\mathrm{pred}}}$ and $g_\phi$ is the nonlinear residual MLP specified in Table~\ref{tab:leon-b-layers}. The input LayerNorm of $g_\phi$ is distinct from $\LN_{\mathrm{in}}$. Both instantiations use this construction, including the GELU hidden activation in $g_\phi$. The observable module is evaluated once before evolution. With $E$ denoting the visual encoder, $\Phi=\varphi\circ E\circ R=\phi_{\mathrm{core}}\circ\mathcal P_{\mathrm{WAM}}\circ E\circ R$, giving $\mathbf h^{(0)}=\varphi(X)=\Phi(s)$ as in the main text and Appendix~\ref{app:leon-theory}.

\paragraph{Increment readout.}
Let $\Delta\mathbf{h}_i^{(0)}=0$ and write $\mathbf{h}_i^{(n)}=\mathbf{h}_i^{(0)}+\Delta\mathbf{h}_i^{(n)}$, with $\Delta\mathbf{h}^{(n)}=[\Delta\mathbf{h}_1^{(n)},\ldots,\Delta\mathbf{h}_N^{(n)}]^\top$. The main-text residual readout uses predictor-interface coordinates and leaves the retained adapters implicit. We denote these coordinates by $X_{\mathrm{in}}$ and $\widetilde X_{\mathrm{in}}^+$. The implemented readout is
\begin{equation}
 \begin{aligned}
  \widetilde X_{\mathrm{in}}^+&=X_{\mathrm{in}}+(\Delta\mathbf{h}^{(S)})W_\psi^\top,\\
  \widehat X^+&=\mathcal R_{\mathrm{WAM}}(\widetilde X_{\mathrm{in}}^+),
 \end{aligned}
 \label{eq:leon-b-readout}
\end{equation}
where $W_\psi\in\R^{d_{\mathrm{pred}}\times d_h}$ is a trainable linear readout and $\mathcal R_{\mathrm{WAM}}$ is the retained WAM output map. The readout has no bias or normalization layer. It is applied once to the accumulated increment, which is maintained directly during evolution. Consequently, $\Delta\mathbf{h}^{(S)}=0$ gives $\widetilde X_{\mathrm{in}}^+=X_{\mathrm{in}}$, independently of the compression from $d_{\mathrm{pred}}$ to $d_h$. This identity holds at the LEON predictor interface; the complete prediction additionally includes $\mathcal R_{\mathrm{WAM}}$.

\subsection{Shared Conditional Vector Field}
\label{app:leon-field}

For the current observable $\mathbf{h}_i^{(n)}$ and fixed external condition $c$, define the fused representation $\rho_i^{(n)}=g_{\mathrm{fuse}}(\mathbf{h}_i^{(n)},c)\in\R^{d_h}$. The coefficient MLP produces $q_i^{(n)}=f_\xi(\rho_i^{(n)})$, and both instantiations use the componentwise transformation $\xi_i^{(n)}=\tanh(q_i^{(n)})\in(-1,1)^M$. LEON evaluates the conditional vector field
\begin{equation}
 F_\theta(\mathbf{h}_i^{(n)},c)
 =\left(D+\sum_{k=1}^{M}\xi_{i,k}^{(n)}B_k\right)\mathbf{h}_i^{(n)}
  +g_b\bigl(\mathbf{h}_i^{(n)}+g_p(\xi_i^{(n)})\bigr),
 \label{eq:leon-b-field}
\end{equation}
where $D$ is diagonal and each $B_k\in\R^{d_h\times d_h}$ is learned directly as an unrestricted matrix. The tensor storing these matrices has shape $M\times d_h\times d_h$, with entries indexed as $B[k,\mathrm{out},\mathrm{in}]$. The coefficient count $M$ controls the dimension of their linear span rather than the ranks of its elements. We write $A(\xi)=D+\sum_k\xi_kB_k$ for the matrix family and $b(\mathbf{h},\xi)=g_b(\mathbf{h}+g_p(\xi))$ for the complementary term, consistently with Appendix~\ref{app:leon-theory}. Both terms use the same coefficient vector. Although $A$ is affine in $\xi$, the complete field is generally nonlinear in the state and condition.

\begin{table}[htbp]
 \centering
 \caption{Layer specifications of the common LEON modules. Dimensions list input, hidden, and output widths for MLPs and input and output widths for linear maps. MLP normalization follows the convention in Section~\ref{app:leon-notation}. The coefficient MLP is followed by the componentwise $\tanh$ activation defined in Section~\ref{app:leon-field}.}
 \label{tab:leon-b-layers}
 \small
 \begin{tabularx}{\linewidth}{@{}p{0.29\linewidth}p{0.30\linewidth}>{\raggedright\arraybackslash}Xc@{}}
  \toprule
  Module & Layer dimensions & Hidden activation & Bias \\
  \midrule
  Observable projection $W_\phi$ & $1024\!\to\!768$ & --- & Yes \\
  Observable residual $g_\phi$ & $1024\!\to\!1024\!\to\!768$ & GELU & Yes \\
  Coefficient MLP $f_\xi$ & $768\!\to\!768\!\to\!96$ & SiLU & Yes \\
  Coefficient embedding $g_p$ & $96\!\to\!256\!\to\!768$ & GELU & Yes \\
  Complementary MLP $g_b$ & $768\!\to\!768\!\to\!768$ & Model-specific & Yes \\
  Increment readout $W_\psi$ & $768\!\to\!1024$ & --- & No \\
  \bottomrule
 \end{tabularx}
\end{table}

\paragraph{VLA-JEPA conditioning.}
The retained action encoder maps $K=8$ action tokens of dimension $2048$ to encoded tokens $c_j\in\R^{1024}$. A learned affine projection followed by LayerNorm produces $\bar c_j\in\R^{768}$; let $\bar C=[\bar c_1,\ldots,\bar c_K]^\top$. The global summary is $c^{\mathrm{glob}}=g_{\mathrm{glob}}(K^{-1}\sum_{j=1}^{K}\bar c_j)$. For each visual token, a local summary is obtained by multi-head attention, $c_i^{\mathrm{loc},(n)}=\operatorname{MHA}(\LN_q(\mathbf{h}_i^{(n)}),\bar C,\bar C)$, with eight heads of dimension $96$ and total width $768$. Attention uses scaled dot products with scale $96^{-1/2}$ and softmax over the $K$ action tokens. The query, key, value, and output projections are affine $768\!\to\!768$ maps with biases; no attention mask is applied within a prediction pair, and attention dropout is zero. The fused representation is $\rho_i^{(n)}=g_{\mathrm{merge}}([c^{\mathrm{glob}};c_i^{\mathrm{loc},(n)}])$, where the bracket denotes concatenation in the stated order. The merge map applies LayerNorm to the concatenated vector and then an affine projection. Queries depend on the current observable; keys and values are derived only from the action tokens associated with that prediction pair.

\paragraph{LaWAM conditioning.}
The retained action projection maps one aggregated latent action of dimension $32$ to $c\in\R^{1024}$. The global summary is $c^{\mathrm{glob}}=g_{\mathrm{glob}}(c)$, and the fused representation is $\rho_i^{(n)}=g_{\mathrm{merge}}([\mathbf{h}_i^{(n)};c^{\mathrm{glob}}])$. Both maps are normalized two-layer MLPs with the dimensions and activations in Table~\ref{tab:leon-b-differences}. The global summary is computed once before the Euler updates in this implementation. In both models, state-independent conditioning representations have fixed values over the prediction interval, whereas state-dependent fusion is evaluated at each current observable. Neither conditional field contains attention between visual tokens; visual context is supplied by the retained feature encoders.

\begin{table}[htbp]
 \centering
 \caption{Model-specific LEON configurations. All MLPs use the normalization convention in Section~\ref{app:leon-notation}.}
 \label{tab:leon-b-differences}
 \small
 \begin{tabularx}{\linewidth}{@{}>{\raggedright\arraybackslash}p{0.26\linewidth}>{\raggedright\arraybackslash}X>{\raggedright\arraybackslash}X@{}}
  \toprule
  Component or setting & VLA-JEPA + LEON & LaWAM + LEON \\
  \midrule
  Action encoder $\mathcal{E}_Z$ & Affine $2048\!\to\!1024$ for each of $8$ tokens & Affine $32\!\to\!1024$ for one latent action \\
  Action-token projection & Affine $1024\!\to\!768$ with bias, then LayerNorm & Not used \\
  Global-summary MLP $g_{\mathrm{glob}}$ & $768\!\to\!768\!\to\!768$; GELU & $1024\!\to\!768\!\to\!768$; SiLU \\
  Context merge $g_{\mathrm{merge}}$ & LayerNorm, then affine $1536\!\to\!768$ & MLP $1536\!\to\!768\!\to\!768$; SiLU \\
  Local action attention & $8$ heads; head dimension $96$ & None \\
  Attention projections & Affine $768\!\to\!768$ for query, key, value, and output; all with biases & Not used \\
  Hidden activation of $g_b$ & $\tanh$ & SiLU \\
  Diagonal parameterization $D$ & $\diag(a-\mathbf 1)$ & $\diag(0.1\tanh\widetilde d)$ \\
  Reference prediction interval $\Delta t_{\mathrm{pred}}$ on LIBERO (s) & $0.1$ & $0.35$ \\
  Reference prediction interval $\Delta t_{\mathrm{pred}}$ on RoboTwin (s) & --- & $35/30$ \\
  \bottomrule
 \end{tabularx}
\end{table}

\paragraph{Diagonal parameterization.}
The VLA-JEPA diagonal is parameterized by $a\in\R^{d_h}$ with initial value $\mathbf 1$; LaWAM uses $\widetilde d\in\R^{d_h}$ with initial value zero. Both give an initial effective diagonal $D=0$. The identity contribution to each Euler update arises from retaining the current state, separately from $D$ in Eq.~\eqref{eq:leon-b-field}.

For the VLA-JEPA instantiation, training additionally penalizes $\max(|q_{i,k}^{(S-1)}|-2,0)^2$, averaged over prediction pairs, visual tokens, and coefficient components, with weight $10^{-3}$; this uses the final field evaluation, as defined in Section~\ref{app:leon-objectives}. The basis matrices $\{B_k\}$ use a learning-rate multiplier of $0.01$ relative to the reference LEON learning rate, with zero weight decay for both the basis and coefficient-network parameter groups. These settings apply to the VLA-JEPA optimization protocol and are distinct from the shared discretization hyperparameters in Table~\ref{tab:leon-b-common}.

\subsection{Normalized Evolution and Forward Algorithm}
\label{app:leon-algorithm}

For normalized horizon $T=1$ and $S=8$, the step size is $\eta=T/S=1/8$. The accumulated increment evolves according to
\begin{equation}
 \begin{aligned}
  \mathbf{h}_i^{(n)}&=\mathbf{h}_i^{(0)}+\Delta\mathbf{h}_i^{(n)},\\
  \Delta\mathbf{h}_i^{(n+1)}&=\Delta\mathbf{h}_i^{(n)}+\eta F_\theta(\mathbf{h}_i^{(n)},c),
  \qquad n=0,\ldots,S-1,
 \end{aligned}
 \label{eq:leon-b-update}
\end{equation}
starting from $\Delta\mathbf{h}_i^{(0)}=0$. This is algebraically equivalent to the Euler update for $\mathbf{h}_i^{(n)}$ in Appendix~\ref{app:leon-time}. A single set of parameters defines $D$, $\{B_k\}$, the conditioning modules, $f_\xi$, $g_p$, and $g_b$ across all steps. The coefficients are recomputed from the current state, so the effective matrix $A(\xi_i^{(n)})$ may vary with $n$. The field has no explicit time input, step embedding, or learned step size. No projection or normalization is applied to the evolving state between updates; the normalizations within the field modules remain part of $F_\theta$.

\paragraph{Differentiation.}
Training differentiates through all $S$ updates without detaching the evolving state or accumulated increment between steps. Gradients for shared parameters accumulate over their uses across the updates. The fixed condition $c$ retains its gradient path through $\mathcal{E}_Z$, and activation checkpointing recomputes field evaluations during backpropagation.

Algorithm~\ref{alg:leon-b-forward} summarizes the complete forward computation. The token loop describes independent field evaluations conditional on the common external representation; implementations evaluate these operations in batches.

\begingroup
\floatstyle{ruled}
\restylefloat{algorithm}
\algrenewcommand{\algorithmicrequire}{\textbf{Input:}}
\algrenewcommand{\algorithmicensure}{\textbf{Output:}}
\begin{algorithm}[htbp]
 \caption{LEON forward computation with a shared conditional vector field}
 \label{alg:leon-b-forward}
 \begin{algorithmic}[1]
  \Require Visual features $X$; action representation $Z$; horizon $T=1$; steps $S=8$
  \Ensure Future-feature prediction $\widehat X^+$
  \State $X_{\mathrm{in}}\gets\mathcal P_{\mathrm{WAM}}(X)$
  \State $c\gets\mathcal{E}_Z(Z)$; $\eta\gets T/S$
  \State $\mathbf{h}^{(0)}\gets\phi_{\mathrm{core}}(X_{\mathrm{in}})$ \Comment{evaluates $\varphi(X)$ once}
  \State $\Delta\mathbf{h}^{(0)}\gets0$
  \For{$n=0,\ldots,S-1$}
   \State $\mathbf{h}^{(n)}\gets \mathbf{h}^{(0)}+\Delta\mathbf{h}^{(n)}$
   \For{$i=1,\ldots,N$}
    \State $\rho_i^{(n)}\gets g_{\mathrm{fuse}}(\mathbf{h}_i^{(n)},c)$
    \State $\xi_i^{(n)}\gets\tanh(f_\xi(\rho_i^{(n)}))$
    \State $v_i^{(n)}\gets A(\xi_i^{(n)})\mathbf{h}_i^{(n)}+b(\mathbf{h}_i^{(n)},\xi_i^{(n)})$
    \State $\Delta\mathbf{h}_i^{(n+1)}\gets\Delta\mathbf{h}_i^{(n)}+\eta v_i^{(n)}$
   \EndFor
  \EndFor
  \State $\widetilde X_{\mathrm{in}}^+\gets X_{\mathrm{in}}+(\Delta\mathbf{h}^{(S)})W_\psi^\top$ \Comment{one increment readout}
  \State \Return $\mathcal R_{\mathrm{WAM}}(\widetilde X_{\mathrm{in}}^+)$
 \end{algorithmic}
\end{algorithm}
\endgroup

\paragraph{Prediction horizon.}
The benchmark-specific reference prediction intervals $\Delta t_{\mathrm{pred}}$ in Table~\ref{tab:leon-b-differences}, denoted by $\Tref$ in Appendix~\ref{app:leon-time}, define the change of variables $\tau=(t-t_0)/\Delta t_{\mathrm{pred}}$ described in Appendix~\ref{app:leon-time}. The reference duration is absorbed into the normalized vector field, and both implementations use $\eta=1/8$. The fixed-horizon configurations retain prediction pairs affected by trajectory boundaries, so the unit model horizon does not assert an identical measured time gap for every pair. For a different number of integration steps at the same model horizon, the step size is $T/S$.

\paragraph{Retained WAM output maps.}
In VLA-JEPA, $\mathcal R_{\mathrm{WAM}}$ applies LayerNorm in dimension $1024$, with numerical constant $10^{-6}$, followed by an affine projection to dimension $2048$. In LaWAM, it applies an affine projection from dimension $1024$ to $768$, followed by the terminal feature LayerNorm. These endpoint maps are applied after the increment readout. They lie outside the shared conditional field.

\subsection{Implementation Details}
\label{app:leon-initialization}

\paragraph{Initialization.}
Sample a standard Gaussian matrix in $\R^{1024\times768}$ using a local random generator with seed $1729$, and obtain its reduced QR factor $Q$, where $Q^\top Q=I$. Initialize $W_\phi=Q^\top$, $\beta_\phi=0$, and $W_\psi=Q$. The two projection matrices are subsequently optimized independently; the transpose relation and orthogonality specify their initialization. The local generator fixes the initial projection and is separate from the training seed.

Initialize the weights and biases of the final affine layers of $g_\phi$, $f_\xi$, and $g_b$ to zero, and initialize the diagonal parameters as specified above. Each basis entry is sampled independently from a zero-mean Gaussian with standard deviation $10^{-4}$. Thus the basis matrices are initially nonzero, but $\xi=0$, $b=0$, and $D=0$ imply $F_\theta=0$. Consequently, the initial model satisfies $\Delta\mathbf{h}^{(S)}=0$ and $\widetilde X_{\mathrm{in}}^+=X_{\mathrm{in}}$ at the LEON interface. Table~\ref{tab:leon-b-initialization} specifies the remaining LEON parameters. The attention output projection follows the affine-layer initialization rule.

\begin{table}[htbp]
 \centering
 \caption{Initialization of LEON parameters not fixed by the structured assignments above. Here $d_{\mathrm{in}}$ is the input width of an affine layer. Normal truncation bounds are expressed in weight units.}
 \label{tab:leon-b-initialization}
 \small
 \begin{tabularx}{\linewidth}{@{}p{0.29\linewidth}>{\raggedright\arraybackslash}X>{\raggedright\arraybackslash}X@{}}
  \toprule
  Parameter group & VLA-JEPA + LEON & LaWAM + LEON \\
  \midrule
  Remaining affine weights & $\mathcal N(0,0.02^2)$ truncated to $[-2,2]$ & $\mathcal U(-d_{\mathrm{in}}^{-1/2},d_{\mathrm{in}}^{-1/2})$ \\
  Remaining affine biases & Zero & $\mathcal U(-d_{\mathrm{in}}^{-1/2},d_{\mathrm{in}}^{-1/2})$ \\
  Packed attention input weight & Xavier uniform on a $3d_h\times d_h$ matrix & Not used \\
  Attention input bias & Zero & Not used \\
  LayerNorm scale and shift & One and zero, respectively & One and zero, respectively \\
  \bottomrule
 \end{tabularx}
\end{table}

\paragraph{Numerical precision.}
Table~\ref{tab:leon-b-precision} distinguishes parameter storage, computation, and accumulation within LEON. BF16 autocast applies to eligible operations; explicitly FP32 computations override this setting. These arithmetic choices accompany the same mathematical update in Eq.~\eqref{eq:leon-b-update}.

\begin{table}[htbp]
 \centering
 \caption{Numerical precision within LEON. BF16 autocast specifies a computation mode rather than a uniform dtype for every operation. The retained WAM adapters follow their original precision settings.}
 \label{tab:leon-b-precision}
 \small
 \begin{tabularx}{\linewidth}{@{}p{0.38\linewidth}>{\raggedright\arraybackslash}X>{\raggedright\arraybackslash}X@{}}
  \toprule
  Storage or computation & VLA-JEPA + LEON & LaWAM + LEON \\
  \midrule
  Parameters, states, and accumulated increments & FP32 & FP32 \\
  Eligible field operations & BF16 autocast & BF16 autocast \\
  Coefficient MLP and output transformation & FP32; autocast disabled & BF16 autocast \\
  Coefficient-weighted basis reduction & FP32 & FP32 \\
  Euler accumulation & FP32 & FP32 \\
  Increment readout and residual addition & FP32; autocast disabled & FP32; autocast disabled \\
  \bottomrule
 \end{tabularx}
\end{table}

\paragraph{Matrix contraction and computational cost.}
The matrix-dependent contribution is evaluated as $\sum_k\xi_k(B_k\mathbf{h})$, without constructing a separate effective matrix for each token. The implementation partitions tokens and basis matrices into blocks and recomputes intermediate contractions during the backward pass. Blocking controls intermediate memory use while preserving the matrix parameterization. The basis tensor contains $Md_h^2=96\times768^2=56{,}623{,}104$ trainable parameters, and the diagonal contributes another $d_h$ parameters. These parameters are stored once and shared across all $S$ evaluations. The basis contraction requires $O(Md_h^2)$ operations per token per step, giving $O(SNMd_h^2)$ over one prediction. The linear observable projection and increment readout each require $O(Nd_{\mathrm{pred}}d_h)$ operations; the residual observable MLP and conditional-field networks contribute their respective additional costs. Parameter sharing reduces field parameter storage relative to $S$ independently parameterized fields, while retaining $S$ field evaluations.

\par\addvspace{4pt}\noindent\textbf{Limitations.}
LEON derives its transition structure from controlled observable dynamics, while the individual learned operator components are not explicitly identified with physical generators. Further understanding the precise correspondence between latent operator evolution and underlying physical dynamics remains a direction for future work.

%% file: appendices/appendix_c_integration.tex
\section{Integration with Existing WAMs}
\label{app:leon-integration}

LEON implements the transition module of a WAM through the observable construction, shared Euler updates, and increment readout specified in Appendix~\ref{app:leon-specification}. This appendix describes its integration with VLA-JEPA~\citep{sun2026vlajepa} and LaWAM~\citep{chen2026lawam}, from the construction of prediction pairs and conditioning representations to the retained feature interfaces and parameter initialization. The two implementations are presented in parallel, followed by the visual sampling convention, training objectives, and gradient paths that determine how the replacement transition is optimized and used for action generation.

Following the main text and Appendix~\ref{app:leon-notation}, $X=E(o)$ denotes the source visual features, and $X^+$ and $\widehat X^+$ denote the future target and prediction. These features lie in $\R^{N\times d_x}$, with $d_x=d_{\mathrm{vis}}$. The retained input mapping produces $X_{\mathrm{in}}=\mathcal P_{\mathrm{WAM}}(X)\in\R^{N\times d_{\mathrm{pred}}}$, with $d_{\mathrm{pred}}=1024$. The encoded action-related condition is $c=\mathcal{E}_Z(Z)$. Within one prediction pair, the observable state $\mathbf{h}^{(n)}$, accumulated increment $\Delta\mathbf{h}^{(n)}$, and Euler step index $n$ follow Appendix~\ref{app:leon-notation}. The index $r$, introduced below, denotes a position in the encoded video sequence and is distinct from the internal Euler step.

\subsection{Integration with VLA-JEPA}
\label{app:leon-vlajepa-interface}

VLA-JEPA~\citep{sun2026vlajepa} learns latent-action representations through future-feature prediction and jointly trains a flow-matching action head on robot demonstrations. Given the current observation and instruction, its VLM produces latent-action tokens and embodied-action tokens. The world model predicts future visual features from encoded visual history and the latent-action tokens, using features extracted by a frozen visual encoder as supervision. Within the VLM, the embodied-action tokens attend to the predictively trained latent-action tokens, incorporating transition information into the representations that condition action generation. At inference, the action head generates action chunks conditioned on the embodied-action tokens, without evaluating the future-feature prediction branch. We replace the transition module of the world model with LEON while retaining the feature interfaces and action-generation pathway.

\paragraph{Conditioning representations.}
For each prediction pair, $Z$ contains $K=8$ latent-action tokens of dimension $2048$. The retained action encoder $\mathcal{E}_Z$ applies an affine map to each token, producing $c\in\R^{K\times1024}$. Each token group is associated with one transition in the encoded video sequence. Its value is determined by the initial visual observation, task instruction, and the VLM token context. LEON holds $c$ fixed over the prediction interval and evaluates the state-dependent fusion specified in Appendix~\ref{app:leon-field}. The embodied-action tokens condition the action head through the retained VLM attention structure.

\paragraph{Prediction pairs.}
The retained eight-frame video input is encoded into four temporal feature groups. Features from the two camera views are concatenated along the channel dimension within each group, giving $d_{\mathrm{vis}}=2048$. For consecutive groups $r$ and $r+1$, with $r=0,1,2$, the source features $X$ are taken from group $r$, the target $X^+$ from group $r+1$, and $Z$ from the corresponding latent-action token group. All three prediction pairs are supervised. Each pair uses its encoded source features rather than the preceding prediction as input. The pairs are evaluated in parallel, and each pair undergoes $S=8$ internal Euler updates.

\paragraph{Transition replacement.}
The affine projection $P_{\mathrm{in}}$ maps $X$ from dimension $2048$ to $d_{\mathrm{pred}}=1024$, producing $X_{\mathrm{in}}$. LEON replaces the original predictor Transformer computation between the retained input and output adapters. It constructs $\mathbf{h}^{(0)}=\varphi(X)=\phi_{\mathrm{core}}(X_{\mathrm{in}})$ once, evolves the observables with one shared conditional field, and reads out the accumulated increment as in Eq.~\eqref{eq:leon-b-overall}. The retained output map $\mathcal R_{\mathrm{WAM}}$ applies LayerNorm in dimension $1024$, followed by an affine projection to dimension $2048$. Each prediction therefore has the spatial organization and feature dimension of its V-JEPA target. The replacement transition operates independently on the encoded prediction pairs; visual temporal context is supplied by the retained video encoder.

\paragraph{Parameter initialization and training.}
The retained VLM, action head, prediction-input projection, action encoder $\mathcal{E}_Z$, and terminal prediction adapters inherit their compatible pretrained WAM parameters. The original predictor Transformer parameters are replaced by the LEON parameters initialized according to Appendix~\ref{app:leon-initialization}. The V-JEPA~2 prediction encoder remains frozen and is evaluated without gradient tracking. LEON, the retained prediction adapters, the VLM, and the action head are optimized during policy fine-tuning. The frozen prediction encoder is separate from the trainable visual backbone within the VLM. Table~\ref{tab:leon-c-parameters} summarizes the optimization scope.

\paragraph{Training and inference.}
During policy training, the prediction loss supervises the latent-action pathway jointly with the action objective. Transition information reaches the embodied-action representations through their attention to the latent-action tokens and through the shared VLM parameters. At inference, the VLM still computes the latent-action and embodied-action representations, and the action head generates actions from the embodied-action representations and retained robot-state inputs. Future-feature prediction and the V-JEPA prediction encoder are not evaluated in this action-generation path.

\subsection{Integration with LaWAM}
\label{app:leon-lawam-interface}

LaWAM~\citep{chen2026lawam} first pretrains a latent action model on current--future feature pairs extracted by a frozen visual encoder. Its inverse-dynamics encoder infers a latent action from each feature pair, and its forward decoder predicts future features from the current features and inferred latent action. The encoder and decoder are jointly trained through future-feature prediction, latent-space regularization, and auxiliary supervision of end-effector motion. The pretrained decoder is retained as the latent world model, while the inverse-dynamics encoder provides latent-action targets for training a policy conditioned on the current observation and instruction. At inference, the world model predicts a future visual subgoal from the current features and policy-predicted latent action. The action expert then generates an action chunk conditioned on the current features, predicted subgoal, and VLM context. We replace the transition module of the world model with LEON while retaining the visual-subgoal interface to the action expert.

\paragraph{Conditioning representations.}
The retained VLM processes the current visual observation and task instruction. Its latent-action aggregation module produces $Z\in\R^{32}$, which the retained affine action encoder $\mathcal{E}_Z$ maps to $c\in\R^{1024}$. LEON combines the fixed condition with the evolving observable through the fusion maps in Appendix~\ref{app:leon-field}. The VLM context supplied directly to the action expert is denoted by $H_{\mathrm{VLM}}$. During training, the frozen inverse-dynamics encoder and its latent bottleneck provide a target for $Z$; the LEON transition is conditioned on the policy prediction $Z$.

\paragraph{Prediction pairs.}
Each training example supplies a current--future pair from the primary camera view. The retained visual sampler selects the current frame and a frame seven sampling intervals later, subject to the original trajectory-boundary clipping rule. The frozen DINOv3 encoder produces $X$ and $X^+$ in dimension $d_{\mathrm{vis}}=768$. The latent-action teacher is evaluated on the same visual pair. Each source--target pair defines one LEON prediction, with $S=8$ internal Euler updates under the policy-predicted latent action.

\paragraph{Transition replacement.}
LEON replaces the original decoder Transformer blocks while retaining the affine input projection $P_{\mathrm{in}}$, spatial positional encoding, action projection $\mathcal{E}_Z$, and terminal output map. The input projection maps $X$ from dimension $768$ to $d_{\mathrm{pred}}=1024$, and the spatial positional encoding is then applied to obtain $X_{\mathrm{in}}$. LEON performs the observable construction, shared evolution, and increment readout of Eq.~\eqref{eq:leon-b-overall}. The retained output map $\mathcal R_{\mathrm{WAM}}$ applies an affine projection from dimension $1024$ to $768$, followed by the terminal feature LayerNorm. The resulting $\widehat X^+$ preserves the feature dimension and spatial arrangement expected by the action expert.

\paragraph{Parameter initialization and training.}
The retained policy modules and decoder adapters inherit their pretrained LaWAM parameters, and the replacement LEON transition is initialized as specified in Appendix~\ref{app:leon-initialization}. Policy fine-tuning optimizes LEON, the retained decoder adapters, the latent-action aggregation module, the action expert, and the trainable VLM parameters. The VLM retains its first $16$ language Transformer blocks, which are trainable, together with its trainable visual backbone and visual merger. Its token embeddings and vocabulary output projection are frozen. The DINOv3 prediction encoder and latent-action teacher also remain frozen. The pretraining of the latent action model supplies the teacher and inherited parameters; the objectives used to optimize the LEON-integrated policy are specified in Section~\ref{app:leon-objectives}.

\paragraph{Training and inference.}
The action expert receives $(X,\widehat X^+,H_{\mathrm{VLM}})$. Throughout the present policy-training configuration, its future-feature input is the model prediction, detached when evaluating the action loss. Prediction and latent-action distillation retain their own gradient paths. At inference, the policy computes $Z$ from the current observation and instruction, LEON predicts $\widehat X^+$, and the action expert uses the predicted visual subgoal to generate actions. The future-prediction branch remains active, while the observed future and latent-action teacher serve only as training supervision.

Table~\ref{tab:leon-c-interfaces} compares the retained interfaces. Table~\ref{tab:leon-c-parameters} distinguishes newly initialized LEON modules from inherited parameter groups and states their optimization status.

\begin{table}[htbp]
 \centering
 \caption{Feature and conditioning interfaces of the two LEON integrations. Projection dimensions are reported in input--output order. LayerNorm placement follows each retained WAM.}
 \label{tab:leon-c-interfaces}
 \small
 \begin{tabularx}{\linewidth}{@{}p{0.27\linewidth}YY@{}}
  \toprule
  Interface specification & VLA-JEPA + LEON & LaWAM + LEON \\
  \midrule
  Prediction features & V-JEPA~2; two-view channel concatenation & DINOv3; primary view \\
  Visual feature dimension $d_{\mathrm{vis}}$ & $2048$ & $768$ \\
  Action-related representation $Z$ & $8$ tokens of dimension $2048$ per prediction pair & One latent vector of dimension $32$ \\
  Action encoder $\mathcal{E}_Z$ & Affine $2048\!\to\!1024$ per token & Affine $32\!\to\!1024$ \\
  Visual input projection $P_{\mathrm{in}}$ & Affine $2048\!\to\!1024$ & Affine $768\!\to\!1024$ \\
  Output map $\mathcal R_{\mathrm{WAM}}$ & LayerNorm, then affine $1024\!\to\!2048$ & Affine $1024\!\to\!768$, then LayerNorm \\
  Prediction at inference & Branch omitted from action generation & Predicted features condition the action expert \\
  \bottomrule
 \end{tabularx}
\end{table}

\begin{table}[htbp]
 \centering
 \caption{Parameter initialization and optimization in the two integrations. LEON modules are initialized according to Appendix~\ref{app:leon-initialization}; retained WAM modules inherit pretrained parameters, and the frozen visual prediction encoders retain their pretrained weights. \emph{Trainable} specifies the optimization scope during policy fine-tuning.}
 \label{tab:leon-c-parameters}
 \small
 \begin{tabularx}{\linewidth}{@{}p{0.29\linewidth}YY@{}}
  \toprule
  Parameter group & VLA-JEPA + LEON & LaWAM + LEON \\
  \midrule
  LEON transition & Newly initialized; trainable & Newly initialized; trainable \\
  Retained prediction adapters & Pretrained; trainable & Pretrained; trainable \\
  Visual prediction encoder & Pretrained; frozen & Pretrained; frozen \\
  VLM visual backbone & Pretrained; trainable & Pretrained; trainable, including visual merger \\
  VLM language Transformer & Pretrained; retained blocks trainable & Pretrained; first $16$ blocks retained and trainable \\
  VLM token embeddings & Pretrained; trainable & Pretrained; frozen \\
  Action head or expert & Pretrained; trainable & Pretrained; trainable \\
  Latent-action teacher & Not used & Pretrained; frozen, including latent bottleneck \\
  \bottomrule
 \end{tabularx}
\end{table}

The initialization specifications describe the insertion of LEON into a pretrained WAM. Continued fine-tuning restores the resulting LEON-integrated checkpoint and preserves the trainable parameter groups in Table~\ref{tab:leon-c-parameters}.

\subsection{Training Protocol}
\label{app:leon-objectives}

\paragraph{Visual sampling and normalized time.}
The reference physical prediction interval $\Delta t_{\mathrm{pred}}$ is determined by the temporal separation of the visual source and target specified by the sampling procedure. On LIBERO, the reference interval is $0.1\,\mathrm{s}$ between consecutive encoded temporal groups in VLA-JEPA and $0.35\,\mathrm{s}$ between the sampled source and future frames in LaWAM. On RoboTwin, LaWAM uses a reference visual prediction interval of $35/30\,\mathrm{s}$. These visual intervals define $\tau=(t-t_0)/\Delta t_{\mathrm{pred}}$ and a unit normalized horizon, with $S=8$ and $\eta=1/S$, as specified in Appendix~\ref{app:leon-algorithm}. The prediction pairs retain the original clipping or padding at trajectory boundaries. Such pairs use the same normalized model horizon and remain included in the stated losses, without an additional boundary mask or sample-dependent rescaling of the evolution interval.

\paragraph{Feature-prediction supervision.}
Both implementations supervise $\widehat X^+$ with the future features $X^+$ from the corresponding frozen prediction encoder. The losses are evaluated directly in the retained visual feature coordinates, including the original encoder normalization, with no additional unit-length normalization. Let $\mathbb E_{\mathcal D}$ denote expectation over training prediction pairs, approximated by the minibatch mean. The prediction objectives are
\begin{equation}
 \begin{aligned}
  \mathcal L_{\mathrm{WM}}^{\text{VLA-JEPA}}
  &=\mathbb E_{\mathcal D}\!\left[
    \frac{1}{N d_{\mathrm{vis}}}
    \sum_{i=1}^{N}\sum_{\ell=1}^{d_{\mathrm{vis}}}
    \left|\widehat X^+_{i\ell}-X^+_{i\ell}\right|
   \right],\\
  \mathcal L_{\mathrm{WM}}^{\text{LaWAM}}
  &=\mathbb E_{\mathcal D}\!\left[
    \frac{\|\widehat X^+-X^+\|_F^2}{N d_{\mathrm{vis}}}
   \right].
 \end{aligned}
 \label{eq:leon-c-prediction-losses}
\end{equation}
Here $\|\cdot\|_F$ is the Frobenius norm, $i$ indexes spatial tokens, and $\ell$ indexes feature coordinates. VLA-JEPA averages the absolute error over all retained prediction pairs, tokens, and feature coordinates; LaWAM uses the corresponding mean squared error.

\paragraph{Latent-action distillation.}
LaWAM supervises the policy latent action with $Z^{\mathrm{teach}}\in\R^{32}$ from the frozen latent action model evaluated on the same source--future visual pair. The target is the mean of the pretrained variational posterior, obtained by the teacher's deterministic inference procedure. It is compared directly with $Z$ in the inherited latent coordinate system, without an additional alignment map or unit-length normalization. Let $\operatorname{sg}$ denote stop-gradient, which preserves a value in the forward computation and has zero derivative in backpropagation. The distillation objective is
\begin{equation}
 \mathcal L_{\mathrm{distill}}
 =\mathbb E_{\mathcal D}\!\left[
   \frac{1}{32}
   \bigl\|Z-\operatorname{sg}(Z^{\mathrm{teach}})\bigr\|_2^2
  \right].
 \label{eq:leon-c-distillation}
\end{equation}
This objective updates the latent-action aggregation module and the trainable VLM parameters producing its inputs. The teacher and its latent bottleneck are evaluated without gradient tracking. VLA-JEPA learns its latent-action tokens through predictive supervision without a teacher-distillation objective.

\paragraph{Joint optimization.}
The following objectives apply to policy fine-tuning after the transition replacement. Let $\mathcal L_{\mathrm{act}}$ denote the retained flow-matching action objective of the corresponding WAM; its action representation and action-head formulation follow the original work~\citep{sun2026vlajepa,chen2026lawam}. The complete objectives are
\begin{equation}
 \begin{aligned}
  \mathcal L_{\mathrm{train}}^{\text{VLA-JEPA}}
  &=\mathcal L_{\mathrm{act}}
    +\beta\mathcal L_{\mathrm{WM}}^{\text{VLA-JEPA}}
    +\lambda_\xi\mathcal L_\xi,\\
  \mathcal L_{\mathrm{train}}^{\text{LaWAM}}
  &=\mathcal L_{\mathrm{act}}
    +\lambda_{\mathrm{WM}}\mathcal L_{\mathrm{WM}}^{\text{LaWAM}}
    +\lambda_{\mathrm{distill}}\mathcal L_{\mathrm{distill}}.
 \end{aligned}
 \label{eq:leon-c-training-objectives}
\end{equation}
The weights are $\beta=\lambda_{\mathrm{WM}}=\lambda_{\mathrm{distill}}=0.1$ and $\lambda_\xi=10^{-3}$. The VLA-JEPA regularizer $\mathcal L_\xi$ is the mean of $\max(|q_{i,k}^{(S-1)}|-2,0)^2$ over prediction pairs, spatial tokens $i$, and coefficient components $k$. Here $q_{i,k}^{(S-1)}$ is the coefficient preactivation before the $\tanh$ activation in the final field evaluation, consistently with Appendix~\ref{app:leon-field}. The LaWAM teacher retains the representation learned during latent action model pretraining, while its parameters remain fixed under Eq.~\eqref{eq:leon-c-training-objectives}.

\paragraph{Gradient propagation.}
The prediction loss updates LEON and the retained prediction adapters in both integrations. Through $c=\mathcal{E}_Z(Z)$, it also updates the latent-action pathway and the trainable VLM parameters producing $Z$. In VLA-JEPA, the prediction objective trains the latent-action representations through the transition module; the embodied-action representations incorporate their transition information through VLM attention. The action objective updates the action head and the VLM through the embodied-action representations, including the internal attention dependencies on the latent-action representations.

In LaWAM, the action expert receives $(X,\operatorname{sg}(\widehat X^+),H_{\mathrm{VLM}})$ during training. The detach operation blocks the action-loss gradient through the predicted visual subgoal, while the direct VLM-context input retains its gradient path. The prediction objective trains the future-prediction branch and its upstream latent-action pathway; distillation independently supervises the policy latent action.

Let $\theta_{\mathrm{pred}}$ collect parameters exclusive to the future-prediction branch, including LEON and the retained prediction adapters. Both integrations satisfy $\nabla_{\theta_{\mathrm{pred}}}\mathcal L_{\mathrm{act}}=0$, while the shared VLM parameters may receive gradients from multiple objectives. Frozen prediction encoders and the LaWAM teacher receive no gradients from these losses. Table~\ref{tab:leon-c-gradients} summarizes the gradient destinations.

\begin{table}[htbp]
 \centering
 \caption{Parameter groups receiving gradients from each training objective. The prediction branch contains LEON and the retained prediction adapters. The latent-action pathway includes the modules producing $Z$ and their trainable VLM dependencies. All entries refer only to trainable parameters; frozen prediction encoders and the LaWAM teacher are excluded.}
 \label{tab:leon-c-gradients}
 \small
 \begin{tabularx}{\linewidth}{@{}p{0.23\linewidth}YY@{}}
  \toprule
  Training objective & VLA-JEPA + LEON & LaWAM + LEON \\
  \midrule
  Action loss $\mathcal L_{\mathrm{act}}$ & Action head and VLM through embodied-action representations and their attention dependencies & Action expert and VLM through the direct context input \\
  Prediction loss $\mathcal L_{\mathrm{WM}}$ & Prediction branch and latent-action pathway & Prediction branch and latent-action pathway \\
  Distillation loss $\mathcal L_{\mathrm{distill}}$ & Not used & Latent-action aggregation and upstream VLM \\
  Coefficient regularizer $\mathcal L_\xi$ & Observable map, conditional field, input adapters, and latent-action pathway & Not used \\
  \bottomrule
 \end{tabularx}
\end{table}

%% file: appendices/appendix_d_protocols.tex
\section{Training and Evaluation Protocols}
\label{app:leon-protocols}

This appendix specifies the experimental protocols for training and evaluating LEON within VLA-JEPA and LaWAM. We describe the benchmark data and optimization settings, define the training budgets and checkpoint-selection rule, and establish the evaluation and statistical reporting procedures for LIBERO, LIBERO-Plus, and RoboTwin~2.0. The network parameterization follows Appendix~\ref{app:leon-specification}, while the prediction interfaces, trainable components, and training objectives follow Appendix~\ref{app:leon-integration}.

\subsection{Datasets}
\label{app:leon-data}

\paragraph{LIBERO.}
We jointly fine-tune on LIBERO-Spatial, LIBERO-Object, LIBERO-Goal, and LIBERO-10 (Long), assigning equal training weights to the four suites. This mixture is used for both LEON instantiations and all retrained VLA-JEPA transition baselines and structural ablations. Data preprocessing and action normalization follow the corresponding VLA-JEPA and LaWAM protocols~\citep{sun2026vlajepa,chen2026lawam}.

\paragraph{RoboTwin.}
We follow the LaWAM multi-task protocol on RoboTwin~2.0, using the 50-task mixture of $2{,}500$ clean-scene demonstrations and $25{,}000$ demonstrations collected under scene randomization~\citep{chen2026lawam}. Data preprocessing and action normalization follow the original LaWAM protocol. The reference visual prediction interval is $\Delta t_{\mathrm{pred}}=35/30\,\mathrm{s}$, with the LEON evolution parameterized over the unit normalized horizon $T=1$.

\subsection{Optimization Settings}
\label{app:leon-optimization}

All experiments use AdamW with $(\beta_1,\beta_2)=(0.9,0.95)$ and $\epsilon=10^{-8}$. The global gradient norm is clipped at $1.0$. Learning rates increase linearly from zero during warmup and subsequently follow cosine decay. Table~\ref{tab:leon-d-optimization} lists the peak learning rates; warmup lengths and training budgets are given in Table~\ref{tab:leon-d-budgets}. Unlisted trainable parameters use a base rate of $10^{-5}$ in VLA-JEPA and $10^{-4}$ in LaWAM.

Weight decay is $10^{-8}$, except for the VLA-JEPA coefficient network $f_\xi$ and basis matrices $\{B_k\}$, which use zero weight decay. On RoboTwin, the complete AdamW update of each $B_k$, including its decoupled weight-decay contribution, is multiplied by $0.8$; the table reports the nominal learning rate. The shared LEON module is evaluated explicitly in FP32 for RoboTwin. The remaining numerical and differentiation conventions are specified in Appendix~\ref{app:leon-specification}.

\begin{table}[htbp]
 \centering
 \caption{Peak learning rates by parameter group. The remaining prediction parameters comprise LEON and the retained WAM prediction adapters, excluding the coefficient network and basis matrices listed separately. The RoboTwin basis-matrix update additionally uses the $0.8$ multiplier specified in the text.}
 \label{tab:leon-d-optimization}
 \small
 \begin{tabularx}{\linewidth}{@{}Yccc@{}}
  \toprule
  Parameter group
  & \shortstack{VLA-JEPA + LEON\\LIBERO}
  & \shortstack{LaWAM + LEON\\LIBERO}
  & \shortstack{LaWAM + LEON\\RoboTwin} \\
  \midrule
  VLM & $3.3\times10^{-5}$ & $10^{-4}$ & $10^{-4}$ \\
  Action head/expert & $3.3\times10^{-4}$ & $10^{-4}$ & $10^{-4}$ \\
  Remaining prediction parameters & $10^{-5}$ & $10^{-4}$ & $3\times10^{-4}$ \\
  Coefficient network $f_\xi$ & $10^{-5}$ & $10^{-4}$ & $3\times10^{-4}$ \\
  Basis matrices $\{B_k\}$ & $10^{-7}$ & $10^{-4}$ & $3\times10^{-4}$ \\
  \bottomrule
 \end{tabularx}
\end{table}

\subsection{Training Schedule}
\label{app:leon-training-schedule}

The retained VLA-JEPA modules are initialized from the released VLA-JEPA pretraining checkpoint. For LaWAM, we use the released policy-integration pretraining checkpoint, including its pretrained latent action model. LEON modules are initialized as specified in Appendix~\ref{app:leon-initialization}. Table~\ref{tab:leon-d-budgets} defines the training budgets, including warmup, and global batch sizes. All update counts refer to optimizer updates.

Given the cost of repeating benchmark fine-tuning, we report the final checkpoint from one run per WAM configuration in Table~\ref{tab:leon-d-budgets}. Each independent repeat requires $40{,}000$ optimizer updates on LIBERO or $100{,}000$ on RoboTwin. The RoboTwin fine-tuning run reported for LaWAM~\citep{chen2026lawam} used approximately $1{,}280$ H100 GPU-hours, equivalent to about $320$ hours on four H100 GPUs assuming the same throughput per GPU. Prior VLA and WAM studies likewise report benchmark success rates over evaluation rollouts~\citep{kim2024openvla,kim2025openvlaoft,bu2025univla,sun2026vlajepa,bi2026motus,kim2026cosmospolicy,chen2026lawam}. The standard deviations in Table~\ref{tab:leon-e-libero-seeds} describe variation across rollout seeds for fixed checkpoints. Variability across independent fine-tuning runs was not estimated.

The Transformer and Residual MLP transition baselines use the same LIBERO data mixture, pretrained initialization of non-transition components, optimization schedule, global batch size, training budget, and checkpoint-selection rule as VLA-JEPA + LEON. The Transformer transition module is randomly initialized before benchmark fine-tuning. The four structural ablations follow the same training protocol. Parameter-group rules apply to the parameters present in each variant, and every retrained VLA-JEPA variant uses its own final $40{,}000$-update checkpoint.

\begin{table}[htbp]
 \centering
 \caption{Training budgets and global batch sizes for LEON, retrained transition baselines, and structural ablations. Training updates include warmup. Every model is evaluated using the final checkpoint at its prescribed budget.}
 \label{tab:leon-d-budgets}
 \small
 \begin{tabularx}{\linewidth}{@{}Ylrrr@{}}
  \toprule
  Model & Benchmark & Training updates & Warmup updates & Global batch size \\
  \midrule
  VLA-JEPA + LEON & LIBERO & $40{,}000$ & $6{,}000$ & $256$ \\
  LaWAM + LEON & LIBERO & $40{,}000$ & $1{,}500$ & $256$ \\
  LaWAM + LEON & RoboTwin & $100{,}000$ & $2{,}000$ & $1{,}024$ \\
  Transformer transition baseline & LIBERO & $40{,}000$ & $6{,}000$ & $256$ \\
  Residual MLP transition baseline & LIBERO & $40{,}000$ & $6{,}000$ & $256$ \\
  VLA-JEPA structural ablations & LIBERO & $40{,}000$ & $6{,}000$ & $256$ \\
  \bottomrule
 \end{tabularx}
\end{table}

\subsection{Evaluation Protocol}
\label{app:leon-evaluation}

We report task success rates (SRs) as percentages using the benchmark success criteria. Observation processing, action normalization, and policy inference follow the corresponding original WAM protocols~\citep{sun2026vlajepa,chen2026lawam}. Table~\ref{tab:leon-d-evaluation} summarizes the evaluation settings and checkpoints for LEON, the retrained transition baselines, and the structural ablations.

\paragraph{LIBERO.}
Each evaluation seed runs $50$ trials per task across the four ten-task suites. Within a seed, each suite score averages its tasks equally, and the overall score averages the four suites equally. For the main benchmark results of the two LEON instantiations and the released WAM references, each checkpoint is held fixed across rollout seeds $\{0,1,2\}$. We report the mean and sample standard deviation across the three seed-level scores. These statistics describe rollout variability for fixed model weights. The VLA-JEPA transition comparisons and structural ablations use rollout seed $0$ only. The full LEON reference row uses the seed-$0$ result from the same checkpoint evaluated in the main benchmark.

\paragraph{LIBERO-Plus.}
We follow the original VLA-JEPA evaluation protocol. Each evaluated VLA-JEPA variant, including LEON, the retrained transition baselines, and the structural ablations, reuses its own final LIBERO checkpoint without additional training or adaptation. Evaluation uses rollout seed $0$. The overall score is the unweighted mean of the seven perturbation-family success rates.

\paragraph{RoboTwin.}
We follow the LaWAM evaluation protocol on all $50$ tasks, with $100$ trials per task under each of the Clean and Randomized conditions~\citep{chen2026lawam}. The same final LEON checkpoint is used for both conditions. Each condition score averages the task success rates equally; Clean and Randomized scores are reported separately, and the overall score is their arithmetic mean.

\begin{table}[htbp]
 \centering
 \caption{Evaluation protocols for LEON, retrained transition baselines, and structural ablations. Main LIBERO results use three rollout seeds for each fixed checkpoint. Transition comparisons and structural ablations use rollout seed $0$ only. On LIBERO-Plus, each evaluated VLA-JEPA variant reuses its own final LIBERO checkpoint. RoboTwin follows the original LaWAM protocol under both evaluation conditions.}
 \label{tab:leon-d-evaluation}
 \small
 \begin{tabularx}{\linewidth}{@{}p{0.18\linewidth}YY@{}}
  \toprule
  Benchmark & Evaluation protocol & Checkpoint \\
  \midrule
  LIBERO & Main benchmark results: $40$ tasks; $50$ trials per task and seed; rollout seeds $\{0,1,2\}$ & Final $40{,}000$-update checkpoint of each LEON instantiation \\
  LIBERO & Transition comparisons and structural ablations: $40$ tasks; $50$ trials per task; rollout seed $0$ & Final $40{,}000$-update checkpoint of each VLA-JEPA variant \\
  LIBERO-Plus & Original VLA-JEPA protocol; rollout seed $0$ & Final LIBERO checkpoint of each evaluated VLA-JEPA variant \\
  RoboTwin~2.0 & $50$ tasks; $100$ trials per task under both Clean and Randomized conditions & LaWAM + LEON, final $100{,}000$-update checkpoint \\
  \bottomrule
 \end{tabularx}
\end{table}

\paragraph{Reference baselines.}
On LIBERO, we re-evaluate the released VLA-JEPA and LaWAM checkpoints under the same evaluation protocol and with the same rollout seeds $\{0,1,2\}$ as their respective LEON counterparts. In LaWAM + LEON, the retained policy modules and decoder adapters inherit their pretrained LaWAM weights. To allow additional optimization of the randomly initialized LEON transition, we extend LIBERO fine-tuning from the $25{,}000$ updates in the published LaWAM protocol~\citep{chen2026lawam} to $40{,}000$ updates, keeping the global batch size at $256$. On RoboTwin, we use the published LaWAM results as the reference~\citep{chen2026lawam}. Both LaWAM configurations use $100{,}000$ fine-tuning updates, a global batch size of $1{,}024$, and $100$ evaluation trials per task under each condition. The independently retrained transition comparisons and structural ablations in Section~\ref{app:leon-transition-comparisons} use VLA-JEPA.

%% file: appendices/appendix_e_results_ablations.tex
\section{Supplementary Results and Ablations}
\label{app:leon-supplementary-results}

This appendix supplements the main evaluation with LIBERO results for individual rollout seeds and definitions of the transition comparisons. We report the suite-level performance of the released WAM references and both LEON instantiations, describe the RoboTwin comparison, and specify the retrained transition baselines and component ablations. Benchmark training, checkpoint selection, evaluation, and aggregation follow Appendix~\ref{app:leon-protocols}. Section~\ref{app:leon-closed-loop-protocol} specifies the closed-loop evaluation protocol and stage-level results for Figure~4.

\subsection{Detailed Benchmark Results}
\label{app:leon-detailed-results}

\paragraph{LIBERO.}
Table~\ref{tab:leon-e-libero-seeds} reports evaluations with rollout seeds $\{0,1,2\}$. The VLA-JEPA and LaWAM reference rows correspond to reevaluations of their officially released weights, as specified in Section~\ref{app:leon-evaluation}. Each reference checkpoint and each final LEON checkpoint is held fixed across the three seeds. Overall scores are computed within each seed by averaging the four suites equally; means and sample standard deviations are then computed across seeds.

The released VLA-JEPA and LaWAM checkpoints obtain overall success rates of $(97.88\pm0.38)\%$ and $(98.08\pm0.25)\%$, respectively. VLA-JEPA + LEON obtains $(98.95\pm0.17)\%$, and LaWAM + LEON obtains $(98.75\pm0.18)\%$. The corresponding increases in mean overall success rate are $1.07$ and $0.67$ percentage points relative to the respective released-checkpoint references. These statistics characterize rollout variability for fixed model weights. The separately retrained transition comparisons are specified in Section~\ref{app:leon-transition-comparisons}.

\begin{table}[htbp]
 \centering
 \caption{LIBERO success rates (\%). Each checkpoint is evaluated with rollout seeds $0$, $1$, and $2$. Mean SR and sample SD are computed across seeds; SD is expressed in percentage points (pp). Overall scores are averaged within each seed before aggregation. $\dagger$ denotes reevaluation of officially released WAM weights. LEON rows use the final $40{,}000$-update checkpoints.}
 \label{tab:leon-e-libero-seeds}
 \small
 \setlength{\tabcolsep}{4pt}
 \begin{tabularx}{0.96\linewidth}{@{}Y*{5}{>{\centering\arraybackslash}X}@{}}
  \toprule
  Task suite & Seed 0 & Seed 1 & Seed 2 & Mean SR (\%) & Sample SD (pp) \\
  \midrule
  \multicolumn{6}{@{}l}{\textit{VLA-JEPA}$^\dagger$} \\
  Spatial & $98.40$ & $97.40$ & $98.20$ & $98.00$ & $0.53$ \\
  Object & $99.80$ & $99.60$ & $99.80$ & $99.73$ & $0.12$ \\
  Goal & $98.60$ & $98.00$ & $98.60$ & $98.40$ & $0.35$ \\
  LIBERO-10 & $95.80$ & $94.80$ & $95.60$ & $95.40$ & $0.53$ \\
  \addlinespace[2pt]
  Overall & $98.15$ & $97.45$ & $98.05$ & $97.88$ & $0.38$ \\
  \midrule
  \multicolumn{6}{@{}l}{\textit{VLA-JEPA + LEON}} \\
  Spatial & $99.60$ & $98.80$ & $99.20$ & $99.20$ & $0.40$ \\
  Object & $99.60$ & $99.80$ & $99.60$ & $99.67$ & $0.12$ \\
  Goal & $98.40$ & $99.00$ & $99.20$ & $98.87$ & $0.42$ \\
  LIBERO-10 & $98.60$ & $97.40$ & $98.20$ & $98.07$ & $0.61$ \\
  \addlinespace[2pt]
  Overall & $99.05$ & $98.75$ & $99.05$ & $98.95$ & $0.17$ \\
  \midrule
  \multicolumn{6}{@{}l}{\textit{LaWAM}$^\dagger$} \\
  Spatial & $98.20$ & $98.00$ & $98.20$ & $98.13$ & $0.12$ \\
  Object & $98.80$ & $99.00$ & $99.00$ & $98.93$ & $0.12$ \\
  Goal & $98.80$ & $96.00$ & $99.60$ & $98.13$ & $1.89$ \\
  LIBERO-10 & $97.00$ & $98.20$ & $96.20$ & $97.13$ & $1.01$ \\
  \addlinespace[2pt]
  Overall & $98.20$ & $97.80$ & $98.25$ & $98.08$ & $0.25$ \\
  \midrule
  \multicolumn{6}{@{}l}{\textit{LaWAM + LEON}} \\
  Spatial & $99.00$ & $99.60$ & $98.80$ & $99.13$ & $0.42$ \\
  Object & $99.80$ & $100.00$ & $99.60$ & $99.80$ & $0.20$ \\
  Goal & $98.60$ & $98.00$ & $99.00$ & $98.53$ & $0.50$ \\
  LIBERO-10 & $97.80$ & $96.60$ & $98.20$ & $97.53$ & $0.83$ \\
  \addlinespace[2pt]
  Overall & $98.80$ & $98.55$ & $98.90$ & $98.75$ & $0.18$ \\
  \bottomrule
 \end{tabularx}
\end{table}

\paragraph{RoboTwin.}
The RoboTwin comparison evaluates LaWAM + LEON against the published LaWAM reference on all $50$ tasks under Clean and Randomized conditions~\citep{chen2026lawam}. Both conditions use the same final LEON checkpoint and the evaluation protocol in Section~\ref{app:leon-evaluation}. Task-level differences are expressed in percentage points relative to the published reference. Condition scores average the task success rates equally, and the overall score averages Clean and Randomized.

The complete task-level results are reported in Section~\ref{app:leon-robotwin-results}.

\subsection{Transition Comparisons and Ablations}
\label{app:leon-transition-comparisons}

We compare alternative transition architectures and four LEON component variants within VLA-JEPA. Each model is trained separately under the matched protocol in Section~\ref{app:leon-training-schedule} and evaluated using its own final LIBERO checkpoint. LIBERO-Plus evaluation follows Section~\ref{app:leon-evaluation} without additional training or adaptation.

\paragraph{Retrained transition baselines.}
The Transformer baseline retains the original VLA-JEPA predictor architecture~\citep{sun2026vlajepa}, with its transition parameters randomly initialized. The Residual MLP baseline retains LEON's observable map, action-conditioning module, and increment readout. The initial observable and its condition are concatenated once and processed by LayerNorm, an affine projection to width $1024$, and GELU. The MLP uses eight residual blocks to match the number of LEON updates, with a fixed residual scale of $1/8$. Each block applies LayerNorm followed by four affine layers of width $1024$, with GELU between consecutive affine layers. Independent weights allow each stage to learn a distinct update and give the blocks $8\times4\times1024^2$ weight parameters across their affine layers, compared with $4\times1024^2$ for a single block shared across stages. A bias-free linear projection maps the accumulated block increments to dimension $768$ before the retained increment readout. Each block's final affine layer is initialized to zero, giving an initial identity transition at the predictor feature interface. The remaining initialization follows Appendix~\ref{app:leon-initialization}. At this update depth, increasing the MLP parameter count requires wider blocks or additional layers within each block. In preliminary experiments, a larger MLP designed to approach the parameter count of the Transformer transition did not converge within the prescribed $40{,}000$ training updates.

\paragraph{Component variants.}
Table~\ref{tab:leon-e-component-variants} defines the component variants. All four retain $S=8$ Euler updates with $\eta=1/8$, shared field parameters, and the WAM input and output adapters. Except for the stated modifications, components follow Appendix~\ref{app:leon-specification}.

The retained VLA-JEPA input projection produces $1024$-dimensional predictor features. Full LEON maps these features to $768$-dimensional observables and maps the accumulated increment back to predictor coordinates. The \emph{w/o Observable Adaptation} variant omits both maps and evolves the predictor features directly. Retaining a state width of $768$ would require a separate projection whose outputs would themselves define a new set of observables, even if the projection were fixed. The conditioning and field networks therefore operate at the native predictor width of $1024$, with $M=96$ and the $256$-dimensional hidden layer of $g_p$ retained. Because each basis matrix acts on the state, the basis matrices contain $96\times1024^2=100{,}663{,}296$ parameters in this variant and $96\times768^2=56{,}623{,}104$ in full LEON. This comparison evaluates the choice of evolution coordinates together with the associated change in state width.

For the fixed-operator variant, $\bar A=D+\sum_{k=1}^{M}\bar\xi_k B_k$, where $\bar\xi=\tanh(\bar q)$ and $\bar q\in\mathbb R^M$ is a trainable, zero-initialized vector shared across tokens and updates. The complementary branch retains the dynamically inferred coefficients $\xi_i^{(n)}$. In the operator-only and additive-only variants, the retained branch keeps its full-model architecture.

\begin{table}[htbp]
 \centering
 \caption{LEON component variants. Token and internal step indices are omitted. Dynamic coefficients $\xi$ depend on the current state and fixed external condition.}
 \label{tab:leon-e-component-variants}
 \small
 \begin{tabularx}{\linewidth}{@{}p{0.27\linewidth}cY@{}}
  \toprule
  Variant & State dimension & Modification \\
  \midrule
  w/o Observable Adaptation & $1024$ & Identity observable map and increment readout; conditioning width $1024$ \\
  Fixed Operator & $768$ & $\bar A\mathbf h+b(\mathbf h,\xi)$; $\bar A$ is independent of the state and condition \\
  Operator Only & $768$ & $A(\xi)\mathbf h$; $g_p$ and $g_b$ are removed \\
  Additive Only & $768$ & $b(\mathbf h,\xi)$; $D$ and $\{B_k\}$ are removed \\
  \bottomrule
 \end{tabularx}
\end{table}

\par
\paragraph{LIBERO results.}
Table~\ref{tab:leon-e-libero-ablations} reports the transition comparisons with rollout seed $0$. The full-model row reproduces the corresponding seed-$0$ results from Table~\ref{tab:leon-e-libero-seeds}; the overall score averages the four suites equally.

\begin{table}[htbp]
 \centering
 \caption{LIBERO success rates (\%) for transition comparisons within VLA-JEPA. Every model is evaluated with rollout seed $0$ using its own final $40{,}000$-update checkpoint. Overall is the unweighted mean across the four suites. The best value in each column is bold.}
 \label{tab:leon-e-libero-ablations}
 \small
 \setlength{\tabcolsep}{4pt}
 \begin{tabularx}{\linewidth}{@{}p{0.26\linewidth}*{5}{>{\centering\arraybackslash}X}@{}}
  \toprule
  Method & Spatial & Object & Goal & LIBERO-10 & Overall \\
  \midrule
  LEON (full) & $\mathbf{99.60}$ & $99.60$ & $98.40$ & $\mathbf{98.60}$ & $\mathbf{99.05}$ \\
  \midrule
  Transformer & $97.60$ & $98.60$ & $98.00$ & $95.40$ & $97.40$ \\
  Residual MLP & $98.60$ & $99.80$ & $97.20$ & $93.80$ & $97.35$ \\
  \midrule
  w/o Observable Adaptation & $98.60$ & $99.60$ & $97.40$ & $96.60$ & $98.05$ \\
  Fixed Operator & $97.80$ & $99.20$ & $97.20$ & $95.80$ & $97.50$ \\
  Operator Only & $99.40$ & $\mathbf{100.00}$ & $\mathbf{99.20}$ & $94.40$ & $98.25$ \\
  Additive Only & $98.60$ & $99.40$ & $97.20$ & $93.20$ & $97.10$ \\
  \bottomrule
 \end{tabularx}
\end{table}

The full configuration achieves the highest overall success rate of $99.05\%$, compared with $97.40\%$ for Transformer and $97.35\%$ for Residual MLP. Operator Only is the strongest component variant overall at $98.25\%$. Each component variant has its largest deficit relative to the full model on LIBERO-10. Operator Only achieves the highest Object and Goal scores, whereas the full model leads on Spatial and LIBERO-10.

\paragraph{LIBERO-Plus results.}
Table~\ref{tab:leon-e-libero-plus} reports success rates across seven LIBERO-Plus perturbation families: background textures, lighting conditions, language instructions, camera viewpoints, object layouts, robot initial states, and sensor noise. Evaluation uses perturbed instances of the $40$ tasks from LIBERO-Spatial, LIBERO-Object, LIBERO-Goal, and LIBERO-10. We follow the public VLA-JEPA evaluation implementation~\citep{sun2026vlajepa}, retaining the benchmark-provided task lists, perturbation configurations, and initial states.\footnote{The evaluation entry points are \texttt{eval\_libero\_plus.sh} and \texttt{libero\_plus\_init.py} in the official VLA-JEPA repository (\texttt{https://github.com/ginwind/VLA-JEPA}). Perturbation-family assignments follow the LIBERO-Plus file \texttt{task\_classification.json}.}

Each perturbed task instance is evaluated once with rollout seed $0$. For each family, the success rate is the fraction of successful rollouts over all evaluated instances pooled across the four suites. Mean SR is the unweighted arithmetic mean of the seven family-level success rates. All models use the same evaluation configuration and their own final LIBERO checkpoints, without additional training or adaptation, as specified in Section~\ref{app:leon-evaluation}.

The complete LEON configuration achieves a mean success rate of $83.45\%$, compared with $79.44\%$ for Residual MLP and $79.54\%$ for Transformer.

\begin{table}[htbp]
 \centering
 \caption{LIBERO-Plus success rates (\%). All methods are instantiated within VLA-JEPA and evaluated with rollout seed $0$ using their own final LIBERO checkpoints. Mean SR is the unweighted mean across the seven perturbation families. Backg., Lang., and Robot init. denote Background, Language, and Robot Initialization, respectively. The best value in each column is bold.}
 \label{tab:leon-e-libero-plus}
 \small
 \setlength{\tabcolsep}{2.5pt}
 \begin{tabularx}{\linewidth}{@{}p{0.26\linewidth}*{8}{>{\centering\arraybackslash}X}@{}}
  \toprule
  Method & Backg. & Light & Lang. & Camera & Layout
  & \shortstack{Robot\\init.} & Noise & \shortstack{Mean\\SR} \\
  \midrule
  LEON (full) & $\mathbf{97.03}$ & $\mathbf{98.34}$ & $\mathbf{85.04}$ & $68.67$ & $\mathbf{87.54}$ & $\mathbf{67.68}$ & $\mathbf{79.89}$ & $\mathbf{83.45}$ \\
  \midrule
  Transformer & $93.50$ & $94.40$ & $84.26$ & $63.30$ & $84.59$ & $64.26$ & $72.45$ & $79.54$ \\
  Residual MLP & $92.75$ & $93.70$ & $84.45$ & $65.54$ & $83.93$ & $63.87$ & $71.83$ & $79.44$ \\
  \midrule
  w/o Observable Adaptation & $94.70$ & $93.78$ & $84.91$ & $\mathbf{68.98}$ & $84.33$ & $64.58$ & $75.33$ & $80.94$ \\
  Fixed Operator & $93.40$ & $92.29$ & $83.67$ & $65.35$ & $85.44$ & $65.42$ & $73.95$ & $79.93$ \\
  Operator Only & $94.33$ & $95.36$ & $84.19$ & $67.10$ & $86.43$ & $66.45$ & $76.00$ & $81.41$ \\
  Additive Only & $94.98$ & $95.01$ & $83.86$ & $67.73$ & $85.64$ & $65.55$ & $69.46$ & $80.32$ \\
  \bottomrule
 \end{tabularx}
\end{table}

The complete configuration has the highest mean success rate among the evaluated models. Relative to the full model, the variants without observable adaptation and with a fixed operator have lower means by $2.51$ and $3.52$ percentage points, respectively. Operator Only and Additive Only obtain $81.41\%$ and $80.32\%$, both below the combined update. LEON also achieves the highest success rate in six of the seven perturbation families; for Camera, the variant without observable adaptation scores $68.98\%$, compared with $68.67\%$ for the full model.

\subsection{RoboTwin Results}
\label{app:leon-robotwin-results}

RoboTwin 2.0 evaluates multi-task bimanual manipulation under Clean and Randomized conditions. Table~\ref{tab:leon-e-robotwin-tasks} reports success rates on all $50$ tasks, with $100$ trials per task under each condition, following the LaWAM evaluation protocol~\citep{chen2026lawam}. The LaWAM reference results are taken from the published paper. LaWAM + LEON uses the same final $100{,}000$-update checkpoint for both conditions, as specified in Section~\ref{app:leon-evaluation}.

LaWAM + LEON achieves mean success rates of $93.32\%$ on Clean and $90.42\%$ on Randomized, compared with $92.64\%$ and $89.80\%$ for the LaWAM reference. The corresponding increases are $0.68$ and $0.62$ percentage points. Averaging the two condition scores gives an overall success rate of $91.87\%$, compared with $91.22\%$ for LaWAM, an increase of $0.65$ percentage points.

\begingroup
\small
\setlength{\tabcolsep}{4pt}
\setlength{\LTleft}{0pt}
\setlength{\LTright}{0pt}
\setlength{\LTcapwidth}{\linewidth}
\renewcommand{\arraystretch}{1.08}
\begin{longtable}{@{}p{\dimexpr0.34\linewidth-6\tabcolsep\relax}*{6}{>{\centering\arraybackslash}p{\dimexpr0.11\linewidth-\tabcolsep\relax}}@{}}
 \caption{RoboTwin task success rates (\%) under Clean and Randomized conditions. LaWAM results are from the published reference~\citep{chen2026lawam}. $\Delta$ denotes LaWAM + LEON minus LaWAM in percentage points. Average is the unweighted mean over all $50$ tasks.}
 \label{tab:leon-e-robotwin-tasks}\\
  \toprule
  Task & \multicolumn{3}{c}{Clean} & \multicolumn{3}{c}{Randomized} \\
  \cmidrule(lr){2-4}\cmidrule(l){5-7}
  & \shortstack{LaWAM\\+ LEON} & LaWAM & $\Delta$
  & \shortstack{LaWAM\\+ LEON} & LaWAM & $\Delta$ \\
  \midrule
 \endfirsthead
 \caption[]{RoboTwin task success rates (continued).}\\
  \toprule
  Task & \multicolumn{3}{c}{Clean} & \multicolumn{3}{c}{Randomized} \\
  \cmidrule(lr){2-4}\cmidrule(l){5-7}
  & \shortstack{LaWAM\\+ LEON} & LaWAM & $\Delta$
  & \shortstack{LaWAM\\+ LEON} & LaWAM & $\Delta$ \\
  \midrule
 \endhead
  \midrule
  \multicolumn{7}{r@{}}{\footnotesize Continued on next page}\\
 \endfoot
  \bottomrule
 \endlastfoot
  Adjust Bottle & $98$ & $100$ & $-2$ & $100$ & $100$ & $0$ \\
  Beat Block Hammer & $89$ & $90$ & $-1$ & $87$ & $93$ & $-6$ \\
  Blocks Ranking RGB & $94$ & $97$ & $-3$ & $100$ & $100$ & $0$ \\
  Blocks Ranking Size & $89$ & $93$ & $-4$ & $93$ & $89$ & $+4$ \\
  Click Alarmclock & $100$ & $100$ & $0$ & $100$ & $100$ & $0$ \\
  Click Bell & $100$ & $100$ & $0$ & $100$ & $100$ & $0$ \\
  Dump Bin Bigbin & $96$ & $97$ & $-1$ & $97$ & $95$ & $+2$ \\
  Grab Roller & $100$ & $100$ & $0$ & $100$ & $100$ & $0$ \\
  Handover Block & $96$ & $96$ & $0$ & $85$ & $87$ & $-2$ \\
  Handover Mic & $97$ & $93$ & $+4$ & $95$ & $98$ & $-3$ \\
  Hanging Mug & $57$ & $51$ & $+6$ & $59$ & $43$ & $+16$ \\
  Lift Pot & $100$ & $100$ & $0$ & $96$ & $99$ & $-3$ \\
  Move Can Pot & $96$ & $98$ & $-2$ & $95$ & $93$ & $+2$ \\
  Move Pillbottle Pad & $98$ & $97$ & $+1$ & $86$ & $90$ & $-4$ \\
  Move Playingcard Away & $96$ & $100$ & $-4$ & $100$ & $100$ & $0$ \\
  Move Stapler Pad & $87$ & $94$ & $-7$ & $95$ & $87$ & $+8$ \\
  Open Laptop & $98$ & $100$ & $-2$ & $99$ & $100$ & $-1$ \\
  Open Microwave & $68$ & $41$ & $+27$ & $57$ & $43$ & $+14$ \\
  Pick Diverse Bottles & $92$ & $91$ & $+1$ & $79$ & $88$ & $-9$ \\
  Pick Dual Bottles & $100$ & $100$ & $0$ & $97$ & $95$ & $+2$ \\
  Place A2B Left & $96$ & $98$ & $-2$ & $87$ & $91$ & $-4$ \\
  Place A2B Right & $95$ & $89$ & $+6$ & $93$ & $94$ & $-1$ \\
  Place Bread Basket & $94$ & $92$ & $+2$ & $87$ & $85$ & $+2$ \\
  Place Bread Skillet & $88$ & $90$ & $-2$ & $84$ & $83$ & $+1$ \\
  Place Burger Fries & $96$ & $93$ & $+3$ & $97$ & $96$ & $+1$ \\
  Place Can Basket & $91$ & $92$ & $-1$ & $65$ & $65$ & $0$ \\
  Place Cans Plasticbox & $100$ & $100$ & $0$ & $95$ & $95$ & $0$ \\
  Place Container Plate & $100$ & $100$ & $0$ & $100$ & $100$ & $0$ \\
  Place Dual Shoes & $95$ & $98$ & $-3$ & $90$ & $94$ & $-4$ \\
  Place Empty Cup & $97$ & $99$ & $-2$ & $100$ & $100$ & $0$ \\
  Place Fan & $92$ & $92$ & $0$ & $91$ & $93$ & $-2$ \\
  Place Mouse Pad & $93$ & $91$ & $+2$ & $89$ & $84$ & $+5$ \\
  Place Object Basket & $93$ & $92$ & $+1$ & $88$ & $90$ & $-2$ \\
  Place Object Scale & $94$ & $95$ & $-1$ & $85$ & $88$ & $-3$ \\
  Place Object Stand & $99$ & $92$ & $+7$ & $95$ & $93$ & $+2$ \\
  Place Phone Stand & $96$ & $93$ & $+3$ & $89$ & $94$ & $-5$ \\
  Place Shoe & $97$ & $100$ & $-3$ & $100$ & $100$ & $0$ \\
  Press Stapler & $100$ & $98$ & $+2$ & $98$ & $97$ & $+1$ \\
  Put Bottles Dustbin & $92$ & $94$ & $-2$ & $88$ & $92$ & $-4$ \\
  Put Object Cabinet & $85$ & $90$ & $-5$ & $86$ & $82$ & $+4$ \\
  Rotate QRcode & $89$ & $94$ & $-5$ & $94$ & $89$ & $+5$ \\
  Scan Object & $93$ & $96$ & $-3$ & $85$ & $90$ & $-5$ \\
  Shake Bottle & $100$ & $100$ & $0$ & $100$ & $100$ & $0$ \\
  Shake Bottle Horizontally & $100$ & $100$ & $0$ & $100$ & $100$ & $0$ \\
  Stack Blocks Three & $92$ & $90$ & $+2$ & $83$ & $75$ & $+8$ \\
  Stack Blocks Two & $100$ & $100$ & $0$ & $99$ & $97$ & $+2$ \\
  Stack Bowls Three & $87$ & $90$ & $-3$ & $76$ & $80$ & $-4$ \\
  Stack Bowls Two & $97$ & $100$ & $-3$ & $100$ & $99$ & $+1$ \\
  Stamp Seal & $90$ & $89$ & $+1$ & $91$ & $88$ & $+3$ \\
  Turn Switch & $74$ & $47$ & $+27$ & $66$ & $56$ & $+10$ \\
  \midrule
  Average & $93.32$ & $92.64$ & $+0.68$ & $90.42$ & $89.80$ & $+0.62$ \\
\end{longtable}
\endgroup

\subsection{Closed-Loop Evaluation Protocol}
\label{app:leon-closed-loop-protocol}

\paragraph{Experimental setup.}
Figure~4 evaluates the LIBERO-10 task of placing the white mug on the left plate and the yellow-and-white mug on the right plate in \texttt{LIVING\_ROOM\_SCENE5}. We use all $50$ released initial states, indexed $0$--$49$, with the same state assigned to both policies under both perturbations. Each policy--condition pair therefore comprises $50$ rollouts. Environment and policy random seeds are reset to $0$ for each rollout. After $10$ settling steps, execution continues until native task success or a limit of $520$ control actions. Both policies generate seven-action chunks using four action-head inference steps. The LEON policy reuses the VLA-JEPA + LEON checkpoint evaluated on LIBERO in Table~\ref{tab:leon-e-libero-seeds}. The Transformer policy reuses the retrained Transformer baseline checkpoint evaluated in Table~\ref{tab:leon-e-libero-ablations}. Both are final $40{,}000$-update checkpoints and are evaluated without additional training or adaptation, retaining their associated action-normalization statistics.

\paragraph{Visual perturbations.}
The silver-background condition replaces only the table's RGB texture with the LIBERO-Plus asset \texttt{FabricRope001\_DISP\_4K.png}, used as a color texture. The blur condition applies the LIBERO-Plus Gaussian-blur function at severity $2$, corresponding to $\sigma=2$ pixels, to both external-camera and wrist-camera images. Blurring is applied to the $256\times256$ observations before resizing to $224\times224$ policy inputs. Both conditions preserve the original dynamics, camera geometry, and task goals. These two appearance conditions use LIBERO-Plus resources for a single-task evaluation.

\paragraph{Evaluation metrics.}
First-mug placement records whether the white mug satisfies the original placement predicate on the left plate at rollout termination. Full-task completion uses the native LIBERO success criterion, requiring both mugs to satisfy their respective placement predicates simultaneously. The task does not impose a placement order. An annotation $x/50$ counts rollouts satisfying the stated terminal criterion over all $50$ initial states for the corresponding policy and condition.

\paragraph{Stage-level results.}
Table~\ref{tab:leon-e-closed-loop-stages} reports the terminal success counts corresponding to Figure~4. Both metrics use all $50$ paired initial states for each policy and perturbation condition.

\begin{table}[htbp]
 \centering
 \caption{Stage-level success counts for the Figure~4 closed-loop evaluation. Each entry reports successful trials out of $50$. First-mug placement refers to the white mug on the left plate. Full-task completion requires both native placement goals to hold simultaneously at rollout termination.}
 \label{tab:leon-e-closed-loop-stages}
 \small
 \setlength{\tabcolsep}{4pt}
 \begin{tabularx}{\linewidth}{@{}YY*{2}{>{\centering\arraybackslash}X}@{}}
  \toprule
  Visual condition & Method & \shortstack{First-mug\\placement} & \shortstack{Full-task\\completion} \\
  \midrule
  Silver background & LEON & $50/50$ & $50/50$ \\
   & Transformer & $48/50$ & $44/50$ \\
  \midrule
  Dual-view Gaussian blur & LEON & $45/50$ & $43/50$ \\
   & Transformer & $24/50$ & $18/50$ \\
  \bottomrule
 \end{tabularx}
\end{table}

\paragraph{Trajectory visualization.}
Illustrative trajectories are selected for clear differences in task-stage completion, with the same initial state used for each policy pair. Frames show initialization, verified placement or transport events when reached, and the actual terminal state. Wider views are rendered by replaying recorded actions without additional policy rollouts. Clean observations serve as visual references. Frame selection does not alter the $50$-rollout denominators, and the results characterize the two evaluated checkpoints on this task.

%% file: appendices/appendix_f_dynamical_system_diagnostics.tex
\begingroup
\raggedbottom
\setlength{\LTleft}{0pt}
\setlength{\LTright}{0pt}
\setlength{\LTpre}{4pt}
\setlength{\LTpost}{6pt}
\setlength{\LTcapwidth}{\linewidth}
\section{Controlled Dynamical-System Diagnostics}
\label{app:leon-diagnostics}

We evaluate LEON on three dynamical systems with known governing equations to examine prediction under controlled changes in initial conditions and system parameters. The damped spring tests identification from noisy observations and amplitude extrapolation in a system with exact linear closure. The damped pendulum tests recursive prediction across energy regimes under nonlinear dynamics. The driven oscillator tests stiffness-conditioned prediction and the roles of the operator and complementary terms through retrained ablations and test-time interventions. For each system, LEON is implemented using physical states or compact observations. This appendix specifies the experimental settings and reports results over five seeds.

\subsection{Experimental Protocol}
\label{app:leon-diagnostic-protocol}

\paragraph{Data generation.}
Reference trajectories are generated by fourth-order Runge--Kutta integration with sampling interval $\Delta t=0.05\,\mathrm{s}$. All methods use the same training and test trajectories within each seed. Neural predictors and trainable fixed-operator baselines use the same system-specific normalization. DMD is fitted directly to snapshots in physical coordinates, as described in Section~\ref{app:leon-spring}. Predictions are generated recursively after the measured context. LEON and the MLP use the most recent observation, whereas the Transformer uses a rolling history. Table~\ref{tab:leon-f-protocol} summarizes the data and training budgets. The five seeds are $300$--$304$ for the spring, $0$--$4$ for the pendulum, and $1400$--$1404$ for the driven oscillator. Each seed determines the data realization and model initialization where applicable. All five runs are included in every comparison.

\begingroup
\small
\setlength{\tabcolsep}{3pt}
\renewcommand{\arraystretch}{1.08}
\begin{longtable}{@{\extracolsep{\fill}}lccc@{}}
\caption{Data and training budgets. Trajectory counts are per seed, and test counts apply to each evaluation regime. Lengths include the initial state sample. Prediction horizons count physical sampling intervals after the measured context.}
\label{tab:leon-f-protocol}\\[-5pt]
\toprule
Setting & Spring & Pendulum & Driven oscillator \\
\midrule
\endfirsthead
\multicolumn{4}{@{}l@{}}{\tablename~\thetable\ (continued)}\\[2pt]
\toprule
Setting & Spring & Pendulum & Driven oscillator \\
\midrule
\endhead
\endfoot
\bottomrule
\endlastfoot
Training / ID validation / test trajectories & $16/24/128$ & $128/48/128$ & $32/48/128$ \\
  Training and ID validation length & $520$ & $180$ & $129$ \\
  Test length & $520$ & $190$ & $105$ \\
  Context length & $8$ & $8$ & $4$ \\
  Supervised prediction steps & $\{1,10,30,60\}$ & $\{1,5,10,20,40,80\}$ & $\{1,\ldots,8\}$ \\
  Evaluation horizon & $400$ & $160$ & $80$ \\
  Evaluation duration (s) & $20$ & $8$ & $4$ \\
  Training updates & $1{,}400$ & $1{,}600$ & $1{,}400$ \\
  Global batch size & $96$ & $64$ & $96$ \\
\end{longtable}
\endgroup

\paragraph{Model architecture.}
Let $\mathbf z_\ell\in\mathbb R^{d_z}$ denote the observation at physical sample $\ell$, and let $\overline{\mathbf z}_\ell=(\mathbf z_\ell-\boldsymbol\mu)\oslash\boldsymbol\sigma$ use the training normalization specified for each system, with $\boldsymbol\mu=0$ for the spring and driven oscillator. An affine projection and a nonlinear residual construct $\mathbf x_\ell=\gamma[P_{\mathrm{in}}\overline{\mathbf z}_\ell+f_{\mathrm{in}}(\overline{\mathbf z}_\ell)]$. LEON predicts the next observation by
\begin{equation}
 \begin{aligned}
  \mathbf h_\ell^{(0)}&=\phi(\mathbf x_\ell),\\
  \mathbf h_\ell^{(n+1)}&=\mathbf h_\ell^{(n)}
   +\eta F_\theta(\mathbf h_\ell^{(n)},\mathbf c_\ell),
   \qquad n=0,\ldots,S-1,\\
  \widehat{\mathbf z}_{\ell+1}&=\mathbf z_\ell+
   \boldsymbol\sigma\odot\left[
    \gamma^{-1}P_{\mathrm{out}}W_\psi
    (\mathbf h_\ell^{(S)}-\mathbf h_\ell^{(0)})\right].
 \end{aligned}
 \label{eq:leon-f-transition}
\end{equation}
The field is $F_\theta(\mathbf h,\mathbf c)=A(\boldsymbol\xi_{\mathrm{op}})\mathbf h+b(\mathbf h,\boldsymbol\xi_b)$, where $A(\boldsymbol\xi)=D+\sum_{k=1}^{M}\xi_kB_k$, $D=\operatorname{diag}(0.1\tanh\mathbf d)$, and each $B_k\in\mathbb R^{d_h\times d_h}$ is an unrestricted learned matrix. The affine observable map $\phi$ is evaluated once per physical prediction. Matrices and field networks are shared across internal updates, and $W_\psi$ and $P_{\mathrm{out}}$ are bias-free linear increment readouts. Each physical sampling interval is represented by a unit normalized horizon, with $\eta=1/S$. Thus $\Delta t$ is the physical sampling interval, $S$ counts internal updates for one predicted sample, and the supervised rollout horizon counts successive physical predictions.

The spring and pendulum use constant external conditions and state-dependent coefficients shared by both terms. Their complementary term is $b(\mathbf h,\boldsymbol\xi)=g_b(\mathbf h+g_p(\boldsymbol\xi))$. The oscillator uses the same form for $b$ and a separate stiffness-dependent coefficient network for the operator, as specified in Section~\ref{app:leon-driven-oscillator}. Table~\ref{tab:leon-f-models} lists the network configurations. Input residual networks have two layers of width $d_x$ and a zero-initialized output layer. The spring and oscillator use bias-free layers with LeakyReLU of slope $0.1$. The pendulum uses affine layers with SiLU. Conditioning and coefficient networks have two affine layers with hidden width $d_h$, and the fused coefficients use a Tanh output activation. These networks and $g_b$ use SiLU, whereas $g_p$ uses GELU. The coefficient embedding and complementary network apply LayerNorm to their inputs. The observable map is affine and has no input LayerNorm.

\begingroup
\small
\setlength{\tabcolsep}{3pt}
\renewcommand{\arraystretch}{1.08}
\begin{longtable}{@{\extracolsep{\fill}}lccc@{}}
\caption{Network configurations. MLP tuples specify input, hidden, and output dimensions. Transformer tuples specify the number of layers, model width, and feed-forward width. All Transformers use four attention heads. Parameter counts include active trainable parameters.}
\label{tab:leon-f-models}\\[-5pt]
\toprule
Configuration & Spring & Pendulum & Driven oscillator \\
\midrule
\endfirsthead
\multicolumn{4}{@{}l@{}}{\tablename~\thetable\ (continued)}\\[2pt]
\toprule
Configuration & Spring & Pendulum & Driven oscillator \\
\midrule
\endhead
\endfoot
\bottomrule
\endlastfoot
Observation dimension $d_z$ & $2$ & $3$ & $2$ \\
  Input feature dimension $d_x$ & $64$ & $64$ & $64$ \\
  Observable dimension $d_h$ & $32$ & $48$ & $32$ \\
  Basis matrices $M$ & $32$ & $24$ & $4$ \\
  Internal updates $S$ & $16$ & $16$ & $1$ \\
  Normalized step size $\eta$ & $1/16$ & $1/16$ & $1$ \\
  Input scaling $\gamma$ & $0.25$ & $1$ & $0.25$ \\
  Coefficient embedding $g_p$ & $(32,64,32)$ & $(24,256,48)$ & $(4,256,32)$ \\
  Complementary network $g_b$ & $(32,64,32)$ & $(48,48,48)$ & $(32,32,32)$ \\
  LEON parameters & $58{,}624$ & $106{,}408$ & $32{,}592$ \\
  \midrule
  Residual MLP widths & $(2,11724,2)$ & $(3,15201,3)$ & $(4,4656,2)$ \\
  Residual MLP parameters & $58{,}622$ & $106{,}410$ & $32{,}594$ \\
  Transformer configuration & $(2,48,191)$ & $(2,64,277)$ & $(1,48,230)$ \\
  Transformer parameters & $58{,}608$ & $106{,}477$ & $32{,}584$ \\
\end{longtable}
\endgroup

\paragraph{Training.}
Neural predictors and trainable fixed-operator baselines are trained from scratch with AdamW and $(\beta_1,\beta_2)=(0.9,0.999)$. The active parameter counts of the MLP and Transformer differ from that of the full LEON model by less than $1\%$. MLPs use one SiLU hidden layer and predict residual increments. Transformers use pre-normalized causal attention, GELU feed-forward layers, zero dropout, and a final LayerNorm. The spring Transformer uses a two-layer output head with hidden width $32$. The other two Transformers use linear output heads.

Spring and pendulum training average normalized squared prediction errors at the supervised endpoints, introducing longer horizons progressively. The oscillator supervises all eight physical prediction steps of its training rollout and uses the conditional regularization described in Section~\ref{app:leon-driven-oscillator}. Table~\ref{tab:leon-f-optimization} gives the optimization settings. Cosine schedules decay to $10\%$ of the peak learning rate. The fixed-operator baselines are trained separately, with the architectures specified in their respective system descriptions.

\begingroup
\small
\setlength{\tabcolsep}{3pt}
\renewcommand{\arraystretch}{1.08}
\begin{longtable}{@{\extracolsep{\fill}}llccccc@{}}
\caption{Optimization settings. Peak LR is the peak learning rate for the input/output modules or the complete baseline. Gradient clipping uses the stated maximum gradient norm. Spring LEON uses learning-rate multipliers of $4$, $5$, and $0.5$ for the coefficient/conditioning networks, basis matrices, and complementary term, respectively. Oscillator LEON and its retrained variants use multiplier $4$ for all three groups and zero weight decay for the complementary term. All pendulum parameters use one learning rate. Training budgets include warmup.}
\label{tab:leon-f-optimization}\\[-5pt]
\toprule
System & Predictor & Peak LR & \shortstack{Warmup\\updates} & \shortstack{Weight\\decay} & \shortstack{Gradient\\clipping} & Schedule \\
\midrule
\endfirsthead
\multicolumn{7}{@{}l@{}}{\tablename~\thetable\ (continued)}\\[2pt]
\toprule
System & Predictor & Peak LR & \shortstack{Warmup\\updates} & \shortstack{Weight\\decay} & \shortstack{Gradient\\clipping} & Schedule \\
\midrule
\endhead
\endfoot
\bottomrule
\endlastfoot
Spring & LEON & $1.6\!\times\!10^{-3}$ & $0$ & $10^{-5}$ & $5$ & Cosine \\*
   & MLP & $10^{-4}$ & $0$ & $10^{-3}$ & $2$ & Cosine \\*
   & Transformer & $6.4\!\times\!10^{-3}$ & $0$ & $10^{-3}$ & $5$ & Constant \\*
   & Fixed operator & $8\!\times\!10^{-4}$ & $0$ & $10^{-4}$ & $1$ & Constant \\
  \midrule
  Pendulum & LEON & $8\!\times\!10^{-4}$ & $100$ & $10^{-4}$ & $1$ & Cosine \\*
   & MLP & $2\!\times\!10^{-4}$ & $100$ & $10^{-2}$ & $1$ & Cosine \\*
   & Transformer & $3.2\!\times\!10^{-3}$ & $100$ & $10^{-2}$ & $1$ & Cosine \\*
   & Fixed operator & $8\!\times\!10^{-4}$ & $100$ & $10^{-4}$ & $1$ & Cosine \\
  \midrule
  Oscillator & LEON / variants & $4\!\times\!10^{-3}$ & $200$ & $10^{-3}$ & $0.25$ & Cosine \\*
   & MLP & $4\!\times\!10^{-3}$ & $100$ & $10^{-6}$ & $1$ & Cosine \\*
   & Transformer & $8\!\times\!10^{-3}$ & $100$ & $10^{-6}$ & $1$ & Cosine \\
\end{longtable}
\endgroup

\paragraph{Model selection.}
Pendulum checkpoints minimize normalized multihorizon error on fixed ID validation windows. Spring and oscillator checkpoints additionally use independent validation sets between the training and OOD test ranges. These contain $24$ spring trajectories with $|q_0|\in[1.25,2.25]$ and $|v_0|\in[1.75,2.25]$, and $48$ oscillator trajectories with $\kappa\in[4.05,4.4]$. Eligible checkpoints satisfy ID short-horizon normalized-MSE thresholds of $0.02$ and $0.01$, respectively. Selection minimizes the square root of the weighted long-horizon MSE, using weights $0.25$ for ID validation and $0.75$ for the additional validation set. Long validation rollouts span $20$ and $4\,\mathrm{s}$ for the spring and oscillator, respectively. Test trajectories are excluded from model selection.

\paragraph{Evaluation metrics.}
For $J$ trajectories, $H$ predicted samples, and physical-state dimension $d_s$, cumulative state RMSE is
\begin{equation}
 R_s(H)=\left[\frac{1}{JHd_s}\sum_{j=1}^{J}\sum_{\ell=1}^{H}
  \|\widehat{\mathbf s}_{j,\ell}-\mathbf s_{j,\ell}\|_2^2\right]^{1/2}.
 \label{eq:leon-f-state-rmse}
\end{equation}
Spring and oscillator errors are computed from position and velocity in physical coordinates. Pendulum angle errors use wrapped differences $e_\theta=\operatorname{atan2}(\sin(\widehat\theta-\theta),\cos(\widehat\theta-\theta))$. Angle and angular-velocity RMSE are reported separately. Unless specified as a terminal error, a metric at $t$ includes the first $H=t/\Delta t$ predicted samples. Teacher-forced one-step evaluation uses measured histories at every eligible test transition. Metrics are computed within each seed and then averaged over five seeds. Entries with $\pm$ report the mean and sample standard deviation. Other entries report the mean. Trajectory quantiles are computed within each seed before averaging.

\subsection{Damped Spring}
\label{app:leon-spring}

\paragraph{System dynamics.}
The unit-mass damped spring has stiffness $4$ and damping coefficient $0.12$. For $\mathbf s=(q,v)^\top$, its dynamics and exact sampled flow are
\begin{equation}
 \dot{\mathbf s}=F_{\mathrm s}\mathbf s,
 \qquad
 F_{\mathrm s}=\begin{pmatrix}0&1\\-4&-0.12\end{pmatrix},
 \qquad
 \mathbf s(t+\Delta t)=K_{\mathrm s}\mathbf s(t),
 \quad K_{\mathrm s}=\exp(\Delta t F_{\mathrm s}).
 \label{eq:leon-f-spring-flow}
\end{equation}
The state coordinates form an exactly closed linear representation, and the same transition matrix governs all initial amplitudes. This system tests whether LEON can identify the underlying dynamics from limited noisy observations and retain prediction accuracy outside the training amplitude range. Since a fixed linear transition is sufficient to represent the exact flow, the experiment evaluates system identification and extrapolation under the stated data and training conditions.

\paragraph{Dataset.}
Initial states for training, ID validation, and ID testing are sampled from $q_0\sim\mathcal U[-1,1]$ and $v_0\sim\mathcal U[-1.5,1.5]$. Amplitude-OOD tests independently sample both magnitudes from $[2.5,4]$ with random signs. Training observations contain independent Gaussian noise with standard deviation $0.015$. Validation and test states are clean. Neural predictors and the trainable fixed operator divide states by training standard deviations with lower bounds $(0.25,0.35)$, without mean subtraction.

\paragraph{Baselines.}
Dynamic mode decomposition (DMD) estimates a fixed linear transition from adjacent training snapshots. Let $\mathbf S_-,\mathbf S_+\in\mathbb R^{2\times N_{\mathrm{pair}}}$ collect the corresponding current and next noisy states in physical coordinates. The ridge-regularized least-squares estimate is
\begin{equation}
 K_{\mathrm{DMD}}=\mathbf S_+\mathbf S_-^\top
  \bigl(\mathbf S_-\mathbf S_-^\top+\lambda_{\mathrm{DMD}}I_2\bigr)^{-1},
 \qquad \lambda_{\mathrm{DMD}}=10^{-5}.
 \label{eq:leon-f-dmd}
\end{equation}
DMD is fitted once per seed using the same noisy training trajectories as the other methods. It requires no gradient-based optimization or checkpoint selection. The fitted matrix depends on the training snapshots and observation noise, whereas $K_{\mathrm s}$ in Eq.~\eqref{eq:leon-f-spring-flow} is the exact transition of the governing equations. The trainable fixed-operator baseline uses $K_0=I+0.20\tanh R$ with four parameters. LEON retains both update terms. Table~\ref{tab:leon-f-spring-prediction} reports cumulative state RMSE. LEON achieves the lowest mean error at all listed horizons in both regimes.

\begingroup
\small
\setlength{\tabcolsep}{3pt}
\renewcommand{\arraystretch}{1.08}
\begin{longtable}{@{\extracolsep{\fill}}llrrrr@{}}
\caption{Spring cumulative state RMSE in physical coordinates. Results are means $\pm$ sample standard deviations over five seeds. Each regime contains $128$ test trajectories per seed.}
\label{tab:leon-f-spring-prediction}\\[-5pt]
\toprule
Regime & Predictor & $0.5$ s & $3$ s & $10$ s & $20$ s \\
\midrule
\endfirsthead
\multicolumn{6}{@{}l@{}}{\tablename~\thetable\ (continued)}\\[2pt]
\toprule
Regime & Predictor & $0.5$ s & $3$ s & $10$ s & $20$ s \\
\midrule
\endhead
\endfoot
\bottomrule
\endlastfoot
ID & DMD & \(0.00734\pm 0.00094\) & \(0.03688\pm 0.00531\) & \(0.08464\pm 0.01116\) & \(0.10372\pm 0.01233\) \\*
   & Fixed operator & \(0.13461\pm 0.00572\) & \(0.16484\pm 0.00691\) & \(0.14289\pm 0.00655\) & \(0.12443\pm 0.00616\) \\*
   & Residual MLP & \(0.00789\pm 0.00380\) & \(0.03233\pm 0.01642\) & \(0.08270\pm 0.04830\) & \(0.11922\pm 0.07852\) \\*
   & Transformer & \(0.01817\pm 0.00639\) & \(0.04169\pm 0.00873\) & \(0.09100\pm 0.02244\) & \(0.13788\pm 0.04201\) \\*
   & LEON & \(0.00231\pm 0.00123\) & \(0.00431\pm 0.00191\) & \(0.00753\pm 0.00363\) & \(0.00964\pm 0.00484\) \\
  \midrule
  Amplitude OOD & DMD & \(0.03710\pm 0.00595\) & \(0.17739\pm 0.02710\) & \(0.42760\pm 0.05994\) & \(0.52407\pm 0.06739\) \\*
   & Fixed operator & \(0.69202\pm 0.01115\) & \(0.83706\pm 0.01346\) & \(0.72405\pm 0.01229\) & \(0.62698\pm 0.01162\) \\*
   & Residual MLP & \(0.06895\pm 0.01098\) & \(0.22787\pm 0.06913\) & \(0.45015\pm 0.14982\) & \(0.51813\pm 0.12075\) \\*
   & Transformer & \(1.21388\pm 0.12382\) & \(3.10572\pm 0.21546\) & \(3.04065\pm 0.18316\) & \(2.49459\pm 0.15472\) \\*
   & LEON & \(0.01634\pm 0.01173\) & \(0.04928\pm 0.04855\) & \(0.07985\pm 0.06882\) & \(0.09032\pm 0.07096\) \\
\end{longtable}
\endgroup

\paragraph{Physical consistency.}
Linearity implies scale equivariance, $K_{\mathrm s}(a\mathbf s)=aK_{\mathrm s}\mathbf s$. The mechanical energy $E(q,v)=2q^2+v^2/2$ obeys $\dot E=-0.12v^2$, so the energy diagnostics assess the predicted dissipative evolution. Energy MAE averages the absolute predicted-versus-reference energy difference over trajectories and forecast times. The energy-balance residual averages $|E(\widehat{\mathbf s}_{\ell+1})-E(\widehat{\mathbf s}_\ell)+0.12\Delta t(\widehat v_\ell^2+\widehat v_{\ell+1}^2)/2|$, using a trapezoidal approximation to the dissipated energy. Scale-equivariance error averages $\|T(3\mathbf s)-3T(\mathbf s)\|_2/\max(\|3T(\mathbf s)\|_2,10^{-6})$ over common validation probes, where $T$ denotes a one-step predictor. Transformer probes repeat the state over the context. Table~\ref{tab:leon-f-spring-physics} shows lower energy errors for LEON among the reported methods, whereas DMD preserves scale equivariance up to numerical precision.

\begingroup
\small
\setlength{\tabcolsep}{3pt}
\renewcommand{\arraystretch}{1.08}
\begin{longtable}{@{\extracolsep{\fill}}lrrr@{}}
\caption{Spring physical consistency metrics, averaged over five seeds. Scale equivariance uses common validation probes. Energy metrics use amplitude-OOD rollouts. Lower values indicate better agreement.}
\label{tab:leon-f-spring-physics}\\[-5pt]
\toprule
Predictor & Scale-equivariance error & Energy MAE ($20$ s) & Energy-balance residual \\
\midrule
\endfirsthead
\multicolumn{4}{@{}l@{}}{\tablename~\thetable\ (continued)}\\[2pt]
\toprule
Predictor & Scale-equivariance error & Energy MAE ($20$ s) & Energy-balance residual \\
\midrule
\endhead
\endfoot
\bottomrule
\endlastfoot
Transformer & \(4.588\times10^{-2}\) & \(7.31097\) & \(0.04994\) \\*
  DMD & \(3.819\times10^{-8}\) & \(2.89906\) & \(0.02141\) \\*
  LEON & \(3.064\times10^{-4}\) & \(0.39522\) & \(0.00869\) \\
\end{longtable}
\endgroup

\paragraph{Amplitude extrapolation.}
We additionally evaluate the same frozen checkpoints on a phase-controlled initial-state grid. For $A_0=1.25r$, initial states are $q_0=A_0\cos\varphi$ and $v_0=-2A_0\sin\varphi$, with $\varphi=2\pi j/32$ for $j=0,\ldots,31$. Here $A_0=\sqrt{q_0^2+(v_0/2)^2}$ is the initial energy-coordinate radius. The reference value $1.25$ is the maximum radius within the training rectangle. Whether an initial state lies inside this rectangle also depends on its phase. Measured context spans $0$--$0.35\,\mathrm{s}$, and forecasts cover $0.40$--$20.35\,\mathrm{s}$. Reported durations are measured from the $0.35\,\mathrm{s}$ prediction anchor. Errors pool the $32$ phases within each seed before aggregation. This sweep is a separate evaluation from the randomly sampled amplitude-OOD test and does not involve retraining or checkpoint selection.

\begingroup
\small
\setlength{\tabcolsep}{3pt}
\renewcommand{\arraystretch}{1.08}
\begin{longtable}{@{\extracolsep{\fill}}rrrrrrr@{}}
\caption{Phase-controlled amplitude sweep: mean cumulative $20$-s state RMSE over five seeds. Each amplitude has $32$ phases per seed. Outside denotes the percentage of initial states outside the training rectangle. All methods use the same phase grid and frozen checkpoints.}
\label{tab:leon-f-amplitude-sweep}\\[-5pt]
\toprule
$A_0$ & Outside (\%) & DMD & Fixed operator & MLP & Transformer & LEON \\
\midrule
\endfirsthead
\multicolumn{7}{@{}l@{}}{\tablename~\thetable\ (continued)}\\[2pt]
\toprule
$A_0$ & Outside (\%) & DMD & Fixed operator & MLP & Transformer & LEON \\
\midrule
\endhead
\endfoot
\bottomrule
\endlastfoot
$0.3125$ & $0$ & $0.04492$ & $0.05283$ & $0.05776$ & $0.06729$ & $0.00399$ \\
  $0.625$ & $0$ & $0.08983$ & $0.10566$ & $0.10855$ & $0.13770$ & $0.00847$ \\
  $0.75$ & $0$ & $0.10780$ & $0.12679$ & $0.12574$ & $0.16885$ & $0.01015$ \\
  $0.9375$ & $43.75$ & $0.13475$ & $0.15848$ & $0.14812$ & $0.16486$ & $0.01243$ \\
  $1$ & $43.75$ & $0.14373$ & $0.16905$ & $0.15485$ & $0.15595$ & $0.01315$ \\
  $1.25$ & $100$ & $0.17966$ & $0.21131$ & $0.17982$ & $0.14127$ & $0.01682$ \\
  $1.5625$ & $100$ & $0.22458$ & $0.26414$ & $0.21087$ & $0.29181$ & $0.02334$ \\
  $1.875$ & $100$ & $0.26949$ & $0.31697$ & $0.24488$ & $0.53858$ & $0.03113$ \\
  $2.5$ & $100$ & $0.35933$ & $0.42262$ & $0.32528$ & $1.13793$ & $0.04869$ \\
  $3.125$ & $100$ & $0.44916$ & $0.52828$ & $0.42166$ & $1.81401$ & $0.06856$ \\
  $3.75$ & $100$ & $0.53899$ & $0.63393$ & $0.53100$ & $2.52143$ & $0.09231$ \\
  $5$ & $100$ & $0.71865$ & $0.84524$ & $0.77658$ & $3.87844$ & $0.15223$ \\
  $6.25$ & $100$ & $0.89831$ & $1.05655$ & $1.04273$ & $4.98575$ & $0.22850$ \\
  $7.5$ & $100$ & $1.07798$ & $1.26786$ & $1.31820$ & $5.82138$ & $0.32362$ \\
\end{longtable}
\endgroup

\subsection{Damped Pendulum}
\label{app:leon-pendulum}

\paragraph{System dynamics.}
The damped pendulum has nominal length $L=1$, gravitational acceleration $g=9.81$, and angular damping coefficient $c=0.12$. It follows $\dot\theta=\omega$ and $\dot\omega=-9.81\sin\theta-0.12\omega$, with energy $E(\theta,\omega)=\omega^2/2+9.81(1-\cos\theta)$ and dissipation $\dot E=-0.12\omega^2$. The nonlinear restoring force produces energy-dependent motion, with qualitatively different trajectories near the conservative separatrix. This system tests whether LEON's learned observable dynamics support recursive prediction beyond the low-energy states used for training.

\paragraph{Dataset.}
Initial states are sampled uniformly from $[-\pi,\pi)\times[-4.5,4.5)$ and accepted according to initial energy. Training, validation, and ID testing use $E_0\in[0.5,12)$. Near OOD, Far OOD, and High OOD use $[12,16)$, $[16,19)$, and $[20.5,28)$, respectively. Far and High OOD lie below and above the conservative separatrix energy $19.62$. These regimes are defined by initial energy, which subsequently decreases under damping. Trajectories contain no observation noise.

The observation is $\mathbf z=(\sin\theta,\cos\theta-1,\omega)$, centered and scaled by training statistics with standard-deviation floors $(0.35,0.20,0.55)$. Predictions decode angle as $\widehat\theta=\operatorname{atan2}(\widehat z_1,\widehat z_2+1)$. The separately trained fixed-operator baseline evolves learned observables with input-independent coefficients, using a two-block predictor with input width $24$, observable width $20$, and $10{,}830$ active parameters. The pendulum curve labelled \emph{w/o modulation} in Figure~\ref{fig:controlled-tests}(a) corresponds to this independently trained baseline.

\paragraph{Prediction accuracy.}
Table~\ref{tab:leon-f-pendulum-prediction} reports cumulative angle and angular-velocity errors across the four energy regimes. All statistics include all five runs, including the large MLP angular-velocity errors.

\begingroup
\small
\setlength{\tabcolsep}{3pt}
\renewcommand{\arraystretch}{1.08}
\begin{longtable}{@{\extracolsep{\fill}}llrrrr@{}}
\caption{Pendulum cumulative prediction errors, reported as means $\pm$ sample standard deviations over five seeds. Angle errors use wrapped differences. Angular-velocity errors are measured in physical coordinates.}
\label{tab:leon-f-pendulum-prediction}\\[-5pt]
\toprule
Regime & Predictor & \shortstack{Angle $1$ s\\(rad)} & \shortstack{Angle $4$ s\\(rad)} & \shortstack{Angle $8$ s\\(rad)} & \shortstack{Angular velocity\\$8$ s (rad/s)} \\
\midrule
\endfirsthead
\multicolumn{6}{@{}l@{}}{\tablename~\thetable\ (continued)}\\[2pt]
\toprule
Regime & Predictor & \shortstack{Angle $1$ s\\(rad)} & \shortstack{Angle $4$ s\\(rad)} & \shortstack{Angle $8$ s\\(rad)} & \shortstack{Angular velocity\\$8$ s (rad/s)} \\
\midrule
\endhead
\endfoot
\bottomrule
\endlastfoot
ID & Transformer & \(0.01490\pm 0.01189\) & \(0.02476\pm 0.03055\) & \(0.03527\pm 0.04823\) & \(0.10207\pm 0.13833\) \\*
   & Fixed operator & \(0.03117\pm 0.00394\) & \(0.03621\pm 0.00887\) & \(0.04248\pm 0.01386\) & \(0.12519\pm 0.04573\) \\*
   & Residual MLP & \(0.10265\pm 0.01548\) & \(0.18507\pm 0.04261\) & \(0.24901\pm 0.06686\) & \(0.68062\pm 0.15805\) \\*
   & LEON & \(0.00445\pm 0.00083\) & \(0.00428\pm 0.00088\) & \(0.00515\pm 0.00061\) & \(0.01524\pm 0.00204\) \\
  \midrule
  Near OOD & Transformer & \(0.14366\pm 0.05756\) & \(0.24999\pm 0.14789\) & \(0.31877\pm 0.19189\) & \(0.86295\pm 0.53351\) \\*
   & Fixed operator & \(0.17207\pm 0.03591\) & \(0.22335\pm 0.02889\) & \(0.31602\pm 0.05238\) & \(0.88547\pm 0.15043\) \\*
   & Residual MLP & \(0.56892\pm 0.16314\) & \(0.73053\pm 0.12348\) & \(0.85677\pm 0.13032\) & \((9.917\pm 21.845)\times10^{3}\) \\*
   & LEON & \(0.04075\pm 0.00318\) & \(0.04195\pm 0.00348\) & \(0.04890\pm 0.00834\) & \(0.13030\pm 0.02297\) \\
  \midrule
  Far OOD & Transformer & \(0.53883\pm 0.13283\) & \(0.86887\pm 0.25052\) & \(1.03745\pm 0.27081\) & \(2.70250\pm 0.70937\) \\*
   & Fixed operator & \(0.59509\pm 0.12163\) & \(0.71016\pm 0.07162\) & \(0.90660\pm 0.09555\) & \(2.36247\pm 0.23851\) \\*
   & Residual MLP & \(1.10344\pm 0.27427\) & \(1.36222\pm 0.11852\) & \(1.48400\pm 0.13680\) & \((5.460\pm 12.014)\times10^{4}\) \\*
   & LEON & \(0.20114\pm 0.03115\) & \(0.22864\pm 0.03547\) & \(0.25887\pm 0.03540\) & \(0.65157\pm 0.08699\) \\
  \midrule
  High OOD & Transformer & \(1.09402\pm 0.16812\) & \(1.76109\pm 0.06502\) & \(1.68153\pm 0.04056\) & \(4.18796\pm 0.06697\) \\*
   & Fixed operator & \(1.17646\pm 0.05379\) & \(1.71994\pm 0.04281\) & \(1.62279\pm 0.01069\) & \(3.99983\pm 0.05879\) \\*
   & Residual MLP & \(1.28488\pm 0.12990\) & \(1.75004\pm 0.03326\) & \(1.72567\pm 0.03890\) & \((6.231\pm 13.718)\times10^{4}\) \\*
   & LEON & \(0.31311\pm 0.09101\) & \(1.33494\pm 0.07668\) & \(1.52211\pm 0.08162\) & \(4.76414\pm 0.54270\) \\
\end{longtable}
\endgroup

LEON achieves the lowest mean angle RMSE at all listed horizons in all four regimes, with the largest relative improvements below the separatrix. High-OOD errors remain substantial, and LEON has higher angular-velocity RMSE than the Transformer and fixed-operator baseline in this regime. Wrapped-angle error is bounded by construction. Angular-velocity error additionally measures large state deviations that angle wrapping can obscure.

\paragraph{Error analysis.}
Table~\ref{tab:leon-f-pendulum-local} reports teacher-forced one-step errors and the median and $90$th percentile of per-trajectory $8$-s angle RMSE. The quantiles describe the trajectory distribution within each seed, whereas Table~\ref{tab:leon-f-pendulum-prediction} pools squared errors before taking the square root.

\begingroup
\small
\setlength{\tabcolsep}{3pt}
\renewcommand{\arraystretch}{1.08}
\begin{longtable}{@{\extracolsep{\fill}}llrrrr@{}}
\caption{Pendulum local and trajectory-level errors, averaged over five seeds. Median and P90 denote the median and 90th percentile of $128$ trajectory RMSE values within each seed. Angle and angular-velocity errors are in rad and rad/s, respectively.}
\label{tab:leon-f-pendulum-local}\\[-5pt]
\toprule
Regime & Predictor & \shortstack{One-step\\angle RMSE} & \shortstack{One-step\\angular-velocity RMSE} & \shortstack{Median\\$8$-s angle RMSE} & \shortstack{P90\\$8$-s angle RMSE} \\
\midrule
\endfirsthead
\multicolumn{6}{@{}l@{}}{\tablename~\thetable\ (continued)}\\[2pt]
\toprule
Regime & Predictor & \shortstack{One-step\\angle RMSE} & \shortstack{One-step\\angular-velocity RMSE} & \shortstack{Median\\$8$-s angle RMSE} & \shortstack{P90\\$8$-s angle RMSE} \\
\midrule
\endhead
\endfoot
\bottomrule
\endlastfoot
ID & Transformer & \(0.00800\) & \(0.02771\) & \(0.03006\) & \(0.04329\) \\*
   & Fixed operator & \(0.00610\) & \(0.00965\) & \(0.02167\) & \(0.05361\) \\*
   & Residual MLP & \(0.01268\) & \(0.02392\) & \(0.16951\) & \(0.40860\) \\*
   & LEON & \(0.00077\) & \(0.00254\) & \(0.00352\) & \(0.00726\) \\
  \midrule
  Near OOD & Transformer & \(0.01776\) & \(0.07083\) & \(0.24206\) & \(0.49876\) \\*
   & Fixed operator & \(0.01988\) & \(0.01525\) & \(0.23787\) & \(0.50297\) \\*
   & Residual MLP & \(0.02770\) & \(0.05671\) & \(0.77440\) & \(1.02335\) \\*
   & LEON & \(0.00263\) & \(0.00900\) & \(0.02920\) & \(0.07784\) \\
  \midrule
  Far OOD & Transformer & \(0.04100\) & \(0.16177\) & \(0.90767\) & \(1.46447\) \\*
   & Fixed operator & \(0.03682\) & \(0.02539\) & \(0.79614\) & \(1.31657\) \\*
   & Residual MLP & \(0.04734\) & \(0.09048\) & \(1.31512\) & \(1.90491\) \\*
   & LEON & \(0.00559\) & \(0.02167\) & \(0.15780\) & \(0.42949\) \\
  \midrule
  High OOD & Transformer & \(0.11561\) & \(0.37799\) & \(1.64319\) & \(2.03106\) \\*
   & Fixed operator & \(0.07322\) & \(0.04108\) & \(1.57419\) & \(1.94506\) \\*
   & Residual MLP & \(0.07589\) & \(0.15693\) & \(1.72666\) & \(1.88459\) \\*
   & LEON & \(0.01432\) & \(0.04825\) & \(1.57796\) & \(1.95898\) \\
\end{longtable}
\endgroup

\paragraph{Parameter variations.}
We additionally test transfer to changed system parameters by varying length and damping from $(L,c)=(1,0.12)$ under $\ddot\theta=-(9.81/L)\sin\theta-c\dot\theta$. The eight settings in Table~\ref{tab:leon-f-pendulum-parameters} change either parameter or both. Initial physical states are fixed across changes within each nominal energy regime, and the changed dynamics apply from the start of the measured context. Models retain their trained normalization, checkpoints, and observation interfaces, without conditioning on the changed $L$ or $c$. The terminal error, $[J^{-1}\sum_j e_{\theta,j,160}^2]^{1/2}$, evaluates the final prediction after $8\,\mathrm{s}$. This experiment complements initial-energy extrapolation by measuring sensitivity to changes in the governing dynamics. Performance depends on the parameter setting, with no uniform advantage for LEON across these changes.

\begingroup
\small
\setlength{\tabcolsep}{3pt}
\renewcommand{\arraystretch}{1.08}
\begin{longtable}{@{\extracolsep{\fill}}lrrrrrr@{}}
\caption{Pendulum parameter changes: mean terminal $8$-s wrapped-angle RMSE (rad) over five seeds and $128$ trajectories per seed and setting. Initial-energy labels use the nominal energy definition.}
\label{tab:leon-f-pendulum-parameters}\\[-5pt]
\toprule
Regime & $L$ & $c$ & Transformer & Fixed operator & MLP & LEON \\
\midrule
\endfirsthead
\multicolumn{7}{@{}l@{}}{\tablename~\thetable\ (continued)}\\[2pt]
\toprule
Regime & $L$ & $c$ & Transformer & Fixed operator & MLP & LEON \\
\midrule
\endhead
\endfoot
\bottomrule
\endlastfoot
ID & $1$ & $0.06$ & \(0.3719\) & \(0.3565\) & \(0.5756\) & \(0.3331\) \\*
   & $1$ & $0.24$ & \(0.3076\) & \(0.2815\) & \(0.2274\) & \(0.3007\) \\*
   & $0.7$ & $0.12$ & \(0.6854\) & \(0.6095\) & \(0.7118\) & \(0.5720\) \\*
   & $1.3$ & $0.12$ & \(1.0836\) & \(1.0916\) & \(0.8901\) & \(1.0977\) \\*
   & $0.7$ & $0.06$ & \(0.8789\) & \(0.8107\) & \(0.8549\) & \(0.7802\) \\*
   & $0.7$ & $0.24$ & \(0.4159\) & \(0.3865\) & \(0.4441\) & \(0.3812\) \\*
   & $1.3$ & $0.06$ & \(1.1711\) & \(1.1871\) & \(0.9653\) & \(1.1982\) \\*
   & $1.3$ & $0.24$ & \(0.8109\) & \(0.8119\) & \(0.6593\) & \(0.8164\) \\
  \midrule
  Near OOD & $1$ & $0.06$ & \(1.1943\) & \(1.1281\) & \(1.4037\) & \(0.9037\) \\*
   & $1$ & $0.24$ & \(0.5800\) & \(0.5191\) & \(0.4675\) & \(0.6976\) \\*
   & $0.7$ & $0.12$ & \(1.1271\) & \(1.0916\) & \(0.9554\) & \(1.0194\) \\*
   & $1.3$ & $0.12$ & \(1.4756\) & \(1.3911\) & \(1.1873\) & \(1.4273\) \\*
   & $0.7$ & $0.06$ & \(1.7203\) & \(1.8344\) & \(1.1080\) & \(1.7888\) \\*
   & $0.7$ & $0.24$ & \(0.4086\) & \(0.3432\) & \(0.7906\) & \(0.5125\) \\*
   & $1.3$ & $0.06$ & \(1.5469\) & \(1.4694\) & \(1.2594\) & \(1.6991\) \\*
   & $1.3$ & $0.24$ & \(1.1776\) & \(1.1371\) & \(1.1235\) & \(1.0056\) \\
  \midrule
  Far OOD & $1$ & $0.06$ & \(1.9502\) & \(1.9104\) & \(1.6012\) & \(1.7466\) \\*
   & $1$ & $0.24$ & \(0.4624\) & \(0.4527\) & \(1.3861\) & \(1.0831\) \\*
   & $0.7$ & $0.12$ & \(1.7292\) & \(1.7606\) & \(1.3888\) & \(1.2080\) \\*
   & $1.3$ & $0.12$ & \(1.5458\) & \(1.4849\) & \(1.5788\) & \(1.8194\) \\*
   & $0.7$ & $0.06$ & \(1.1889\) & \(1.4462\) & \(1.1910\) & \(1.7260\) \\*
   & $0.7$ & $0.24$ & \(0.8915\) & \(0.7905\) & \(1.1914\) & \(0.8368\) \\*
   & $1.3$ & $0.06$ & \(1.1126\) & \(1.1375\) & \(1.4470\) & \(1.6944\) \\*
   & $1.3$ & $0.24$ & \(1.1784\) & \(1.0442\) & \(1.4858\) & \(1.0007\) \\
  \midrule
  High OOD & $1$ & $0.06$ & \(1.9836\) & \(1.9803\) & \(1.9692\) & \(1.9176\) \\*
   & $1$ & $0.24$ & \(0.7093\) & \(0.6438\) & \(1.5582\) & \(1.5469\) \\*
   & $0.7$ & $0.12$ & \(1.3740\) & \(1.3639\) & \(1.4677\) & \(1.8935\) \\*
   & $1.3$ & $0.12$ & \(1.6610\) & \(1.7103\) & \(1.5892\) & \(1.8961\) \\*
   & $0.7$ & $0.06$ & \(1.6871\) & \(1.6622\) & \(1.8899\) & \(1.6889\) \\*
   & $0.7$ & $0.24$ & \(1.3148\) & \(1.2288\) & \(1.3641\) & \(1.6102\) \\*
   & $1.3$ & $0.06$ & \(1.8667\) & \(1.8890\) & \(1.7259\) & \(1.7946\) \\*
   & $1.3$ & $0.24$ & \(1.2890\) & \(1.2542\) & \(1.4507\) & \(1.3830\) \\
\end{longtable}
\endgroup

\subsection{Driven Oscillator}
\label{app:leon-driven-oscillator}

\paragraph{System dynamics.}
The unit-mass driven oscillator follows $\dot q=v$ and $\dot v=-\kappa q-0.12v+u(t)$. At fixed stiffness $\kappa$, it is linear in the state and externally forced. For a force held constant over one sampling interval, its exact transition is
\begin{equation}
 \begin{aligned}
  \mathbf s_{\ell+1}&=K_\kappa\mathbf s_\ell+G_\kappa u_\ell,
  \qquad \mathbf s=(q,v)^\top,\\
  K_\kappa&=\exp(\Delta t F_\kappa),
  \qquad G_\kappa=\int_0^{\Delta t}\exp(\tau F_\kappa)\mathbf b_u\,d\tau,\\
  F_\kappa&=\begin{pmatrix}0&1\\-\kappa&-0.12\end{pmatrix},
  \qquad \mathbf b_u=(0,1)^\top.
 \end{aligned}
 \label{eq:leon-f-oscillator-flow}
\end{equation}
The separation between stiffness-dependent state propagation and the forced response provides a controlled setting for evaluating LEON's conditional matrix family and complementary term. Extrapolation to unseen stiffness values tests conditional prediction across system parameters. Retrained ablations and test-time interventions assess the roles of the learned update terms.

\paragraph{Dataset.}
Stiffness is constant within each trajectory and sampled uniformly from $[1,4]$ for training, validation, and ID testing, and from $[4.5,6]$ for OOD testing. Initial position and velocity are sampled uniformly from $[-1,1]$ and $[-1.25,1.25]$, respectively. The force is sampled uniformly from $[-1.2,1.2]$ and held constant for eight sampling intervals. Trajectories contain no observation noise. States are divided by their training root-mean-square scale plus $10^{-6}$, without mean subtraction. Stiffness is normalized as $\bar\kappa=(\kappa-2.5)/1.5$. All predictors are evaluated on the same trajectories with the same stiffness values and recorded force sequences.

The operator coefficients are $\boldsymbol\xi_{\mathrm{op}}=f_\kappa(\bar\kappa)$, where $f_\kappa$ is a learned SiLU network of widths $(1,32,4)$ with a linear output. The complementary coefficients are obtained by fusing the current observable with the encoded pair $(\bar\kappa,u_\ell)$. The complementary term is the only component of the full predictor that uses the force $u_\ell$ applied during the current prediction interval. Each physical prediction uses one internal Euler update ($S=1$, $\eta=1$). Training supervises eight successive physical predictions. Evaluation predicts $80$ samples after the four-sample context.

\paragraph{Regularization.}
The training loss is normalized-state rollout MSE over eight physical prediction steps plus a stiffness-curvature penalty with weight $\lambda=10$. The penalty uses scaled second finite differences at three normalized-stiffness probes with spacing $0.1$, while keeping states and forces fixed. Probe centers are clipped to $[-0.8,0.8]$ so that all probes lie within the training range. LEON regularizes the operator and complementary increments separately. The MLP and Transformer regularize their total predicted increment. This penalty encourages smooth dependence on stiffness. Although the continuous-time generator is affine in $\kappa$, the exact finite-time matrices in Eq.~\eqref{eq:leon-f-oscillator-flow} generally depend nonlinearly on $\kappa$. The penalty therefore provides a soft inductive bias for conditional generalization. At test time, the learned networks are evaluated directly at the supplied stiffness values.

\paragraph{Retrained ablations.}
Table~\ref{tab:leon-f-oscillator-comparison} compares LEON with parameter-matched baselines and independently retrained variants. Each variant is trained from scratch and evaluated using its own checkpoint selected according to Section~\ref{app:leon-diagnostic-protocol}. The \emph{w/o additive} variant is trained and evaluated with $b=0$. At fixed state and stiffness, its prediction is identical for all values of $u_\ell$. The increase in rollout error therefore reflects the combined effect of removing the complementary update and its force input. The w/o modulation variant replaces the stiffness coefficient network with four learned constants and retains the stiffness and force inputs to the complementary term. The w/o curvature variant preserves the full architecture, data, initialization, optimizer settings, and model selection criterion, changing only the regularization weight to zero.

\begingroup
\small
\setlength{\tabcolsep}{3pt}
\renewcommand{\arraystretch}{1.08}
\begin{longtable}{@{\extracolsep{\fill}}lrrrrr@{}}
\caption{Driven-oscillator prediction results over five seeds. Rollout RMSE pools $80$ prediction steps and both state coordinates. One-step errors are OOD means. Active parameter counts exclude modules unused in prediction. All ablations in this table are independently retrained and use their own selected checkpoints.}
\label{tab:leon-f-oscillator-comparison}\\[-5pt]
\toprule
Predictor & \shortstack{Active\\parameters} & ID RMSE & OOD RMSE & \shortstack{One-step\\MAE} & \shortstack{One-step\\$\ell_2$ error} \\
\midrule
\endfirsthead
\multicolumn{6}{@{}l@{}}{\tablename~\thetable\ (continued)}\\[2pt]
\toprule
Predictor & \shortstack{Active\\parameters} & ID RMSE & OOD RMSE & \shortstack{One-step\\MAE} & \shortstack{One-step\\$\ell_2$ error} \\
\midrule
\endhead
\endfoot
\bottomrule
\endlastfoot
Full predictor & $32{,}592$ & \(0.000713\pm 0.000138\) & \(0.002873\pm 0.001138\) & \(0.000119\) & \(0.000205\) \\*
  w/o additive (retrained) & $12{,}932$ & \(0.341472\pm 0.015257\) & \(0.312320\pm 0.029392\) & \(0.016275\) & \(0.031078\) \\*
  w/o modulation (retrained) & $32{,}400$ & \(0.011484\pm 0.003058\) & \(0.035376\pm 0.021153\) & \(0.000674\) & \(0.001292\) \\*
  Residual MLP & $32{,}594$ & \(0.015334\pm 0.003464\) & \(0.023635\pm 0.008982\) & \(0.000819\) & \(0.001433\) \\*
  Transformer & $32{,}584$ & \(0.019594\pm 0.004148\) & \(0.099364\pm 0.017486\) & \(0.001787\) & \(0.003011\) \\*
  w/o curvature (retrained) & $32{,}592$ & \(0.002280\pm 0.000606\) & \(0.061265\pm 0.027462\) & \(0.000889\) & \(0.001465\) \\
\end{longtable}
\endgroup

LEON achieves the lowest mean ID and OOD rollout RMSE among these methods. The increased OOD error without curvature regularization shows that this penalty contributes to the reported extrapolation performance. OOD one-step MAE averages coordinate-wise absolute errors. One-step $\ell_2$ error averages the Euclidean norm of each state-error vector. Both metrics use measured histories.

\paragraph{Test-time interventions.}
The driven-oscillator ablation curves in Figure~\ref{fig:controlled-tests}(b) reuse the five trained full-model checkpoints without retraining. For \emph{frozen modulation}, the operator coefficients are fixed at $\boldsymbol\xi_{\mathrm{op}}=f_\kappa(0)$, corresponding to the reference stiffness $\kappa_{\mathrm{ref}}=2.5$. The complementary term retains the actual stiffness and force inputs. This intervention gives a mean 4-s OOD RMSE of $0.894652$ and corresponds to the curve labelled \emph{w/o modulation}. The retrained w/o modulation variant in Table~\ref{tab:leon-f-oscillator-comparison} instead learns constant operator coefficients from scratch and obtains $0.035376\pm0.021153$.

For test-time branch removal, we set either $A(\boldsymbol\xi_{\mathrm{op}})\mathbf h$ or $b(\mathbf h,\boldsymbol\xi_b)$ to zero while retaining the identity residual. These interventions correspond to the \emph{w/o operator} and \emph{w/o additive} curves, respectively, and use the same full-model checkpoints, OOD trajectories, and measured conditions. Table~\ref{tab:leon-f-oscillator-knockouts} reports their aggregate results. Relative RMSE is the ratio of the mean rollout RMSE under each intervention to that of the full predictor over the same five seeds. With the complementary term disabled, the predictor has no direct access to $u_\ell$, and its mean 4-s OOD rollout RMSE is $101.59$ times that of the full predictor. These test-time interventions assess sensitivity of the fitted predictor and are distinct from the independently retrained ablations in Table~\ref{tab:leon-f-oscillator-comparison}. They do not establish independent performance contributions or unique physical interpretations for the update terms.

\begingroup
\small
\setlength{\tabcolsep}{3pt}
\renewcommand{\arraystretch}{1.08}
\begin{longtable}{@{\extracolsep{\fill}}lrr@{}}
\caption{Driven-oscillator test-time branch removal. All rows reuse the same five trained full-model checkpoints and OOD trajectories without retraining. Relative RMSE is the ratio of cross-seed mean errors.}
\label{tab:leon-f-oscillator-knockouts}\\[-5pt]
\toprule
Evaluation condition & OOD RMSE & Relative RMSE \\
\midrule
\endfirsthead
\multicolumn{3}{@{}l@{}}{\tablename~\thetable\ (continued)}\\[2pt]
\toprule
Evaluation condition & OOD RMSE & Relative RMSE \\
\midrule
\endhead
\endfoot
\bottomrule
\endlastfoot
Full predictor & \(0.002873\pm 0.001138\) & $1.00\times$ \\*
  Operator off & \(1.116478\pm 0.028740\) & $388.63\times$ \\*
  Complementary term off & \(0.291857\pm 0.016952\) & $101.59\times$ \\
\end{longtable}
\endgroup

\paragraph{Per-seed results.}
Table~\ref{tab:leon-f-oscillator-seeds} reports OOD rollout RMSE for each seed. The test-time branch-removal results use the corresponding full-model checkpoint, enabling paired comparison within each run.

\begingroup
\small
\setlength{\tabcolsep}{3pt}
\renewcommand{\arraystretch}{1.08}
\begin{longtable}{@{\extracolsep{\fill}}rrrrrr@{}}
\caption{Driven-oscillator OOD rollout RMSE for each training seed. The final two columns report test-time branch removal using the corresponding full-model checkpoint without retraining.}
\label{tab:leon-f-oscillator-seeds}\\[-5pt]
\toprule
Seed & Full predictor & Transformer & Residual MLP & Operator off & \shortstack{Complementary\\term off} \\
\midrule
\endfirsthead
\multicolumn{6}{@{}l@{}}{\tablename~\thetable\ (continued)}\\[2pt]
\toprule
Seed & Full predictor & Transformer & Residual MLP & Operator off & \shortstack{Complementary\\term off} \\
\midrule
\endhead
\endfoot
\bottomrule
\endlastfoot
1400 & $0.003927$ & $0.089964$ & $0.027824$ & $1.132572$ & $0.287439$ \\*
  1401 & $0.004144$ & $0.116122$ & $0.016376$ & $1.115638$ & $0.313232$ \\*
  1402 & $0.001444$ & $0.085920$ & $0.036161$ & $1.155046$ & $0.302001$ \\*
  1403 & $0.002352$ & $0.084317$ & $0.013862$ & $1.082214$ & $0.288382$ \\*
  1404 & $0.002498$ & $0.120499$ & $0.023952$ & $1.096918$ & $0.268233$ \\
\end{longtable}
\endgroup

\paragraph{Condition swapping.}
OOD stiffness values are paired by reversed stiffness order. The physical state at the forecast origin, measured context, and future forces are held fixed, and reference futures are generated under both stiffness values. Let $e_{\alpha\beta,j,r}$ denote trajectory RMSE when conditioning on stiffness $\alpha$ and evaluating against the future under stiffness $\beta$, for trajectory $j$ and seed $r$. With $N=5$ seeds and $J=128$ trajectories, the symmetric errors are
\begin{equation}
 \begin{aligned}
 R_{\mathrm{matched}}&=\frac{1}{2NJ}\sum_{r,j}
   (e_{aa,j,r}+e_{bb,j,r}),\\
 R_{\mathrm{cross}}&=\frac{1}{2NJ}\sum_{r,j}
   (e_{ab,j,r}+e_{ba,j,r}).
 \end{aligned}
 \label{eq:leon-f-condition-sensitivity}
\end{equation}
The swap penalty is $R_{\mathrm{cross}}-R_{\mathrm{matched}}$. Retrieval averages $\tfrac12[\mathbb I(e_{aa}<e_{ab})+\mathbb I(e_{bb}<e_{ba})]$ over trajectories and seeds, testing whether each predicted future is closer to the reference generated under its supplied stiffness. These metrics average per-trajectory RMSE. The rollout RMSE in Table~\ref{tab:leon-f-oscillator-comparison} pools squared errors before taking the square root. Table~\ref{tab:leon-f-oscillator-swap} reports the results. LEON combines the lowest matched error with the highest retrieval rate among the compared predictors. The retrained w/o modulation variant remains sensitive to stiffness through its complementary branch, so retrieval alone does not isolate operator modulation.

\begingroup
\small
\setlength{\tabcolsep}{3pt}
\renewcommand{\arraystretch}{1.08}
\begin{longtable}{@{\extracolsep{\fill}}lrrrr@{}}
\caption{Driven-oscillator condition sensitivity, averaged symmetrically over both stiffness assignments and five seeds. Retrieval is reported as a percentage. Retrained variants use their own checkpoints.}
\label{tab:leon-f-oscillator-swap}\\[-5pt]
\toprule
Predictor & Matched RMSE & Cross RMSE & Swap penalty & Retrieval (\%) \\
\midrule
\endfirsthead
\multicolumn{5}{@{}l@{}}{\tablename~\thetable\ (continued)}\\[2pt]
\toprule
Predictor & Matched RMSE & Cross RMSE & Swap penalty & Retrieval (\%) \\
\midrule
\endhead
\endfoot
\bottomrule
\endlastfoot
w/o modulation (retrained) & \(0.02043\) & \(0.24111\) & \(0.22068\) & $97.03$ \\*
  w/o additive (retrained) & \(0.29260\) & \(0.38666\) & \(0.09406\) & $69.30$ \\*
  Transformer & \(0.05941\) & \(0.27042\) & \(0.21101\) & $93.12$ \\*
  Residual MLP & \(0.01763\) & \(0.24778\) & \(0.23014\) & $98.36$ \\*
  Full predictor & \(0.00219\) & \(0.25187\) & \(0.24968\) & $99.61$ \\
\end{longtable}
\endgroup

\par
\endgroup